\documentclass[]{fairmeta}
\usepackage{xcolor}         % Removed dvipsnames option - colors defined below
\usepackage{wrapfig}
\usepackage{tabularx}
\usepackage{textcomp}
\usepackage{stfloats}
\usepackage{url}
\usepackage{verbatim}
\usepackage{graphicx}
\usepackage{titlesec}
\usepackage{tocloft}
\usepackage{adjustbox}
\usepackage{multirow}
\usepackage{pifont}
\usepackage{tikz}
\usepackage{comment}
\usepackage{colortbl}  %
\usepackage{booktabs} 
\usepackage{hyperref}
\usepackage{graphicx}    %
\usepackage{subcaption} 
\usepackage{amsmath,amssymb} %
\usepackage{todonotes}
\RequirePackage{xspace}
\makeatletter
\DeclareRobustCommand\onedot{\futurelet\@let@token\@onedot}
\def\@onedot{\ifx\@let@token.\else.\null\fi\xspace}

\makeatother

\definecolor{blue}{RGB}{0,0,255}
\definecolor{red}{RGB}{255,0,0}
\definecolor{orange}{RGB}{255,165,0}
\definecolor{RoyalBlue}{RGB}{65,105,225}
\definecolor{OliveGreen}{RGB}{107,142,35}

\definecolor{adptorange}{RGB}{248, 205, 172}
\definecolor{cmpblue}{RGB}{189, 215, 238}
\definecolor{cmpblue}{RGB}{189, 215, 238}

\definecolor{our_red}{RGB}{232,157,160}
\definecolor{our_blue}{RGB}{136,206,230}
\definecolor{our_orange}{RGB}{246,200,168}
\definecolor{our_green}{RGB}{178,211,164}

\definecolor{attn_code0}{RGB}{247,215,200}
\definecolor{attn_code1}{RGB}{238,169,139}
\definecolor{mlp_code0}{RGB}{204,201,221}
\definecolor{mlp_code1}{RGB}{102,95,153}

\definecolor{dark_green}{rgb}{0, 0.5, 0}
\definecolor{dark_red}{rgb}{0.8, 0.2, 0.2}
\definecolor{soft_red}{rgb}{1.0, 0.4, 0.4}
\definecolor{light_blue}{rgb}{0.2, 0.5, 1.0}

\usepackage{algorithm}
\usepackage{algorithmicx}
\usepackage{algpseudocode}

\renewcommand{\Comment}[1]{\hfill\textcolor{light_blue}{// #1}}

\definecolor{token_blue}{RGB}{84, 120, 140}

\usepackage{pifont}       %
\usepackage{bbding}       %
\usepackage{fontawesome}

\usepackage{float}
\usepackage{siunitx}        %
\usepackage{microtype}      %
\usepackage{cleveref}      %
\usepackage{enumitem}
\usepackage[inkscapelatex=false]{svg}
\usepackage{animate}
\usepackage{etoc}          %

\definecolor{darkgreen}{rgb}{0.15, 0.75, 0.15}
\definecolor{cvprblue}{rgb}{0.21,0.49,0.74}
\definecolor{lightblue}{rgb}{0.90, 0.95, 0.99}

\algrenewcommand\algorithmicrequire{\textbf{Input:}}
\algrenewcommand\algorithmicensure{\textbf{Output:}}

\newcommand{\our}{\texttt{FPSAttention}\xspace}

\title{
  FPSAttention: Training-Aware FP8 and Sparsity Co-Design for Fast Video Diffusion
}

\author[*1,2,3]{Akide Liu}
\author[*2]{Zeyu Zhang}
\author[\dag2,4]{Zhexin Li}
\author[\dag3]{Xuehai Bai}
\author[2]{Yizeng Han}
\author[\ddag24]{Jiasheng Tang}
\author[2]{Yuanjie Xing}
\author[2,4]{Jichao Wu}
\author[2,4]{Mingyang Yang}
\author[2]{Weihua Chen}
\author[3]{Jiahao He}
\author[3]{Yuanyu He}
\author[2]{Fan Wang}
\author[1]{Gholamreza Haffari}
\author[\ddag2,3]{Bohan Zhuang}

\affiliation[1]{Monash University}
\affiliation[2]{DAMO Academy, Alibaba Group}
\affiliation[3]{ZIP Lab, Zhejiang University}
\affiliation[4]{Hupan Lab}
\contribution[*]{Co-first authors}
\contribution[\dag]{Co-second authors}
\contribution[\ddag]{Project leads}
\contribution[]{Details see Sec.\ref{sec:contributions}.}

\abstract{
    Diffusion generative models have become the standard for producing high-quality, coherent video content, yet their slow inference speeds and high computational demands hinder practical deployment. Although both quantization and sparsity can independently accelerate inference while maintaining generation quality, naively combining these techniques in existing training-free approaches leads to significant performance degradation, as they fail to achieve proper joint optimization.
    We introduce \texttt{FPSAttention}, a novel training-aware co-design of \texttt{FP}8 quantization and \texttt{S}parsity for video generation, with a focus on the 3D bi-directional attention mechanism. Our approach features three key innovations: 1) A unified 3D tile-wise granularity that simultaneously supports both quantization and sparsity. 2) A denoising step-aware strategy that adapts to the noise schedule, addressing the strong correlation between quantization/sparsity errors and denoising steps. 3) A native, hardware-friendly kernel that leverages FlashAttention and is implemented with optimized Hopper architecture features, enabling highly efficient execution.
    Trained on Wan2.1's 1.3B and 14B models and evaluated on the VBench benchmark, \texttt{FPSAttention} achieves a 7.09$\times$ kernel speedup for attention operations and a 4.96$\times$ end-to-end speedup for video generation compared to the BF16 baseline at 720p resolution—without sacrificing generation quality.}

\date{\today} 
\metadata[Project page and code]{{\href{https://fps.ziplab.co}{\texttt{https://fps.ziplab.co}}}}
\metadata[Correspondence]{\email{fan.w@alibaba-inc.com, bohan.zhuang@gmail.com}}

\begin{document}
\thispagestyle{firstheader}
\maketitle

\section{Introduction} 

\begin{figure}[t]
\centering
\includegraphics[width=\textwidth,page=1]{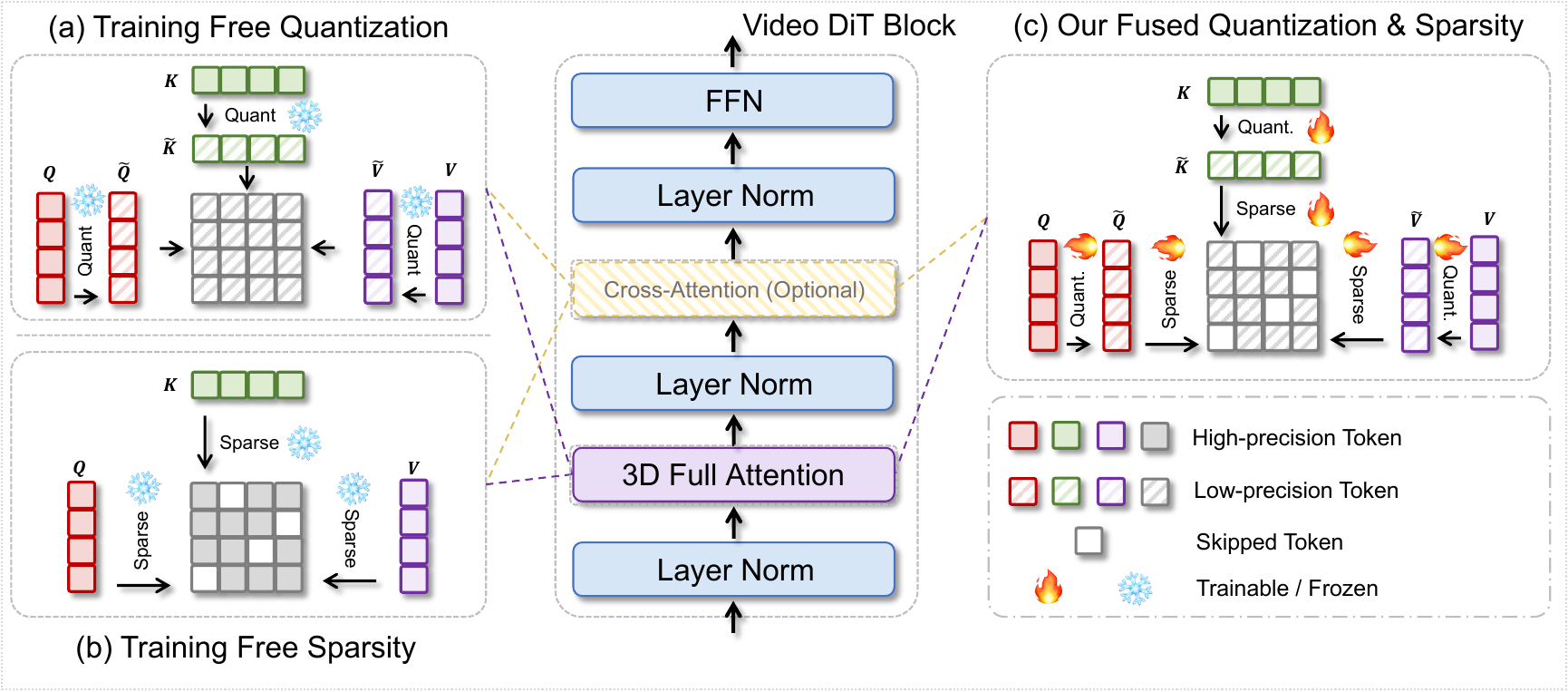}
\vskip -0.1in
\caption{
Comparing previous training-free quantization (a) and training-free sparsity (b) approaches reveals substantial accuracy degradation and a lack of compatibility when used independently. In contrast, our 
\our framework (c) integrates low-precision and sparse patterns in a single training process, yielding near-zero accuracy loss and seamless deployment.
}
\label{fig:intro-compare}
\vspace{-1em}
\end{figure}

\begin{wraptable}{r}{0.3\textwidth}
    \centering
    \small
    \setlength{\tabcolsep}{4pt}
    \vspace{-1em}
    \begin{tabular}{>{\centering\arraybackslash}m{1.2cm}|>{\centering\arraybackslash}m{1.1cm}|>{\centering\arraybackslash}m{1.1cm}}
    \hline
    \textbf{Method} & \textbf{Kernel Speedup} & \textbf{E2E Speedup} \\
    \hline
    BF16 & 1.00$\times$ & 1.00$\times$ \\
    FP8 & 1.84$\times$ & 1.26$\times$ \\
    STA & 5.15$\times$ & 3.60$\times$ \\
    \textbf{FPSAtten} & \textbf{7.09$\times$} & \textbf{4.96$\times$} \\
    \hline
    \end{tabular}
    \caption{
        Efficiency comparison of the BF16 baseline, FP8 quantization, STA sparse attention, and our \our  on Wan2.1-14B at 720p resolution on an NVIDIA H20 GPU. We report both kernel-level and end-to-end speedups relative to the BF16 baseline.
    }
        \vspace{-1em}
    \label{tab:speedup}
    \end{wraptable}

Diffusion models have revolutionized AI through breakthrough image synthesis~\cite{jcst2024,dhariwal2021diffusion,singer2022make} and are advancing into complex video generation~\cite{wang2024artificial,xing2024survey}.
Diffusion Transformers (DiTs)~\cite{peebles2023scalable} now enable efficient, high-quality synthesis~\cite{noarxiv_align,ma2024latte_tmlr}, powering billion-parameter models like Wan2.1~\cite{wang2025wan} that produce coherent, long-duration, high-fidelity videos.

\begin{wrapfigure}{r}{0.35\textwidth}
    \centering
    \vspace{-1em}
    \begin{minipage}{\linewidth}
        \centering
        \animategraphics[width=\linewidth,loop]{10}{video/dog/dog1/dog-}{0}{12}
    \end{minipage}
    
    \vspace{0.5em}
    
    \begin{minipage}{\linewidth}
        \centering
        \animategraphics[width=\linewidth,loop]{10}{video/dog/dog2/dog-}{0}{12}
    \end{minipage}
    \caption{Comparison of video generation results. Top: Training Free FP8 + STA. Bottom: Our Training-Aware \our . Click the image to play the video via Acrobat Reader.}
    \label{fig:intro-video-comparison}
    \vspace{-1em}
\end{wrapfigure}

Despite progress, crippling computational demands persist~\cite{shen2025efficient}: (i) iterative reverse-diffusion requiring hundreds of steps, and (ii) quadratic-complexity spatio-temporal attention ($\mathcal{O}(N^2)$, 
$N$
denotes the number of tokens)~\cite{lu2023vdt,kong2024hunyuanvideo,wang2025wan}.
Particularly, the computational burden of attention becomes prohibitive for high-resolution, long-duration videos, often consuming $>$70\% of inference time \cite{zhang2025fastvideo,xi2025sparse,vaswani2017attention}.
For instance, Wan2.1-14B requires approximately 2.5 hours on an NVIDIA H20 GPU to generate a 5s video. 

To address the efficiency challenge, numerous acceleration methodologies have been proposed \cite{ding2024dollar, ma2024deepcache, shih2023parallel}, among which quantization and sparsity have emerged as predominant techniques \cite{wang2023towards, manduchi2024challenges}.
Quantization reduces numerical precision (e.g., FP32$\to$INT8/FP8), thereby decreasing the memory footprint and enabling faster computations (\Cref{fig:intro-compare}~(a)). The recent post-training quantization (PTQ) method SageAttention \cite{zhang2024sageattention}
quantizes attention modules into INT8 with calibration strategies, providing moderate acceleration with downgraded generation quality. 
Compared to INT8, the emerging format FP8 quantization offers a wider dynamic range\cite{micikevicius2022fp8}, facilitating both training and inference \cite{kuzmin2022fp8, shen2024efficient}.  
Nevertheless, training-free FP8 quantization, despite its theoretical advantages, introduces significant quantization errors that degrade model performance \cite{zhang2024sageattention}. 
Apart from quantization,
sparsity techniques address the quadratic computational complexity of 3D full attention by selectively skipping computations, as illustrated in \Cref{fig:intro-compare}~(b).
Representatively, Sparse-VideoGen \cite{xi2025sparse} implements per-head spatial-temporal masks aligned with GPU blocks, SpargeAttn \cite{zhang2025spargeattn} employs a two-stage filtering mechanism, and Sliding Tile Attention (STA) \cite{zhang2025fastvideo} leverages local 3D sliding windows with kernel-level optimizations.

To enjoy the benefits of both worlds, a straightforward strategy is jointly applying FP8 quantization and sparsity. 
However, a naive combination 
presents significant challenges, as quantization errors can be magnified when combined with sparsity mechanisms \cite{xie2024jointsq}, as shown in \Cref{fig:intro-video-comparison}. Intuitively, sparsity techniques prioritize token retention with high-magnitude attention scores, while quantization disproportionately introduces errors in these high-magnitude values. This intrinsic tension necessitates holistic approaches that consider both techniques simultaneously, potentially framing sparsification as a specialized form of 0-bit quantization to achieve optimal balance between efficiency and generation quality \cite{motetti2024joint}.

Furthermore, existing approaches largely overlook training-aware joint optimization of quantization and sparsity, creating a substantial training-inference gap.  Our empirical analysis (Section~\ref{sec:denoising-step-aware}) reveals that diffusion models can tolerate and even correct for hardware-friendly tile-wise errors.
The error resilience is particularly pronounced when the model is aware of approximations via quantization-aware training (QAT), ensuring consistent performance 
at inference time.

In this paper, we introduce \our, as shown in \Cref{fig:intro-compare}(c), a novel training-aware co-design framework that synergistically integrates FP8 quantization and structured sparsity for 3D attention in video DiTs.

\our proposes three key innovations:
\begin{itemize} [leftmargin=*]
    \item \emph{Unifying Tile-wise Operations}: Implementing a 3D tile-wise granularity for both FP8 quantization and block sparsity (\cref{sec:joint-tile-wise-quantization-and-sparse-attention}), which directly aligns with efficient hardware execution patterns (e.g., GPU Tensor Cores) and forms the basis for structured acceleration. 
    
    \item \emph{Denoising Step-Aware Scheduling}: Introducing an adaptive strategy (\cref{sec:denoising-step-aware}) that dynamically adjusts quantization and sparsity granularity according to the varying error sensitivity and corrective capacity of the model across different denoising timesteps.
    
    \item \emph{Hardware-Optimized Kernel Design}: Developing a native, high-performance kernel (\cref{sec:hardware-optimized-kernel}) leveraging features like FlashAttention and NVIDIA Hopper architecture optimizations to translate theoretical FLOPs reduction into tangible wall-clock speedups.
    
\end{itemize}

As demonstrated in Table~\ref{tab:speedup}, by training on Wan2.1's 1.3B and 14B models and evaluating on the vBench benchmark, FPSAttention achieves a \textbf{7.09}$\times$ kernel speedup for attention operations and a \textbf{4.96}$\times$ end-to-end speedup for video generation compared to the BF16 baseline, all without sacrificing generation quality, significantly outperforming approaches that apply quantization (1.84$\times$ kernel speedup) or sparsity (5.15$\times$ kernel speedup) independently. Our work not only provides a practical solution for accelerating video diffusion but also offers a new perspective on the robustness of diffusion models to aggressive, structured compression.

\section{Related Work}

\paragraph{Quantization for video generation models.}

The computational expense of video generation models, particularly Diffusion Transformers (DiTs) \cite{kong2024hunyuanvideo}, driven by iterative sampling \cite{ho2020denoising} and quadratic attention complexity \cite{vaswani2017attention}, necessitates model quantization techniques \cite{shen2025efficient}. Post-Training Quantization (PTQ) has been explored for its efficiency \cite{he2023ptqd,li2025svdquantabsorbingoutlierslowrank, huang2024tcaqdm, zhao2024mixdqmemoryefficientfewst,huang2024tfmqdmtemporalfeaturemaintenance}; however, applying PTQ to video DiTs presents unique challenges beyond standard image models \cite{wu2024ptq4dit, chen2024q, zhao2024vidit}. Temporal variability of activation statistics across denoising steps \cite{huang2025tcaq} has prompted PTQ methodologies to implement time-step-wise calibration \cite{yi2025hardware}, adaptive quantization, and dynamic smoothing techniques \cite{shao2025tr}. Recent work has evaluated these techniques on standardized benchmarks (e.g., VBench \cite{huang2024vbench}), assessing temporal consistency alongside perceptual quality. While PTQ approaches show promising results, Quantization-Aware Training (QAT) for video diffusion models remains largely unexplored. Our work addresses this gap by introducing an FP8 QAT framework that jointly optimizes quantization and sparsity,  %
enabling efficient video generation while maintaining visual fidelity.

\paragraph{Sparse attention for video generation models.}

Recent advancements in sparse video generation and efficient attention mechanisms have improved memory utilization and computational efficiency. Sparse VideoGen \cite{xi2025sparse} leverages sparsely sampled motion priors to produce realistic videos while reducing temporal redundancy. Efficient attention mechanisms have proven crucial for handling long-range dependencies in video data. Sliding Tile Attention\cite{zhang2025fastvideo} introduces a tiled sparse attention mechanism for modeling spatial-temporal correlations, while SpargeAttn \cite{zhang2025spargeattn} proposes progressive sparsification by selectively pruning attention tokens based on importance scores. DiTFastAttn \cite{chen2024ditfastattn} accelerates attention computation by dynamically filtering irrelevant patches, achieving significant speedups without compromising quality. These approaches illustrate the trend of combining structured sparsity with content-aware selection for scalable video generation. However, these methods are typically limited to inference-time acceleration, lack integration with model training procedures, and are not fully compatible with %
quantization techniques, creating challenges for developing holistically efficient video generation frameworks.

\newcommand{\ourname}{FPSAttention} %

\section{Method}
\label{method}

\begin{figure}[t]
  \centering
  \includegraphics[width=\textwidth,page=2]{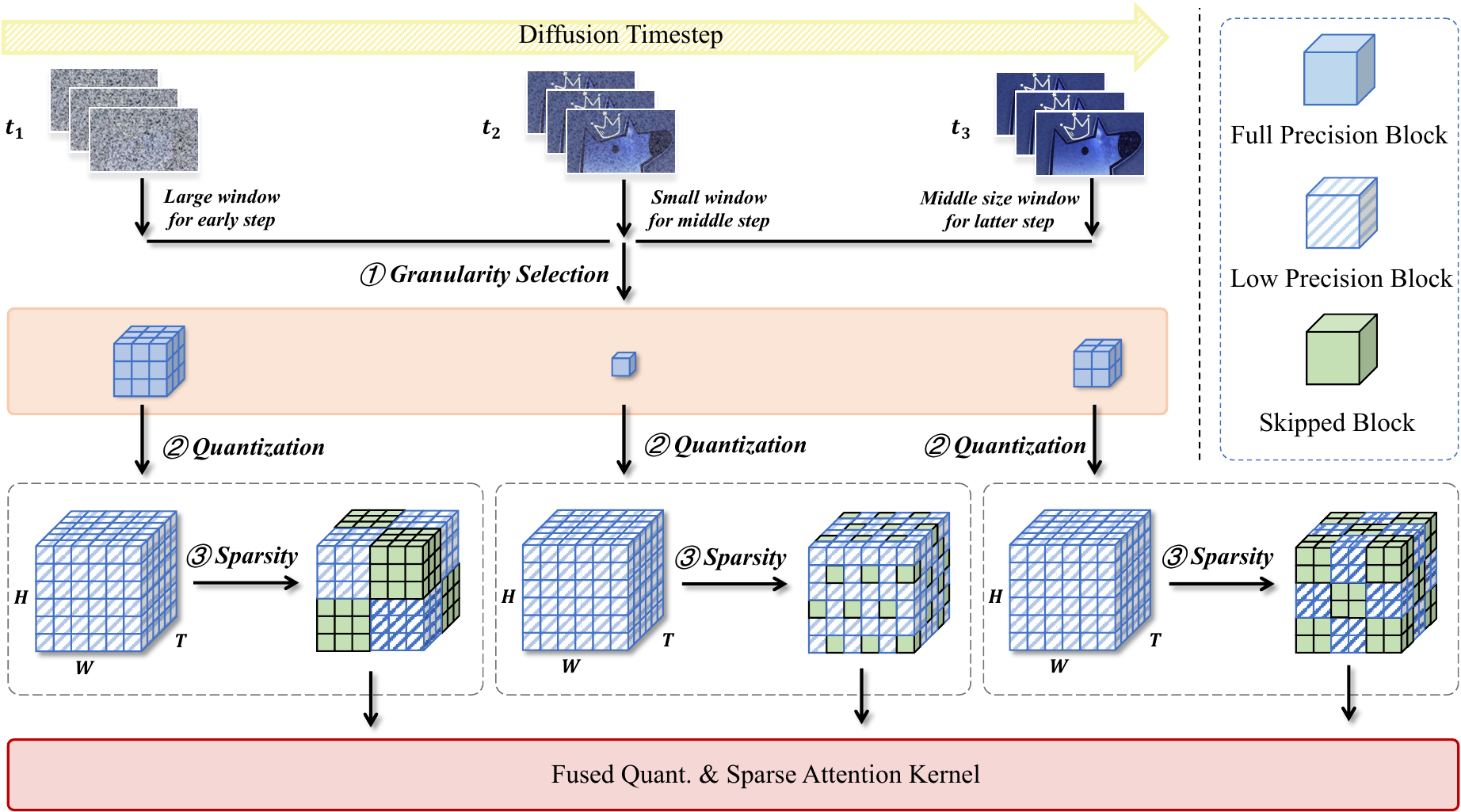}
  \caption{
  Overview of \texttt{\ourname{}}. (1) Our approach synergistically optimizes joint quantization and sparsity patterns within the attention mechanism for efficient  video generation. (2) We introduce a novel denoising step-aware strategy that dynamically adapts the granularity throughout the diffusion process, balancing computational efficiency and perceptual fidelity. Empirical observations are shown in Figure~\ref{fig:FPSAttention_adaptive_schedule}. (3) A fused hardware-friendly kernel is applied for attention operations. 
  }
  \label{fig:method-overview-fp8}
  \vspace{-1em}
\end{figure}

Our technique integrates algorithmic innovation with hardware-conscious kernel optimization to enhance the efficiency of video DiTs. This section begins by establishing fundamental concepts essential to our methodology. Subsequently, we introduce the architecture of our proposed \texttt{\ourname{}}, detailing its two primary algorithmic contributions: a unified tile-wise quantization and sparse attention mechanism, and a denoising step-aware strategy for dynamic adaptation of quantization and sparsity %
hyperparameters.
Finally, we outline our hardware-optimized kernel implementation that plays a crucial role in translating theoretical computational savings into practical efficiency gains. Figure~\ref{fig:method-overview-fp8} provides a high-level conceptual overview of our \texttt{\ourname{}} framework.

\subsection{Background}

This subsection establishes the two foundational techniques that underpin our methodology: 8-bit floating-point (FP8) quantization and Sliding Tile Attention (STA).

\textbf{FP8 quantization.} 
Video Diffusion Transformers (DiTs) process $L = T \times H \times W$ spatiotemporal tokens, where $T$, $H$, and $W$ represent temporal frames, height, and width dimensions. To reduce memory bandwidth requirements for activation tensors, FP8 quantization approximates each value $X_{i,j}$ using an 8-bit floating-point representation. Unlike INT8 quantization, which maps continuous values to a scaled integer grid, %
FP8 conversion preserves the floating-point nature by utilizing dedicated sign, exponent, and mantissa bits (in formats such as E4M3 or E5M2). 

The FP8 conversion employs a scaling factor $s_g$ for each tile of values $g$ to map the original values into the representable dynamic range of FP8:
\begin{equation}
\hat{X}_{\text{FP8}}(X_{i,j}; s_g) = \text{dequantize}(\text{FP8\_convert}(X_{i,j} \cdot s_g)) / s_g.
\end{equation}

To enhance approximation accuracy, tile-wise FP8 quantization employs per-tile scaling factors $\{s_g\}$ that minimize quantization error within each specific tile. This approach preserves attention head-specific and frame-specific activation dynamics while typically reducing data size by half (e.g., from 16-bit to 8-bit). The result is a theoretical $2\times$ reduction in memory bandwidth requirements, with further effective improvements achievable through specialized FP8 hardware acceleration.

\textbf{Sliding Tile Attention (STA).} 
Standard attention operations on $N=L$ tokens with feature dimension $d$ incur a computational complexity of $\mathcal{O}(N^2d)$, creating a significant bottleneck for high-resolution video generation. STA addresses this challenge by partitioning the 3D token space into $M$ non-overlapping tiles $\{\mathcal{T}_u\}$ of dimensions $(T_t, T_h, T_w)$. 

The key innovation of STA is its locality-based attention mechanism: each query tile $u$ attends exclusively to key tiles $v$ within a local neighborhood $\mathcal{W}(u)$, defined by the distance constraint:
\begin{equation}
\mathcal{W}(u) = \left\{v : \|c_u-c_v\|_\infty \leq \left(\frac{W_t}{2T_t}, \frac{W_h}{2T_h}, \frac{W_w}{2T_w}\right)\right\},
\end{equation}
where $(W_t, W_h, W_w)$ denote the window dimensions measured in tile units, and $c_u$, $c_v$ are the centers of tiles $u$ and $v$. This constraint effectively replaces full attention with a tile-wise masked attention:
\begin{equation}
{P}_{q,k}=
\begin{cases}
\text{Softmax}(Q_qK_k^\top/\sqrt{d}), & \text{if token } k \text{ is in a tile } \mathcal{T}_v \text{ where } v\in\mathcal{W}(u), \\
-\infty, & \text{otherwise.}
\end{cases}
\label{eq:sta_mask}
\end{equation}
where $Q$ and $K$ are queries and keys.
STA generates $M\!\times\!|\mathcal{W}(u)|$ dense attention blocks that are compatible with optimized implementations such as FlashAttention\cite{dao2022flashattention1}. This design provides substantial speedup by replacing the irregular sparse patterns of token-wise sliding window attention with structured tile-based computations that align well with GPU memory hierarchies. As demonstrated in \cite{zhang2025fastvideo}, this approach can accelerate attention by 2.8–17$\times$ over FlashAttention-2\cite{dao2023flashattention2} and 1.6–10$\times$ over FlashAttention-3\cite{shah2024flashattention3fastaccurateattention} for video generation tasks.

\subsection{Joint Tile-wise FP8 Sparse Attention}
\label{sec:joint-tile-wise-quantization-and-sparse-attention}

\begin{wrapfigure}{r}{0.5\textwidth}
  \vspace{-2em} %
  \centering
  \includegraphics[width=0.48\textwidth,page=3]{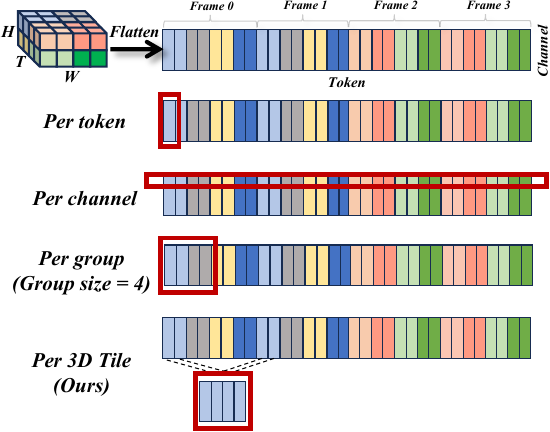}
  \vskip -0.05in
  \caption{Quantization granularities: per-token, per-channel, per-group, and our per 3D-tile, which aligns with hardware compute patterns.}
  \label{fig:quant-comparison}
  \vspace{-1em} %
\end{wrapfigure}

Building upon FP8 quantization and tiled attention techniques, we introduce \texttt{\ourname{}}, a \emph{Joint Tile-wise FP8 Quantization and Sparse Attention} mechanism that synergistically optimizes computational efficiency and accuracy in video DiTs.

Our tile-wise granularity approach (Figure~\ref{fig:quant-comparison}, last row) is motivated by three primary considerations. First, it offers an optimal accuracy-efficiency trade-off compared to conventional methods (per-token, per-channel, per-group; Figure~\ref{fig:quant-comparison}, first three rows) that often fail to align with underlying hardware architectures. While per-group quantization provides a reasonable balance, it frequently overlooks GPU compute tile patterns, thereby reducing hardware utilization efficiency. Second, our approach maintains full compatibility with the STA sparsity design, allowing seamless integration of quantization and sparsity optimizations at matching granularity. Third, our tile-wise design exhibits superior hardware compatibility, aligning precisely with compute tiles in optimized kernels such as FlashAttention, which enables direct translation of theoretical computational savings into practical speedups.

The \texttt{\ourname{}} mechanism processes neural activations through a systematic workflow: (1) organizing query ($Q$) and key ($K$) activations into contiguous 3D tiles aligned with GPU cache layouts for enhanced data locality; (2) quantizing each tile to FP8 precision with a locally optimized scale factor; (3) enforcing tile-granularity sparse attention patterns, leveraging spatial locality and low-bit arithmetic; and (4) dequantizing the aggregated attention output to higher precision (BF16/FP16).

\textbf{Tile-wise FP8 quantization for $Q$ and $K$.}
The matrices $Q, K\!\in\!\mathbb{R}^{L \times d}$ are partitioned along the sequence dimension $L$ into non-overlapping tiles \{$\mathcal{T}_u$\} of dimensions $(T_t, T_h, T_w)$. For each tile $\mathcal{T}_u$, we compute separate scaling factors $s_u^Q$ and $s_u^K$ to map their values optimally to the FP8 representable range via
\begin{equation}
s_u^Q\!=\!\max_{(i,j) \in \mathcal{T}_u} |Q_{i,j}| / M_{\text{FP8\_max}}, \quad s_u^K\!=\!\max_{(i,j) \in \mathcal{T}_u} |K_{i,j}| / M_{\text{FP8\_max}},
\end{equation}
where $M_{\text{FP8\_max}}$ is the maximum representable magnitude in FP8 and bounded by specific format. Then each element is independently quantized:
\begin{equation}
    \hat{Q}_{i,j}\!=\!\text{FP8}(Q_{i,j}; s_u^Q),  \quad
\hat{K}_{i,j}\!=\!\text{FP8}(K_{i,j}; s_u^K).
\end{equation}
This per-tile scaling strategy minimizes quantization error for both $Q$ and $K$ independently, preserving attention dynamics more effectively than global scaling approaches. When combined with the STA formulation, the attention weights $P$ are computed using the quantized $\hat{Q}$ and $\hat{K}$, following Eq.~\ref{eq:sta_mask}. This formulation enforces a regular, block-sparse pattern (Figure~\ref{fig:method-overview-fp8}) that efficiently maps to modern GPU compute architectures.

\textbf{Channel-wise FP8 quantization of $V$ and tensor-wise FP8 quantization of $P$}. 
For the value matrix $V\!\in\!\mathbb{R}^{L \times d}$, a channel-wise FP8 quantization is employed. For each channel, we compute the scaling factor $s^V_j$ to map the values optimally to the FP8 representable range:
$s^V_{j}\!=\!\max_{i \in L} |V_{i,j}| / M_{\text{FP8\_max}}$.
Each element in $V_{j}$ is subsequently quantized as $\widehat{V}_{j}\!=\!\text{FP8}(V_{j};s^V_j)$. 
We observe that keeping the fine granularity for $V$ is critical for the performance. Following SageAttention2~\cite{zhang2024sageattention2}, we use a fixed scalar $\frac{1}{448}$ to quantize $P$, obtaining $\widehat{P}$ in FP8.

\textbf{Aggregation and dequantization.}
The attention output is computed as $\widehat{X} = \widehat{P}\widehat{V}$.
This low-precision output is then dequantized to higher precision (BF16/FP16) to maintain computational stability in subsequent layers.

By quantizing $Q$ and $K$ tiles independently while using channel-wise quantization for $V$, the \texttt{\ourname{}} mechanism captures fine-grained statistical properties of each attention component while maintaining optimal alignment with GPU memory hierarchies. Furthermore, the tile-constrained sparse attention pattern creates $M \times |\mathcal{W}(u)|$ dense, structured attention blocks that avoid the inefficiencies of unstructured sparsity patterns. This approach enables direct wall-clock speedups during execution by maximizing hardware utilization and minimizing memory access overhead, as demonstrated in our experimental results.

\subsection{Denoising Step-aware Quantization and Sparsity Strategy}
\label{sec:denoising-step-aware}

\begin{figure}
    \centering
    \begin{subfigure}[b]{0.32\textwidth}
      \includegraphics[width=\textwidth]{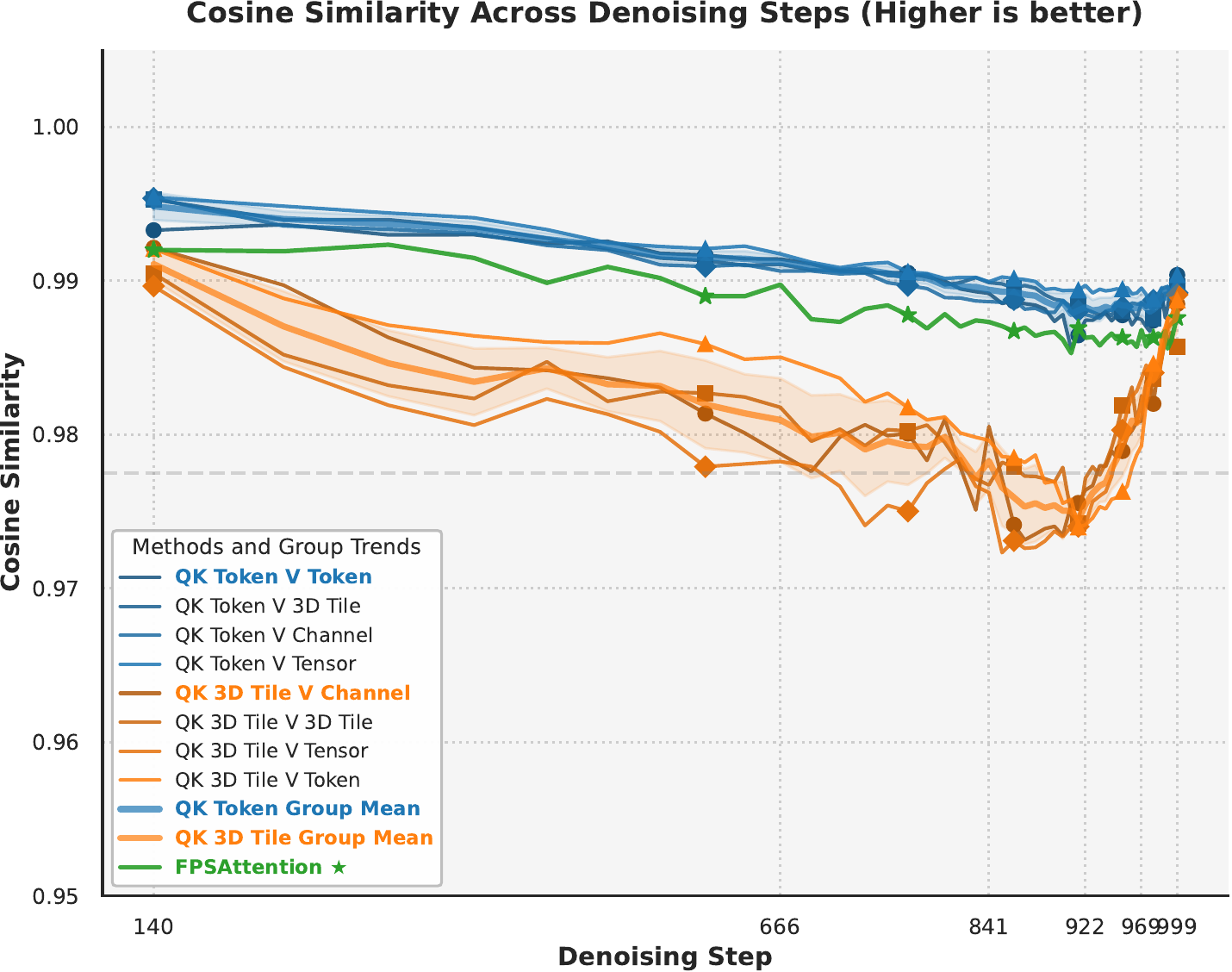}
      \caption{Cosine similarity.}
      \label{fig:cosine_similarity}
    \end{subfigure}
    \hfill
    \begin{subfigure}[b]{0.32\textwidth}
      \includegraphics[width=\textwidth]{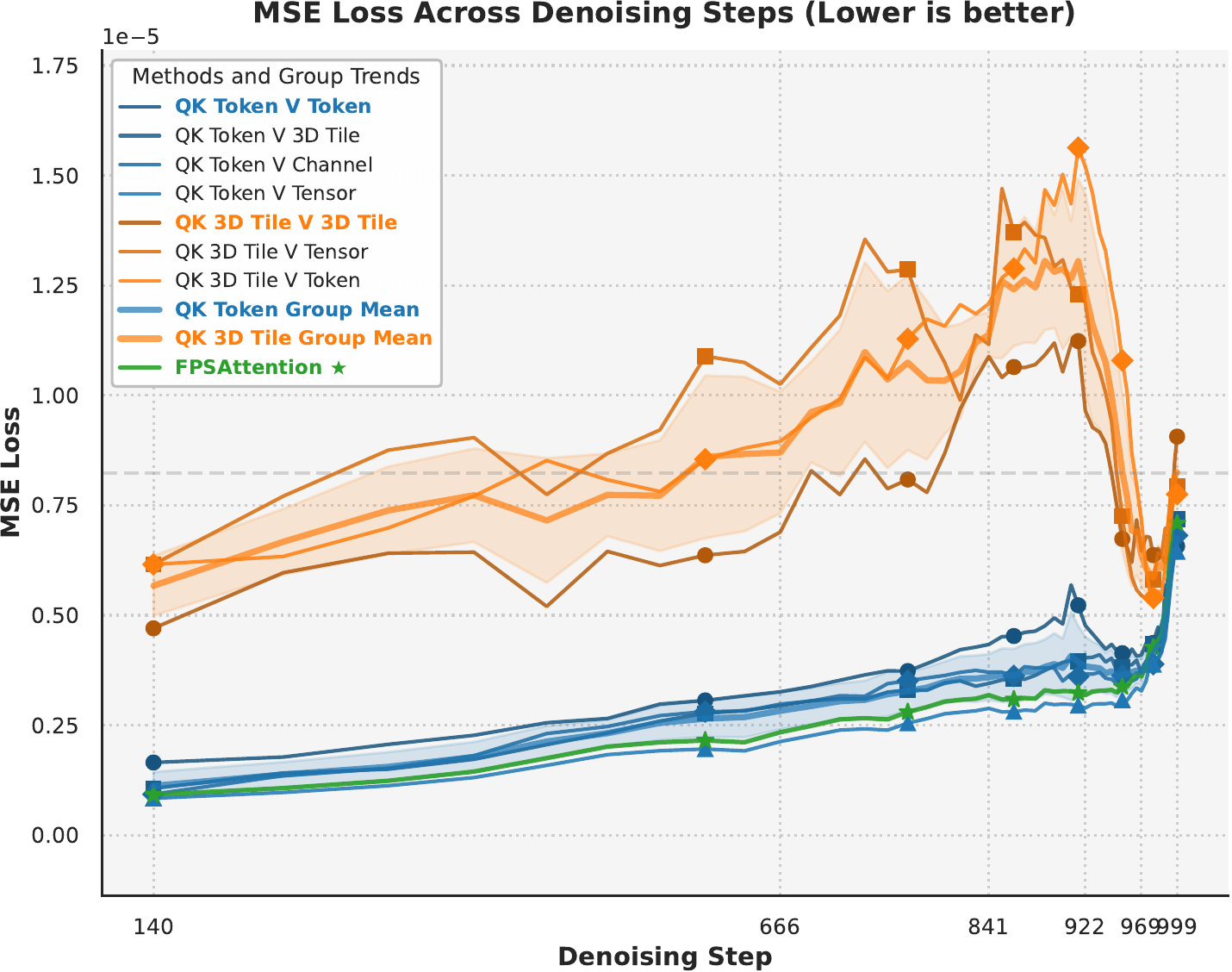}
      \caption{MSE loss.}
      \label{fig:mse_loss}
    \end{subfigure}
    \hfill
    \begin{subfigure}[b]{0.32\textwidth}
      \includegraphics[width=\textwidth]{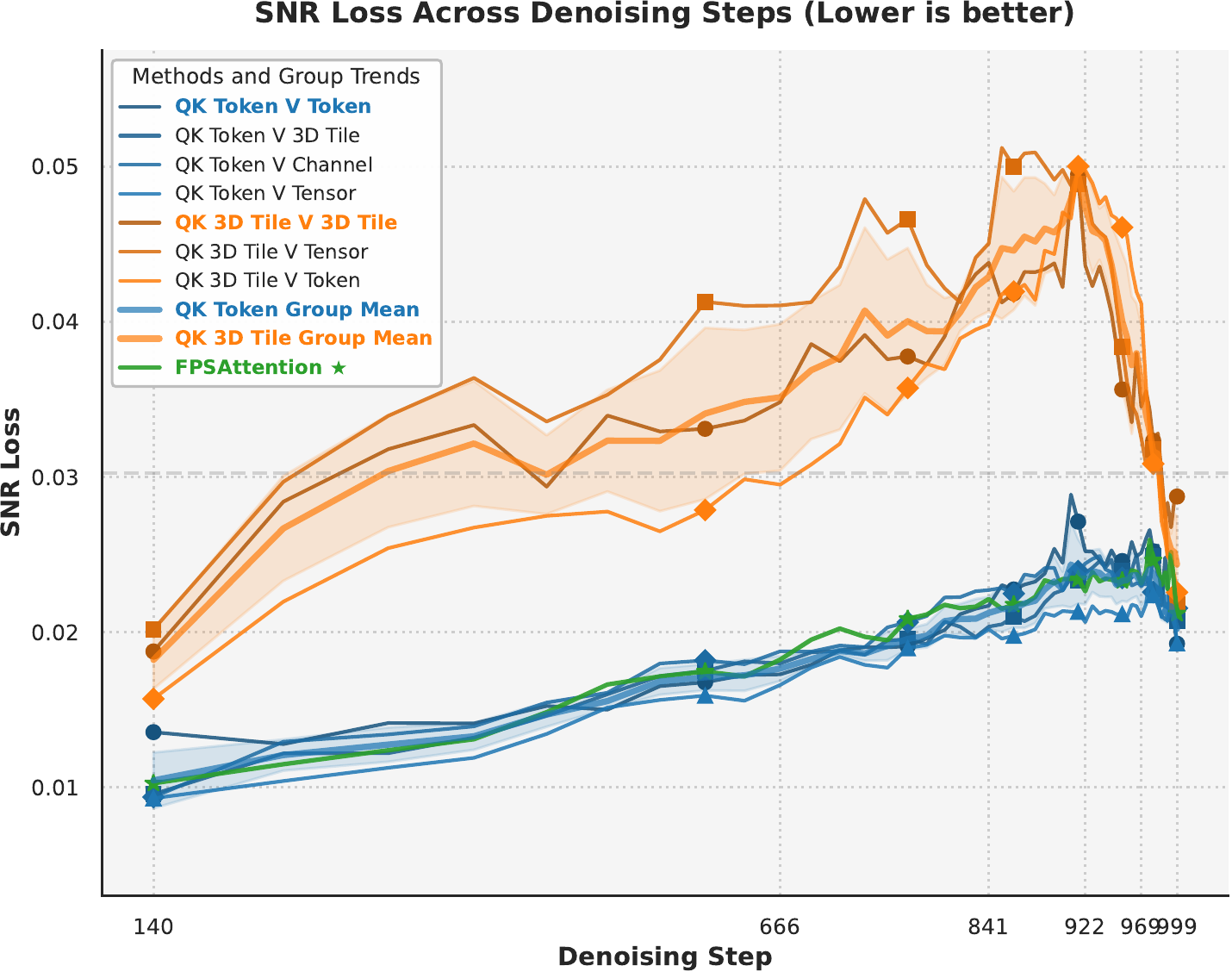}
      \caption{SNR loss.}
      \label{fig:snr_loss}
    \end{subfigure}
    \caption{
    Joint quantization and sparsity error patterns across denoising steps. \textcolor{RoyalBlue}{Blue}: token-level granularity; \textcolor{orange}{orange}: our 3D tile-wise granularity with sparse attention. Key insight: early/late steps tolerate coarser quantization and higher sparsity, while intermediate steps require finer granularity and denser attention. Our \texttt{\ourname{}} (\textcolor{OliveGreen}{green}) closely approximates highest-granularity methods, validating our adaptive scheduling strategy. All measurements from inference are with identical prompts. Performance gaps primarily stem from FP8 quantization rather than sparsity constraints.
    }
    \label{fig:FPSAttention_adaptive_schedule}
\end{figure}

The mechanism described in Section~\ref{sec:joint-tile-wise-quantization-and-sparse-attention} employs uniform quantization and sparsity across denoising steps. 
However, as illustrated in Fig.~\ref{fig:FPSAttention_adaptive_schedule}, 
video DiTs exhibit varying sensitivity to numerical precision and sparsity levels throughout the diffusion process. Specifically, early and late denoising steps demonstrate greater tolerance to coarser quantization and higher sparsity, whereas intermediate steps demand finer numerical precision and lower sparsity. This also suggests that diffusion models can intrinsically correct the approximation errors of attention, which motivates our training-aware scheme to mitigate the training-inference gap.
Based on these observations, we propose an adaptive, denoising step-aware compression schedule, for both training and inference. We adjust the quantization granularity $g(t)$ and sparsity window size $W(t)$ based on the denoising step $t$.

\textbf{Piecewise schedule for quantization and sparsity.}
For $D$ denoising steps, we partition the process into three regimes using thresholds  $t_1\!=\! \alpha_1 D$ and $t_2\!=\!\alpha_2 D$ ($0\!<\!\alpha_1\!<\!\alpha_2\!<\!1$), each associated with different quantization tile sizes $g(t)$ and sparsity window sizes $W(t)$. Smaller $g(t)$ corresponds to a finer quantization granularity, while larger $W(t)$ indicates a denser attention pattern.

We show quantization and sparsity schedule as following. 
We define a time-dependent hyperparameter vector $S(t) = [g(t), W(t)]$, which is governed by:
\begin{equation}
S(t) = 
\begin{cases}
[g_{\text{coarse}}, W_{\text{sparse}}], & t \leq t_1 \text{ (Early-Denoising Steps)}, \\
[g_{\text{fine}}, W_{\text{dense}}], & t_1 < t \leq t_2 \text{ (Mid-Denoising Steps)}, \\
[g_{\text{intermediate}}, W_{\text{medium\_density}}], & t > t_2 \text{ (Late-Denoising Steps)},
\end{cases}
\label{eq:adaptive_schedule}
\end{equation}
with $g_{\text{coarse}}\!>\!g_{\text{intermediate}}\!>\!g_{\text{fine}}$ and $W_{\text{dense}}\!>\!W_{\text{medium\_density}}\!>\!W_{\text{sparse}}$, ensuring the finest quantization granularity and densest attention patterns during these mid-denoising steps, as illustrated in Figure~\ref{fig:FPSAttention_adaptive_schedule}.

These hyperparameters are selected at inference time to match the model’s varying tolerance to quantization and sparsity across different denoising stages, and then transferred to the training to avoid the prohibitive computational overhead.
During training, this configuration allows the model to adaptively compensate for 
joint quantization-sparsity errors,
leading to satisfiable stability and convergence throughout the training process, as shown in Figure~\ref{fig:loss_comparison}. %

\subsection{Hardware-Optimized Kernel Design}
\label{sec:hardware-optimized-kernel}

Our algorithmic designs are complemented by a hardware-optimized kernel implementation for maximum practical efficiency. The implementation addresses several key aspects: memory access coalescing through structured operations that enable efficient GPU memory loads/stores with tiling support; maximized parallelism via tile-wise operations that process independent tiles concurrently; exploitation of 
dedicated acceleration units such as Tensor Cores on NVIDIA Hopper/Ada architectures 
for mixed-precision and FP8 computations; and operation fusion that combines multiple logical steps (attention, sparsity, dequantization) into single triton kernels, significantly reducing overhead and memory traffic while maintaining high tensor core utilization and computational intensity.

\section{Experiments}

\noindent\textbf{Implementation details.} We implemented our proposed framework on the Wan2.1\cite{wan2025wanopenadvancedlargescale} architecture (1.3B and 14B variants), preserving the original model structure while introducing \our, joint FP8 quantization and structured sparsity. The quantization schemes and sparsity patterns were applied across attention mechanisms using score mod and mask mod functions via FlexAttention \cite{dong2024flex} . Fused kernels were compiled using Triton to accelerate inference on Hopper GPUs. The models were trained on high-quality video data (480p$\times$16fps$\times$5s).
We trained \our on Wan2.1-14B using 64 nodes with 8 H20 GPUs for 7 days. For detailed hardware specifications, training procedures, dataset preparation, evaluation protocols, and baseline comparisons, please refer to Appendix. 

\noindent\textbf{Evaluation protocol.}
We evaluate the our method on the public video dataset, VBench \cite{kong2024hunyuanvideo}. We following the common practice \cite{zhao2024real, li2024distrifusion} to sample 5 videos per evaluation prompts defined in the VBench dataset, and assess the video generation quality across 16 VBench dimensions. We also report Peak Signal-to-Noise Ratio (PSNR) \cite{hore2010image}, Structural Similarity Index (SSIM)\cite{ssim}, and Learned Perceptual Image Patch Similarity (LPIPS) \cite{zhang2018unreasonable} metrics. 

\begin{table*}[h]
  \caption{Quality and efficiency benchmarking results. $^\dagger$ We reproduce the results of the baseline methods from the original papers. Note that VBench results here may differ from official results due to the randomness in generated samples and prompt extensions. The quality and efficiency evaluation is based on 480p videos. Specifically, $^\ddagger$ indicates the speedups via 720p with longer sequence length.}
  \label{tab:benchmarking}
  \centering
  \resizebox{\textwidth}{!}{%
  \renewcommand{\arraystretch}{1.2}
  \begin{tabular}{l|ccccc|cccc}
  \toprule
  \multirow{2}{*}{\textbf{Method}} & \multicolumn{5}{c|}{\textbf{Quality}} & \multicolumn{4}{c}{\textbf{Efficiency}} \\
  \cmidrule(lr){2-6} \cmidrule(lr){7-10}
   & PSNR $\uparrow$ & SSIM $\uparrow$ & LPIPS $\downarrow$ & ImageQual $\uparrow$ & SubConsist $\uparrow$ & FLOPS $\downarrow$ & Latency $\downarrow$ & Speedup $\uparrow$ & Speedup$^{\ddagger}$ $\uparrow$ \\
  \midrule
  Wan2.1-1.3B$^\dagger$ & - & - & - & 0.6708 & 0.9536 & 77.52 PFLOPS & 271s & 1.00x & - \\
  \cmidrule(lr){1-10}
  SageAttention & 20.18990 & 0.78241 & 0.18811 & 0.6699 & 0.9453 & 37.61 PFLOPS & 141s & 1.91x & - \\
   SpargeAtten & 17.72979 & 0.72628 & 0.26183 & 0.6541 & 0.8982 & 43.15 PFLOPS & 205s & 1.32x & - \\
   SparseVideoGen & 19.51276 & 0.78891 & 0.20513 & 0.6729 & 0.9292 & 30.67 PFLOPS & 152s & 1.78x & - \\
   STA & 18.78546 & 0.76335 & 0.23187 & 0.6626 & 0.8992 & 31.78 PFLOPS & 143s & 1.89x & - \\
  \rowcolor{lightblue}
   Ours Quant & \textcolor{darkgreen}{20.99712} & \textcolor{darkgreen}{0.79820} & \textcolor{darkgreen}{0.15114} & 0.6798 & 0.9458 & 32.01 PFLOPS & 144s & \textcolor{darkgreen}{1.88x} & - \\
  \rowcolor{lightblue}
   Ours Quant + Sparse & \textcolor{darkgreen}{21.35417} & \textcolor{darkgreen}{0.80835} & \textcolor{darkgreen}{0.15398} & 0.7103 & 0.9338 & 32.01 PFLOPS & 110s & \textcolor{darkgreen}{2.45x} & - \\
  \midrule
  Wan2.1-14B$^\dagger$ & - & - & - & 0.6715 & 0.9528 & 637.52 PFLOPS & 1301s & 1.00x & 1.00x \\
  \cmidrule(lr){1-10}
  SageAttention & 24.33985 & 0.82283 & 0.15607 & 0.6724 & 0.9530 & 301.98 PFLOPS & 646s & 2.01x & 1.94x \\
   SpargeAtten & 21.38291 & 0.81452 & 0.21723 & 0.6350 & 0.9173 & 339.30 PFLOPS & 734s & 1.77x & 2.12x \\
   SparseVideoGen & 23.52881 & 0.80113 & 0.17032 & 0.6868 & 0.9489 & 259.79 PFLOPS & 613s & 2.12x & 3.13x \\
   STA & 22.65635 & 0.82024 & 0.19283 & 0.6577 & 0.9530 & 264.34 PFLOPS & 548s & 2.37x & 3.60x \\
  \rowcolor{lightblue}
   Ours Quant + Sparse & \textcolor{darkgreen}{25.74353} & \textcolor{darkgreen}{0.83171} & \textcolor{darkgreen}{0.07610} & 0.7103 & 0.9435 & 273.01 PFLOPS & 423s & \textcolor{darkgreen}{3.07x} & \textcolor{darkgreen}{4.96x} \\
  \bottomrule
  \end{tabular}%
  }
  \end{table*}

\subsection{Main Results}
\textbf{Quality evaluation.}
We compare \our with several state-of-the-art optimization methods, as shown in Table~\ref{tab:benchmarking}. Our baselines include sparsity-based approaches (SparseVideoGen \cite{xi2025sparse} and STA), quantization methods (SageAttention \cite{zhang2024sageattention}), and hybrid approaches (SpargeAtten \cite{zhang2025spargeattn}, a training-free method that jointly applies attention sparsification and activation quantization).
As demonstrated in Table~\ref{tab:benchmarking}, \our achieves superior performance across all quality metrics. Particularly notable is the average PSNR of 25.74353 on the Wan2.1-14B model, significantly outperforming all baseline methods. This objective metric confirms \our's ability to generate videos with exceptional fidelity to reference images.
Furthermore, \our maintains excellent performance on perceptual metrics, with high Video Quality (0.7103) and strong spatial-temporal consistency (0.9435) on the VBench evaluation.
Interestingly, after joint training, \our exhibits a slight increase in VBench scores, while \cite{yuan2025native,zhang2025fastvideo} also demonstrate performance improvements when trained with structured sparsity—potentially driven by the inductive bias of locality.
These results validate that our joint sparsity and quantization approach preserves visual quality while substantially improving computational efficiency. Visual examples in Figure~\ref{fig:viz-example} further illustrate that \our \textbf{consistently outperforms} baseline methods while maintaining quality comparable to the original model.

\textbf{Test-time efficiency.} 
We evaluate computational efficiency across both 1.3B and 14B parameter variants of the Wan2.1 model, as reported in Table~\ref{tab:benchmarking}. For the 1.3B parameter model, \our achieves up to 2.45$\times$ speedup on 480p videos while maintaining superior quality metrics. More impressively, when tested on the larger 14B parameter model with 720p videos (indicated by $^{\ddagger}$ in the table), \our achieves a substantial \textbf{4.96$\times$ end-to-end speedups} compared to the baseline.

\begin{figure}[t]
    \centering
    \captionsetup[subfloat]{labelsep=none,format=plain,labelformat=empty}
        \subfloat[Baseline, Wan2.1 1.3B, 1$\times$ E2E speedup]
        {
            \subfloat
            {
                \animategraphics[scale=0.22]{16}{video/boat/baseline/boat-}{0}{19}
            }
            \subfloat
            {
                \animategraphics[scale=0.22]{16}{video/tranquil/baseline/tranquil-}{0}{19}
            }
            \subfloat
            {
                \animategraphics[scale=0.22]{16}{video/oven/baseline/oven-}{0}{19}
            }
            \subfloat
            {
                \animategraphics[scale=0.22]{16}{video/underwater/baseline/underwater-}{0}{19}
            } 
            \subfloat
            {
                \animategraphics[scale=0.22]{16}{video/shark/baseline/shark-}{0}{19}
            }
        }
        \vfill
        \subfloat[Our FPSAttention, 4.96$\times$ E2E speedup]
        {
            \subfloat
            {
                \animategraphics[scale=0.22]{16}{video/boat/ours/boat-}{0}{19}
            }
            \subfloat
            {
                \animategraphics[scale=0.22]{16}{video/tranquil/ours/tranquil-}{0}{19}
            }
            \subfloat
            {
                \animategraphics[scale=0.22]{16}{video/oven/ours/oven-}{0}{19}
            }
            \subfloat
            {
                \animategraphics[scale=0.22]{16}{video/underwater/ours/underwater-}{0}{19}
            }
            \subfloat
            {
                \animategraphics[scale=0.22]{16}{video/shark/ours/shark-}{0}{19}
            }
        }
    \caption{
        Examples of generated videos by \our and the Wan2.1 1.3B baseline. We showcase from five different aspects. \our achieves 4.96$\times$ E2E speedup, while maintaining lossless visual quality. Please click the image to play the video clip via Acrobat Reader.
    }
    \label{fig:viz-example}
    \end{figure}

\begin{table}[b]
  \centering
  \begin{minipage}{0.43\textwidth}
  \centering
  \caption{Effect of different tile sizes on model performance on Wan2.1 1.3B. We evaluate various tile size combinations for temporal (t), height (h), and width (w) dimensions.}
  \label{tab:quant_granularity}
  \resizebox{\linewidth}{!}{%
  \begin{tabular}{ccc|ccc}
  \toprule
  \multicolumn{3}{c|}{\textbf{Tile Size}} & \multicolumn{3}{c}{\textbf{Metrics}} \\
  t & h & w & PSNR$\uparrow$ & SSIM$\uparrow$ & LPIPS$\downarrow$ \\
  \midrule
  3 & 4 & 4 & 19.87452 & 0.77634 & 0.16521 \\
  6 & 8 & 8 & 20.12358 & 0.78147 & 0.15982 \\
  12 & 16 & 16 & 20.45621 & 0.78932 & 0.15743 \\
  \rowcolor{blue!10}24 & 32 & 32 & 20.99712 & 0.79820 & 0.15114 \\
  \bottomrule
  \end{tabular}%
  }
  \end{minipage}%
  \hfill
  \begin{minipage}{0.53\textwidth}
  \centering
  \caption{Impact of sparsity window dimensions on model performance and computational efficiency. We evaluate various combinations of temporal (t), height (h), and width (w) window sizes and measure inference kernel speedup.}
  \label{tab:sparsity_window}
  \resizebox{\linewidth}{!}{%
  \begin{tabular}{ccc|ccc|c}
  \toprule
  \multicolumn{3}{c|}{\textbf{Window Size}} & \multicolumn{4}{c}{\textbf{Metrics}} \\
  t & h & w & PSNR$\uparrow$ & SSIM$\uparrow$ & LPIPS$\downarrow$ & Speedup$\uparrow$ \\
  \midrule
  3 & 3 & 1 & 19.23465 & 0.76128 & 0.17251 & 3.24x \\
  5 & 6 & 10 & 19.94731 & 0.77926 & 0.16327 & 3.07x \\
  6 & 6 & 6 & 20.12482 & 0.78341 & 0.16014 & 1.69x \\
  \rowcolor{blue!10} 6 & 6 & 1 & 20.45621 & 0.78932 & 0.15743 & 5.16x \\
  \bottomrule
  \end{tabular}%
  }
  \end{minipage}
  \end{table}

\subsection{Ablation Study}
In this section, we conduct a comprehensive ablation study to analyze the effects of key components in \our . 
We investigate the different tile size selection for quantization and sparsity, and the relations to the hardware awareness.
Additionally, we investigate how sparsity window dimensions affect model performance and computational efficiency.
We also study the training stability to verify whether our \our training %
dynamics
highly aligns the baseline trend.

  \textbf{Effect of tile size for quantization and sparsity}.
  We investigate the impact of joint quantization and sparsity granularity by varying tile sizes along temporal (t), height (h), and width (w) dimensions, with results presented in Table~\ref{tab:quant_granularity}. Our experiments demonstrate that the largest tile configuration (24,32,32) achieves optimal performance with PSNR 20.99712, SSIM 0.79820, and LPIPS 0.15114. However, we observe minimal performance differences between configurations (6,8,8), (12,16,16), and (24,32,32). Based on these findings, we implement a scheduled approach, using different tile sizes for early, mid, and late denoising steps, respectively. We emphasize the hardware-friendly configuration of (6,8,8) yields the best throughput as it aligns with flash attention block size design and is optimized for the Hopper architecture.
  Notably, unbalanced tile sizes, such as (3,4,4), lead to significant performance degradation, this configuration is also misalign with flash attention block size leads to inefficiency.

\textbf{Effect of sparsity window dimensions}.
The relationship between sparsity window dimensions, inference speed, and model performance is analyzed in Table~\ref{tab:sparsity_window}. Our results demonstrate that a configuration (6,6,1) using temporal window of 6, height window of 6, and width window of 1 provides an optimal balance, achieving a substantial 5.16$\times$ kernel speed-up while maintaining high visual quality. %
This suggests that while reducing the temporal and spatial window sizes improves efficiency, there exists a threshold beyond which visual quality deteriorates significantly.

\begin{figure}[t]
    \centering
    \includegraphics[width=0.98\linewidth]{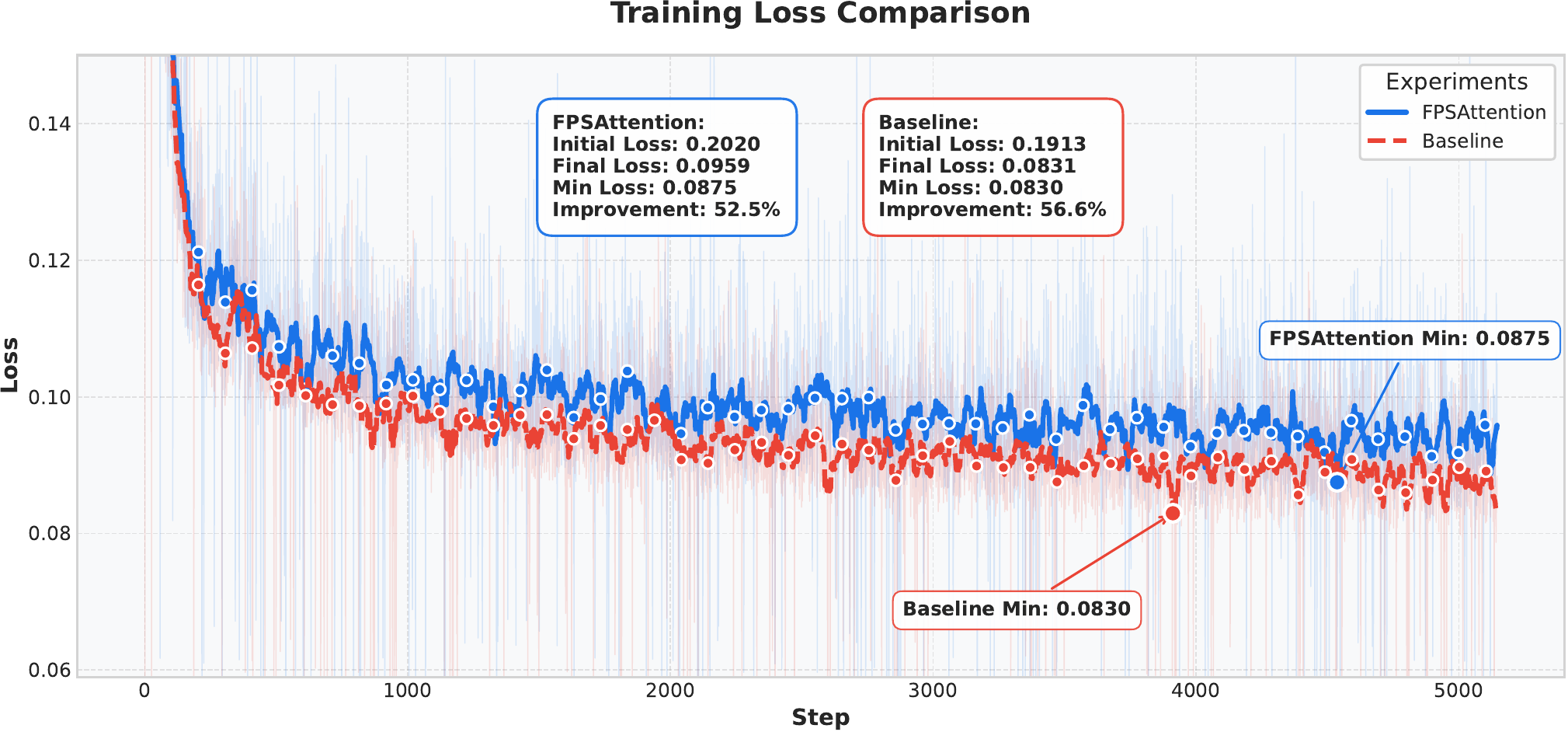}
    \vskip -0.1in
    \caption{Training loss comparison between baseline Wan2.1 1.3B, and our joint FP8 quantization with structured sparsity. The \our shows slightly higher loss initially but stabilizes with comparable final performance.}
    \label{fig:loss_comparison}
    \vspace{-1em}
  \end{figure}

\textbf{Training stability.} Joint FP8 quantization and structured sparsity initially increases training loss by ~15\% compared to full-precision Wan2.1 baseline (Figure~\ref{fig:loss_comparison}). We mitigate these challenges through adaptive learning rate scheduling and gradient accumulation techniques. After 2,000 steps, loss convergence trajectories become nearly identical (<2\% difference), confirming that our 
FP8 sparse attention
preserves critical information pathways despite bitwidth and sparsity constraints.

\section{Conclusion and Future Work}
\label{others}

In this paper, we have introduced \our, a module jointly optimizing FP8 quantization and structured sparsity for video diffusion models. Through unified 3D tile-wise granularity, denoising step-aware adaptation, and hardware-friendly kernel implementation, our approach achieves up to 7.09$\times$ kernel speedup and 4.96$\times$ end-to-end acceleration without compromising generation quality. The tile-aligned approach ensures quantization and sparsity work synergistically, while step-aware scheduling adapts compression hyperparameters to varying sensitivity across diffusion phases.
Despite these results, our approach has limitations: it performs best on FP8-supporting hardware, requires additional training resources, and introduces certain hyperparameters. We currently validate \our on Wan2.1 due to resource constraints. Future work will focus on broadening the applicability by generalizing beyond specific architectures (e.g. Hunyuan~\cite{kong2024hunyuanvideo} based on MMDiT~\cite{esser2024scaling}). We also aim to refine training resource requirements and hyperparameter management, and extend these co-design principles beyond attention mechanisms to other model components.
Furthermore, \our is orthogonal to step distillation techniques \cite{salimans2022progressive, yin2024one}. By incorporating step distillation, additional acceleration can be achieved.

\section{Contributions}
\label{sec:contributions}

This work represents a collaborative effort across multiple institutions and researchers. The following details the specific contributions and affiliations of each author:

\textbf{Author Contributions:}
\begin{itemize}
    \item \textbf{Co-first authors (*):} Akide Liu and Zeyu Zhang contributed equally to this work, with primary responsibility for method design, theoretical analysis, and experimental validation.
    \item \textbf{Co-second authors (\dag):} Zhexin Li and Xuehai Bai contributed equally as co-second authors, focusing on implementation details, algorithm optimization, and evaluation framework development.
    \item \textbf{Project leads (\ddag):} Jiasheng Tang and Bohan Zhuang served as the project leads, providing overall direction, coordination, and strategic guidance throughout the research.
\end{itemize}

\textbf{Internship Affiliations:}
\begin{itemize}
    \item \textbf{Intern at DAMO Academy:} Akide Liu and Zeyu Zhang conducted their research as interns at DAMO Academy, Alibaba Group.
    \item \textbf{Intern at Zhejiang University:} Xuehai Bai conducted research as an intern at Zhejiang University.
\end{itemize}

All authors contributed to the writing and revision of the manuscript. The experimental work was conducted collaboratively across all participating institutions, with computational resources primarily provided by DAMO Academy's infrastructure.

\textbf{Acknowledgement:}
This work was supported by Damo Academy ( Hupan Laboratory ) through Damo Academy ( Hupan Laboratory ) Research Fellow Program.

\newpage

\appendix

\begin{center}
\huge\textbf{Appendix}
\end{center}

\vspace{1em}

\textbf{Contents}
\par\medskip
\noindent A \quad FPSAttention Algorithm \dotfill \pageref{subsec:fpsattention_algorithm}\\
\hspace{1em} B \quad Additional Implementation Details \dotfill \pageref{subsec:additional_impl}\\
\hspace{1em} C \quad Ablation Study: Challenges of Naive Quantization and Sparsity Combination \dotfill \pageref{subsec:ablation_naive}\\
\hspace{1em} D \quad VBench Full Evaluation Results \dotfill \pageref{subsec:vbench_results}\\
\hspace{1em} E \quad Clarification on VBench Evaluation Metrics \dotfill \pageref{subsec:vbench_clarification}\\
\hspace{1em} F \quad Training Hyperparameters \dotfill \pageref{subsec:hyperparameters}\\
\noindent G \quad Limitations \dotfill \pageref{sec:limitations}\\
\noindent H \quad Visualization \dotfill \pageref{sec:visualization}

\newpage

\section{FPSAttention Algorithm}
\label{subsec:fpsattention_algorithm}

Algorithm~\ref{alg:fpsattention_main} presents the core computational workflow of our FPSAttention method, which implements joint tile-wise FP8 quantization with structured sparse attention and denoising step-aware adaptation. The algorithm follows the methodology described in the main paper, incorporating tile-wise quantization for queries and keys, channel-wise quantization for values, tensor-wise quantization for attention weights, and dynamic adaptation based on denoising timesteps.

\begin{algorithm}[H]
    \caption{FPSAttention: Joint Tile-wise FP8 Quantization and Sparse Attention}
    \label{alg:fpsattention_main}
    \begin{algorithmic}[1]
        \Require Input tensors $Q, K, V \in \mathbb{R}^{L \times d}$, denoising step $t$, diffusion steps $D$
        \Require Transition points $\alpha_1, \alpha_2$, quantization granularities $\{g_{\text{coarse}}, g_{\text{fine}}, g_{\text{intermediate}}\}$
        \Require Window sizes $\{W_{\text{sparse}}, W_{\text{dense}}, W_{\text{medium\_density}}\}$, tile scheme $\mathcal{T}$
        \Ensure Output tensor $X \in \mathbb{R}^{L \times d}$ (BF16/FP16)
        
        \State \Comment{1. Denoising Step-aware Parameter Selection}
        \State $t_1 \gets \alpha_1 \cdot D$, $t_2 \gets \alpha_2 \cdot D$
        \If{$t \leq t_1$}
            \State $g(t) \gets g_{\text{coarse}}$, $W(t) \gets W_{\text{sparse}}$ \Comment{Early steps}
        \ElsIf{$t_1 < t \leq t_2$}
            \State $g(t) \gets g_{\text{fine}}$, $W(t) \gets W_{\text{dense}}$ \Comment{Mid steps}
        \Else
            \State $g(t) \gets g_{\text{intermediate}}$, $W(t) \gets W_{\text{medium\_density}}$ \Comment{Late steps}
        \EndIf
        
        \State \Comment{2. Tile-wise FP8 Quantization for Q and K}
        \State Partition $Q, K$ into tiles $\{\mathcal{T}_u\}$ with granularity $g(t)$
        \For{each tile $\mathcal{T}_u$}
            \State $s_u^Q \gets \max_{(i,j) \in \mathcal{T}_u} |Q_{i,j}| / M_{\text{FP8\_max}}$
            \State $s_u^K \gets \max_{(i,j) \in \mathcal{T}_u} |K_{i,j}| / M_{\text{FP8\_max}}$
            \State $\hat{Q}_{i,j} \gets \text{FP8}(Q_{i,j}; s_u^Q)$ for $(i,j) \in \mathcal{T}_u$
            \State $\hat{K}_{i,j} \gets \text{FP8}(K_{i,j}; s_u^K)$ for $(i,j) \in \mathcal{T}_u$
        \EndFor
        
        \State \Comment{3. Channel-wise FP8 Quantization for V}
        \For{each channel $j \in \{1, \ldots, d\}$}
            \State $s^V_j \gets \max_{i \in L} |V_{i,j}| / M_{\text{FP8\_max}}$
            \State $\hat{V}_{i,j} \gets \text{FP8}(V_{i,j}; s^V_j)$ for all $i$
        \EndFor
        
        \State \Comment{4. Structured Sparse Attention Computation via FlexAttention}
        \State Define neighborhood $\mathcal{W}(u)$ based on window size $W(t)$:
        \State $\mathcal{W}(u) = \{v : \|c_u-c_v\|_\infty \leq (W_t/(2T_t), W_h/(2T_h), W_w/(2T_w))\}$
        
        \State \Comment{Configure FlexAttention mask and score modification functions}
        \State Define \texttt{mask\_mod}$(b, h, q, k) = $ \textbf{True} if tile($q$) $\in \mathcal{W}($tile($k$)$)$, \textbf{False} otherwise
        \State Define \texttt{score\_mod}$(S, b, h, q, k) = S$ \Comment{Identity for quantized inputs}
        
        \State \Comment{Execute FlexAttention with quantized inputs and custom modifications}
        \State $\hat{X} \gets$ \texttt{FlexAttention}$(\hat{Q}, \hat{K}, \hat{V}$, \texttt{score\_mod}, \texttt{mask\_mod}$)$
        
        \State \Comment{5. Dequantize Output to Target Precision}
        \State $X \gets \text{Dequantize}(\hat{X})$ \Comment{Dequantize to BF16/FP16}
        
        \Return $X$
    \end{algorithmic}
\end{algorithm}

\subsection*{Key Algorithmic Components}
The algorithm implements the four core innovations described in the main paper:

\begin{itemize}
    \item \textbf{Denoising Step-aware Adaptation}: Lines 2-8 implement the adaptive scheduling strategy from Equation 6 in the main paper, dynamically adjusting quantization granularity $g(t)$ and sparsity window size $W(t)$ based on the current denoising step $t$.
    
    \item \textbf{Tile-wise FP8 Quantization for Q and K}: Lines 10-15 partition queries and keys into 3D tiles with step-dependent granularity and compute per-tile scaling factors $s_u^Q$ and $s_u^K$ to minimize quantization error within each tile.
    
    \item \textbf{Channel-wise FP8 Quantization for V}: Lines 17-20 apply channel-wise quantization to the value matrix, preserving fine-grained channel information that is critical for generation quality.
    
    \item \textbf{FlexAttention-based Sparse Attention}: Lines 22-26 implement structured sparse attention using FlexAttention's \texttt{mask\_mod} and \texttt{score\_mod} interfaces, enabling hardware-optimized execution with tile-wise sparsity patterns that generate exactly $M \times |\mathcal{W}(u)|$ dense attention blocks.
    
    \item \textbf{Output Dequantization}: Line 28 dequantizes the FlexAttention output to the target precision (BF16/FP16) to maintain compatibility with the downstream network components.
\end{itemize}

This implementation ensures full compatibility with the theoretical framework while enabling practical hardware acceleration through structured computation patterns and optimal memory access patterns.

\section{Additional Implementation Details}
\label{subsec:additional_impl}

\textbf{Models.} 
We implement and evaluate FPSAttention on the Wan architecture~\cite{wang2025wan}, leveraging both 1.3B and 13B parameter variants to demonstrate scalability. The Wan models feature a DiT backbone with cross-attention for text conditioning and temporal attention for inter-frame modeling. Our implementation maintains architectural fidelity while seamlessly integrating FP8 quantization across attention and feed-forward components. The joint quantization and sparsity mechanisms are realized through FlexAttention's score and mask modification interfaces, with the resulting fused kernels compiled via Triton for optimal execution on Hopper architectures.

\textbf{Hardware.} 
Experiments utilize a distributed computing cluster with high-performance GPU nodes, each containing 192 CPU cores, 960GB system memory, and 8×NVIDIA H20 GPUs (96GB each). InfiniBand interconnects ensure high-bandwidth inter-node communication for distributed training. Training scales from 16 nodes (1.3B model) to 64 nodes (13B model), requiring approximately 7 days per configuration to achieve convergence.

\textbf{Dataset.} 
Training employs a curated high-quality video dataset processed through a comprehensive filtering pipeline. The preprocessing workflow includes automated subtitle removal, black-border cropping, and monochrome video exclusion, followed by quality-based filtering using established metrics (Q-Align > 3.5, Aesthetic Score > 2.0, optical flow magnitude 0.05–2.0). After deduplication, videos are standardized to 480p resolution, 16fps frame rate, and 5-second duration to optimize the computational efficiency-quality balance across both model scales.

\textbf{Evaluation.}
Performance assessment utilizes the VBench benchmark~\cite{kong2024hunyuanvideo}, following established protocols~\cite{zhao2024real, li2024distrifusion} with 5-video sampling per prompt. Evaluation encompasses 16 comprehensive VBench dimensions covering aesthetic quality, temporal consistency, motion dynamics, and semantic understanding. Additional quantitative metrics include PSNR~\cite{hore2010image}, SSIM, and LPIPS~\cite{zhang2018unreasonable} to provide multi-faceted quality assessment.

\textbf{Baselines.} 
Our comparative analysis includes representative approaches from three categories: (1) sparsity-based methods (SparseVideoGen~\cite{xi2025sparse}, STA), (2) quantization-focused techniques (SageAttention~\cite{zhang2024sageattention}), and (3) joint optimization methods (SpargeAttn~\cite{zhang2025spargeattn}). This selection enables comprehensive evaluation of FPSAttention against both specialized single-optimization approaches and competing joint methods, providing a thorough assessment of our framework's relative performance and efficiency gains.

\section{Ablation Study: Challenges of Naive Quantization and Sparsity Combination}
\label{subsec:ablation_naive}

To validate our core motivation that naive combination of FP8 quantization and sparsity presents significant challenges, we conduct a comprehensive ablation study comparing three key approaches: (1) the baseline full-precision model, (2) a training-free naive combination of quantization and sparsity, and (3) our proposed FPSAttention method with joint optimization. 

Table~\ref{tab:naive_combination_ablation} presents a detailed comparison across all VBench metrics for the Wan 1.3B model. The training-free approach applies standard FP8 quantization and sparse attention patterns without joint optimization or denoising step-aware adaptation. As hypothesized, this naive combination leads to substantial performance degradation across nearly all evaluation metrics.

\begin{table}[h!]
\centering
\caption{Ablation study demonstrating the challenges of naive quantization and sparsity combination. We compare baseline full-precision (Baseline), training-free naive combination (Training-Free), and our joint optimization approach (FPSAttention) on Wan 1.3B across all VBench metrics. The severe degradation in the training-free approach validates the need for holistic joint optimization.}
\label{tab:naive_combination_ablation}
\resizebox{0.8\textwidth}{!}{
\scriptsize
\begin{tabular}{lccc}
\toprule
\textbf{Metric} & \cellcolor{gray!15}\textbf{Baseline} & \cellcolor{gray!15}\textbf{Training-Free} & \cellcolor{gray!15}\textbf{FPSAttention} \\
\midrule
Aesthetic Quality & 0.6105 & 0.2892 & \textbf{0.6240} \\
Appearance Style & 0.7157 & 0.7874 & \textbf{0.7252} \\
Background Consistency & \textbf{0.9503} & 0.9280 & 0.9156 \\
Color & 0.9049 & 0.4836 & \textbf{0.8932} \\
Dynamic Degree & 0.3014 & 0.3750 & \textbf{0.4195} \\
Human Action & 0.7720 & 0.0200 & \textbf{0.7780} \\
Imaging Quality & 0.6708 & 0.6868 & \textbf{0.7103} \\
Motion Smoothness & 0.9527 & 0.9513 & 0.9413 \\
Multiple Objects & 0.6091 & 0.0000 & \textbf{0.6665} \\
Object Class & 0.7710 & 0.0109 & \textbf{0.8185} \\
Overall Consistency & 0.6453 & 0.1206 & \textbf{0.6893} \\
Quality Score & 0.8332 & 0.7473 & \textbf{0.8428} \\
Scene & 0.3030 & 0.0129 & \textbf{0.3870} \\
Semantic Score & 0.6768 & 0.1733 & \textbf{0.7088} \\
Spatial Relationship & 0.7317 & 0.0008 & \textbf{0.7659} \\
Subject Consistency & 0.9457 & 0.8887 & 0.9338 \\
Temporal Flickering & \textbf{0.9844} & 0.9401 & 0.9336 \\
Temporal Style & 0.6382 & 0.1239 & \textbf{0.6558} \\
\midrule
\textbf{Total Score} & 0.8019 & 0.6325 & \textbf{0.8160} \\
\textbf{Performance Drop} & -- & \textcolor{red}{-21.1\%} & \textcolor{blue}{+1.8\%} \\
\bottomrule
\end{tabular}
}
\end{table}

\textbf{Key Findings:}
The results clearly demonstrate the challenges inherent in naive quantization and sparsity combination:

\begin{itemize}
    \item \textbf{Severe Quality Degradation}: The training-free approach achieves only 0.6325 total score compared to the baseline's 0.8019, representing a substantial 21.1\% performance drop.
    
    \item \textbf{Critical Failure Modes}: Several metrics show near-zero performance in the training-free approach, including Human Action (0.02), Multiple Objects (0.0), Object Class (0.011), and Spatial Relationship (0.0008), indicating complete failure in complex semantic understanding tasks.
    
    \item \textbf{Magnified Quantization Errors}: As predicted by our theoretical analysis, sparsity mechanisms amplify quantization errors in high-magnitude attention scores. This is particularly evident in metrics requiring fine-grained semantic understanding, where the interaction between quantization noise and sparse token selection leads to catastrophic information loss.
    
    \item \textbf{Joint Optimization Success}: In contrast, our FPSAttention approach not only avoids the degradation seen in naive combination but actually improves upon the baseline (0.8160 vs 0.8019, +1.8\% improvement), validating the effectiveness of our denoising step-aware joint optimization strategy.
\end{itemize}

This ablation study emphasizes the necessity of our training-aware co-design scheme. 

\section{VBench Full Evaluation Results}
\label{subsec:vbench_results}

Tables~\ref{tab:wan1.3_results} and~\ref{tab:wan13_results} present comprehensive evaluation results of our method compared to various baselines on VBench for Wan 1.3B and Wan 13B models, respectively. These results demonstrate the effectiveness of our joint FP8 quantization and sparsity approach across multiple video quality metrics.

Table~\ref{tab:wan1.3_results} shows performance comparisons across seven methods on the Wan 1.3B model: the baseline (Base), SageAttention (SageAtt)~\cite{zhang2024sageattention}, SpargeAttention (SpargeAtt)~\cite{zhang2025spargeattn}, SparseVideoGen (SparseVG)~\cite{xi2025sparse}, Sliding Tile Attention (STA)~\cite{zhang2025fastvideo}, our quantization-only variant (Ours-Q), and our full joint quantization and sparsity method (Ours-Q+S). The evaluation covers 18 comprehensive metrics including aesthetic quality, motion dynamics, temporal consistency, and semantic understanding. Our full method (Ours-Q+S) achieves the highest total score of 0.8160, demonstrating superior performance compared to methods that apply quantization or sparsity independently.

Table~\ref{tab:wan13_results} presents similar comparisons for the larger 13B model, where our method continues to achieve competitive performance while providing substantial computational savings. The results validate that our approach scales effectively to larger model sizes while maintaining video generation quality across diverse evaluation criteria.

\begin{table}[h!]
\centering
\caption{Performance comparison of different methods on Wan 1.3B across VBench metrics. We compare the baseline (Base), SageAttention (SageAtt), SpargeAttention (SpargeAtt), SparseVideoGen (SparseVG), Sliding Tile Attention (STA), our quantization-only variant (Ours-Q), and our full joint method (Ours-Q+S). Bold values indicate the best performance for each metric.}
\label{tab:wan1.3_results}
\resizebox{\textwidth}{!}{
\scriptsize
\begin{tabular}{lccccccc}
\toprule
\textbf{Metric} & \cellcolor{gray!15}\textbf{Base} & \cellcolor{gray!15}\textbf{SageAtt} & \cellcolor{gray!15}\textbf{SpargeAtt} & \cellcolor{gray!15}\textbf{SparseVG} & \cellcolor{gray!15}\textbf{STA} & \cellcolor{gray!15}\textbf{Ours-Q} & \cellcolor{gray!15}\textbf{Ours-Q+S} \\
\midrule
Aesthetic Quality & 0.6105 & 0.6104 & 0.5668 & 0.563 & 0.5661 & 0.6091 & \textbf{0.624} \\
Appearance Style & 0.7157 & 0.715 & 0.7744 & 0.2253 & 0.7952 & 0.6922 & \textbf{0.7252} \\
Background Consistency & \textbf{0.9503} & 0.95 & 0.9123 & 0.9525 & 0.9284 & 0.9472 & 0.9156 \\
Color & 0.9049 & 0.8866 & 0.8903 & 0.9037 & 0.8974 & \textbf{0.9146} & 0.8932 \\
Dynamic Degree & 0.3014 & 0.307 & 0.3222 & \textbf{0.7139} & 0.5722 & 0.3222 & 0.4195 \\
Human Action & 0.772 & 0.75 & 0.734 & 0.73 & 0.622 & 0.75 & \textbf{0.778} \\
Imaging Quality & 0.6708 & 0.6699 & 0.6541 & 0.6729 & 0.6626 & 0.6798 & \textbf{0.7103} \\
Motion Smoothness & 0.9527 & 0.9527 & 0.9139 & \textbf{0.9726} & 0.9649 & 0.9496 & 0.9413 \\
Multiple Objects & 0.6091 & 0.5837 & 0.4715 & 0.471 & 0.4043 & 0.6011 & \textbf{0.6665} \\
Object Class & 0.7710 & 0.7695 & 0.6859 & 0.6935 & 0.5818 & 0.7851 & \textbf{0.8185} \\
Overall Consistency & 0.6453 & 0.6451 & 0.6492 & \textbf{0.6935} & 0.6236 & 0.6478 & 0.6893 \\
Quality Score & 0.8332 & 0.8337 & 0.8049 & 0.2336 & 0.7994 & 0.8363 & \textbf{0.8428} \\
Scene & 0.3030 & 0.3092 & 0.2192 & 0.1732 & 0.1995 & 0.3200 & \textbf{0.3870} \\
Semantic Score & 0.6768 & 0.6704 & 0.6412 & 0.6342 & 0.6012 & 0.6780 & \textbf{0.7088} \\
Spatial Relationship & 0.7317 & 0.7364 & 0.7217 & 0.6469 & 0.6863 & 0.7438 & \textbf{0.7659} \\
Subject Consistency & 0.9457 & 0.9453 & 0.8982 & 0.9292 & 0.8993 & \textbf{0.9458} & 0.9338 \\
Temporal Flickering & \textbf{0.9844} & 0.9841 & 0.9647 & 0.9883 & 0.9652 & 0.9822 & 0.9336 \\
Temporal Style & 0.6382 & 0.6385 & 0.6247 & 0.2265 & 0.6005 & 0.6475 & \textbf{0.6558} \\
\midrule
\textbf{Total Score} & 0.8019 & 0.8011 & 0.7722 & 0.7827 & 0.7597 & 0.8046 & \textbf{0.8160} \\
\bottomrule
\end{tabular}
}
\end{table}

\section{Clarification on VBench Evaluation Metrics}
\label{subsec:vbench_clarification}

In this study, we observed that some of the baseline methods we reproduced (including some of our own exploratory experiments prior to FPSAttention) might yield VBench scores slightly lower than those reported in their respective official publications. We attribute this primarily to the following factors:
Randomness: The inherent stochasticity in video generation models can lead to slight variations in results and vBench scores across multiple runs, even with identical settings.
Prompt Extension: Many prior works \cite{li2024cogvideox} may employ specific prompt extension strategies to enrich input prompts. This can influence the content and quality scores of the generated videos. We didn't employ this optimization.
Classifier-Free Guidance (CFG) Scale and Other Sampling Strategies: Different CFG scale values and other sampling parameters (e.g., number of sampling steps) significantly impact generation quality. While we endeavored to follow the descriptions in the respective baseline papers, subtle parameter differences might still exist.
It is worth noting that similar observations have been made in other research. For instance, in the work on Sliding Tile Attention (STA) \cite{zhang2025fastvideo}, their reproduced HunyuanVideo baseline also exhibited lower VBench performance compared to Vbench's official leaderboard. 
Despite these potential metric variations, we emphasize that all methods in this study (including our FPSAttention and all compared baselines) were evaluated under an identical VBench evaluation pipeline and parameter settings, ensuring a fair comparison. Our primary research objective is to demonstrate the significant inference speedup achieved by FPSAttention while maintaining comparable (or superior) generation quality relative to baseline methods.

\begin{table}[h!]
\centering
\caption{Performance comparison of different methods on Wan 13B across VBench metrics. We compare the baseline (Base), SageAttention (SageAtt), SpargeAttention (SpargeAtt), SparseVideoGen (SparseVG), Sliding Tile Attention (STA), and our full joint method (Ours-Q+S). Bold values indicate the best performance for each metric.}
\label{tab:wan13_results}
\resizebox{\textwidth}{!}{
\scriptsize
\begin{tabular}{lcccccc}
\toprule
\textbf{Metric} & \cellcolor{gray!15}\textbf{Base} & \cellcolor{gray!15}\textbf{SageAtt} & \cellcolor{gray!15}\textbf{SpargeAtt} & \cellcolor{gray!15}\textbf{SparseVG} & \cellcolor{gray!15}\textbf{STA} & \cellcolor{gray!15}\textbf{Ours-Q+S} \\
\midrule
Aesthetic Quality & 0.6204 & 0.6209 & 0.5875 & 0.6246 & 0.6033 & \textbf{0.624} \\
Appearance Style & 0.2164 & 0.2163 & 0.7586 & 0.2306 & 0.2303 & 0.2073 \\
Background Consistency & \textbf{0.9691} & 0.9687 & 0.9355 & 0.9589 & 0.9573 & 0.9377 \\
Color & 0.8879 & 0.8825 & 0.8768 & 0.8883 & 0.8814 & \textbf{0.8932} \\
Dynamic Degree & 0.6944 & 0.7028 & 0.6028 & 0.6806 & 0.7028 & \textbf{0.8389} \\
Human Action & 0.796 & 0.8 & 0.78 & 0.816 & 0.778 & \textbf{0.816} \\
Imaging Quality & 0.6715 & 0.6724 & 0.635 & 0.6868 & 0.6577 & \textbf{0.7103} \\
Motion Smoothness & 0.9828 & 0.9828 & 0.9413 & 0.982 & 0.9714 & \textbf{0.9804} \\
Multiple Objects & 0.6627 & 0.6477 & 0.6066 & 0.7012 & 0.6576 & 0.666 \\
Object Class & 0.8299 & 0.8312 & 0.7896 & \textbf{0.8712} & 0.81 & 0.8185 \\
Overall Consistency & 0.6912 & 0.6912 & 0.6975 & 0.708 & 0.6893 & 0.6893 \\
Quality Score & 0.6577 & \textbf{0.8428} & 0.8134 & 0.7421 & 0.8246 & 0.7103 \\
Scene & 0.3669 & 0.3049 & 0.3495 & \textbf{0.4129} & 0.3387 & 0.3182 \\
Semantic Score & \textbf{0.7572} & 0.7091 & 0.6969 & 0.7421 & 0.7077 & 0.7088 \\
Spatial Relationship & 0.7364 & 0.7405 & 0.7526 & \textbf{0.8056} & 0.7405 & 0.7661 \\
Subject Consistency & 0.9528 & 0.953 & 0.9173 & 0.9489 & 0.953 & 0.9435 \\
Temporal Flickering & \textbf{0.9922} & \textbf{0.9922} & 0.969 & 0.9891 & \textbf{0.9922} & 0.9754 \\
Temporal Style & 0.2408 & 0.2408 & 0.6607 & 0.2438 & 0.2408 & \textbf{0.6558} \\
\midrule
\textbf{Total Score} & 0.8153 & 0.8158 & 0.7901 & \textbf{0.8196} & 0.8012 & 0.816 \\
\bottomrule
\end{tabular}
}
\end{table}

The results demonstrate that our joint FP8 quantization and sparsity approach achieves competitive or superior performance compared to specialized methods focusing solely on either quantization or sparsity. For the Wan 1.3B model, our method achieves the highest total score (0.8160), outperforming the baseline (0.8019) while providing significant computational benefits. Similarly, for the Wan 13B model, our approach performs on par with the best-performing methods while offering substantial memory and compute savings through the combination of quantization and structured sparsity.

\section{Training Hyperparameters}
\label{subsec:hyperparameters}

Table~\ref{tab:hyperparameters_merged} presents the key hyperparameters used in our experiments for both Wan 1.3B and 13B model training configurations. These hyperparameters were carefully selected to balance training stability, convergence speed, and final model performance while accommodating the constraints imposed by FP8 quantization and structured sparsity.

\begin{table}[h!]
\centering
\caption{Comprehensive hyperparameter configuration for Wan 1.3B and 13B model training and evaluation. The table covers model architecture specifications, training parameters, diffusion scheduler settings, data configuration, and system-level precision settings used in our experiments.}
\label{tab:hyperparameters_merged}
\resizebox{\textwidth}{!}{
\scriptsize
\begin{tabular}{lccc}
\toprule
\textbf{Category} & \textbf{Parameter} & \textbf{Wan 1.3B} & \textbf{Wan 13B} \\
\midrule
\textbf{Model Architecture} & Model Type & WanX21FPS & \textbf{WanX21FPS-13B} \\
& Model Dimension & 1536 & \textbf{5120} \\
& Number of Layers & 30 & \textbf{40} \\
& Number of Heads & 12 & \textbf{40} \\
& FFN Dimension & 8960 & \textbf{13824} \\
& Input/Output Dimension & 16 & 16 \\
& Frequency Dimension & 256 & 256 \\
& Text Dimension & 4096 & 4096 \\
& Patch Size & [1, 2, 2] & [1, 2, 2] \\
\midrule
\textbf{Training} & Learning Rate & 5e-6 & 5e-6 \\
& Weight Decay & 1e-4 & 1e-4 \\
& Gradient Clipping & 1.0 & 1.0 \\
& Warmup Steps & 200 & 200 \\
& EMA Decay & 0.99 & 0.99 \\
& Adam Epsilon & 1e-15 & 1e-15 \\
\midrule
\textbf{Diffusion Scheduler} & Scheduler Type & rflow-wanx & rflow-wanx \\
& Number of Timesteps & 1000 & 1000 \\
& Sample Steps & 50 & 50 \\
& CFG Scale & 5.0 & 5.0 \\
& Sample Shift & 5.0 & 5.0 \\
& Transform Scale & 5.0 & 5.0 \\
& Sample Method & logit-normal & logit-normal \\
\midrule
\textbf{Data \& Sequence} & Text Length & 512 & 512 \\
& Max Sequence Length & 75600 & 75600 \\
& Sample FPS & 16 & 16 \\
& Video Resolution & 480p & 480p \\
& Video Duration & 5s & 5s \\
& Prompt Uncond Probability & 0.1 & 0.1 \\
\midrule
\textbf{System \& Precision} & Data Type & fp8 & fp8 \\
& Training Mode & FSDP & FSDP \\
& Gradient Checkpointing & True & True \\
& Quantization & True & True \\
& Sequence Parallel Degree & 1 & \textbf{4} \\
\bottomrule
\end{tabular}
}
\end{table}

\newpage

\section{Limitations}
\label{sec:limitations}

While FPSAttention demonstrates strong performance across our evaluation scenarios, there are some considerations for broader adoption. Our approach works best with modern FP8-capable GPUs such as NVIDIA Hopper architectures, though it can still provide benefits on older hardware with reduced FP8 acceleration. The method benefits from quantization-aware training to achieve optimal results, which involves a moderate increase in training time compared to post-training quantization approaches. The denoising step-aware scheduling includes several hyperparameters (transition points $\alpha_1$ and $\alpha_2$, quantization granularities, and window sizes) that can be optimized for different model architectures and datasets. Our current evaluation focuses on the Wan2.1 architecture, and the approach shows strong promise for extension to other video diffusion transformer architectures (e.g., HunyuanVideo, CogVideoX). Additionally, while our tile-wise approach achieves good hardware utilization across tested configurations, there are opportunities for architecture-specific optimization to further improve performance on different GPU memory hierarchies.

\section{Visualization}
\label{sec:visualization}

The following qualitative comparison demonstrates that our FPSAttention method generates video frames that are visually nearly identical to the baseline Wan model with size of 1.3B. This visual similarity across diverse scenarios—including boats, fish, dogs, desert landscapes, couples, trains, cars, cats, and robot DJs—validates that our joint FP8 quantization and sparsity optimization achieves essentially lossless performance while providing substantial computational acceleration.

\clearpage
\begin{table}[h!]
    \centering
    \vspace{-2em}
    \caption{Qualitative comparison on the boat group. Prompt: `A boat sailing leisurely along the Seine River with the Eiffel Tower in background in super slow motion' . Top: Baseline; Bottom: FPSAttention.}
    \label{tab:vis_boat}
    \begin{tabular}{cccc}
        \includegraphics[width=0.22\textwidth]{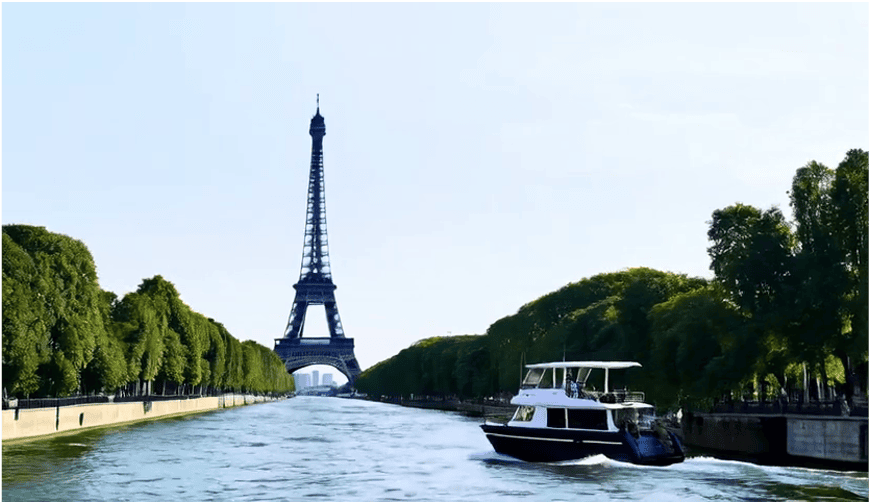} &
        \includegraphics[width=0.22\textwidth]{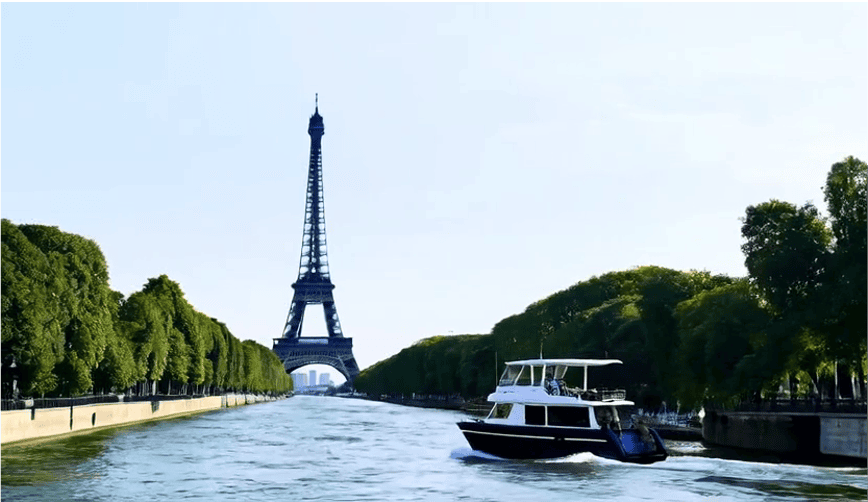} &
        \includegraphics[width=0.22\textwidth]{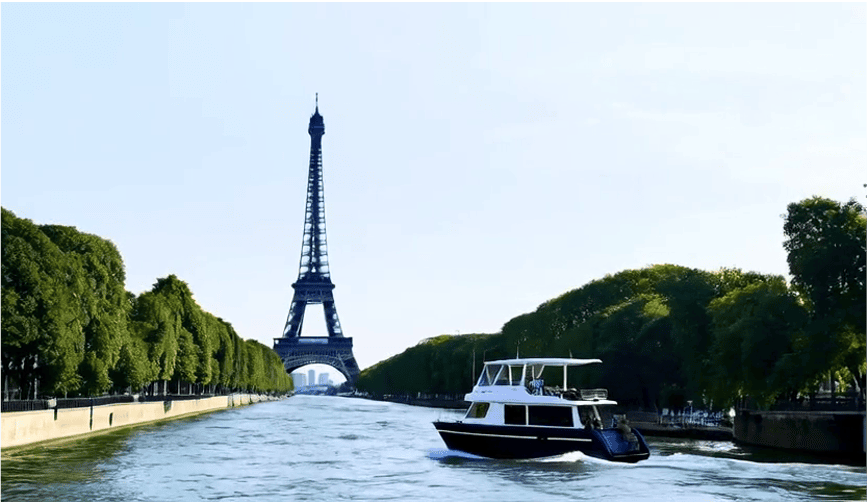} &
        \includegraphics[width=0.22\textwidth]{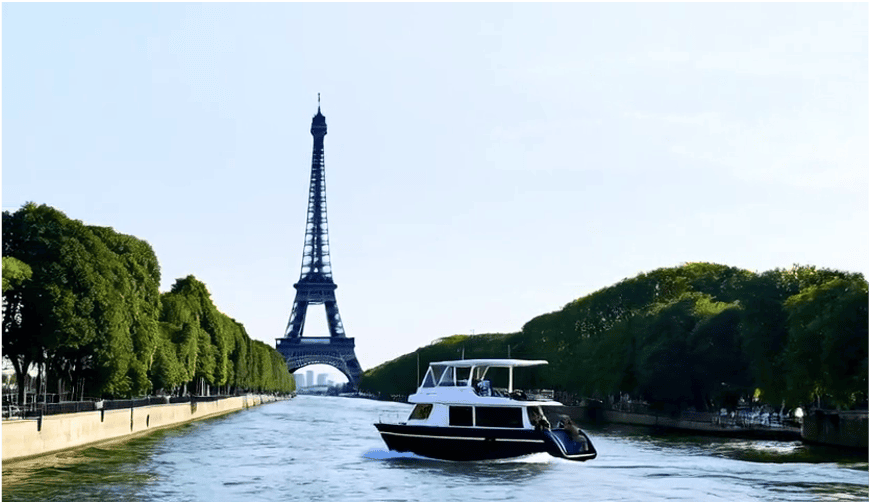} \\
        \multicolumn{4}{c}{\textbf{Baseline}} \\
        \includegraphics[width=0.22\textwidth]{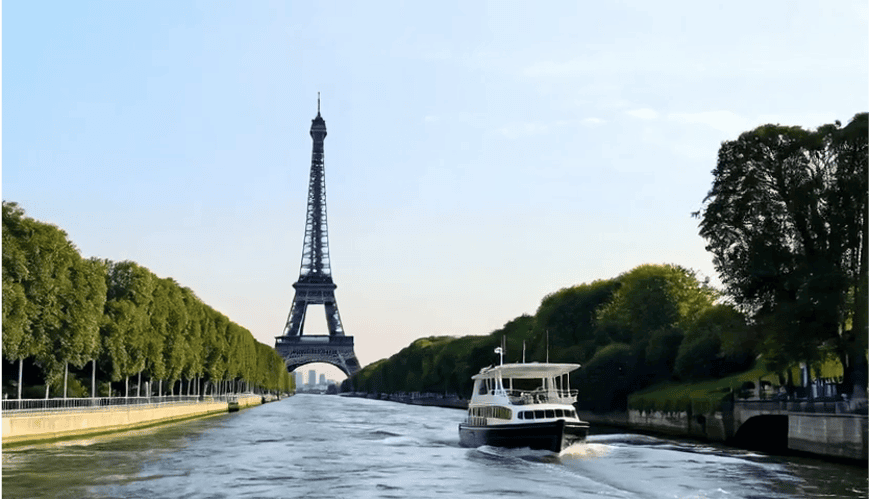} &
        \includegraphics[width=0.22\textwidth]{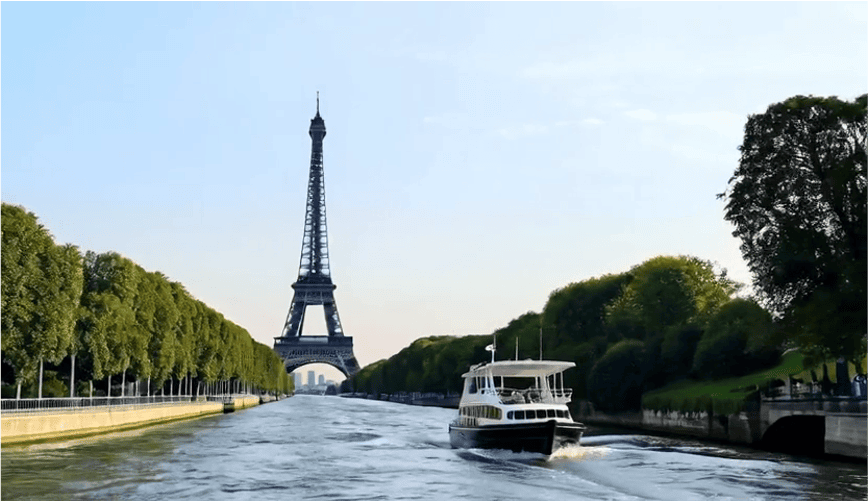} &
        \includegraphics[width=0.22\textwidth]{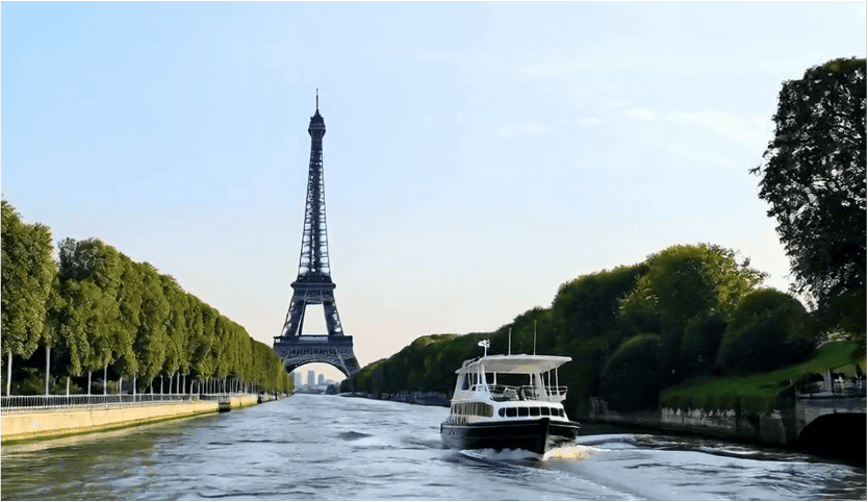} &
        \includegraphics[width=0.22\textwidth]{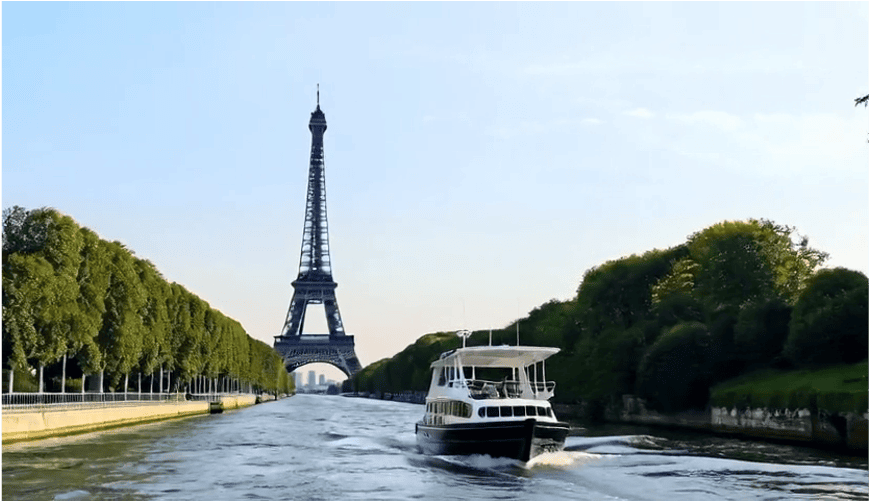} \\
        \multicolumn{4}{c}{\textbf{FPSAttention}} \\
    \end{tabular}
\end{table}

\begin{table}[h!]
    \centering
    \caption{Qualitative comparison on the fish group. Prompt : `Golden fish swimming in the ocean'. Top: Baseline; Bottom: FPSAttention.}
    \label{tab:vis_fish}
    \begin{tabular}{cccc}
        \includegraphics[width=0.22\textwidth]{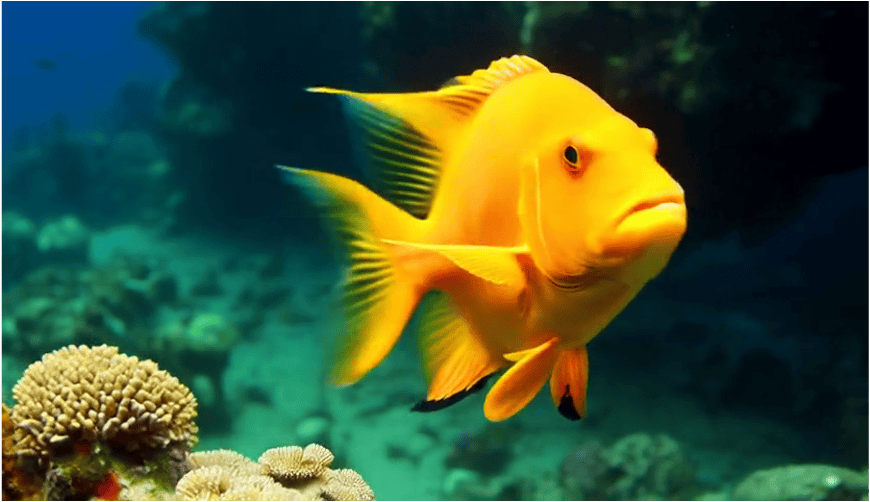} &
        \includegraphics[width=0.22\textwidth]{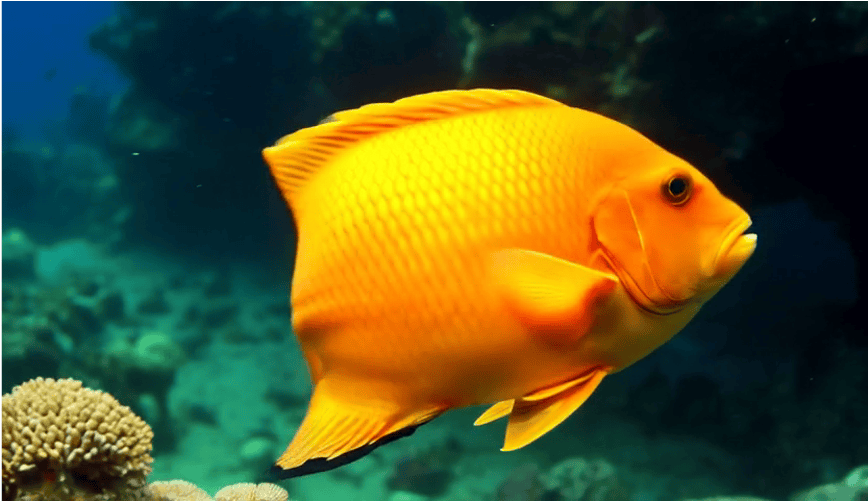} &
        \includegraphics[width=0.22\textwidth]{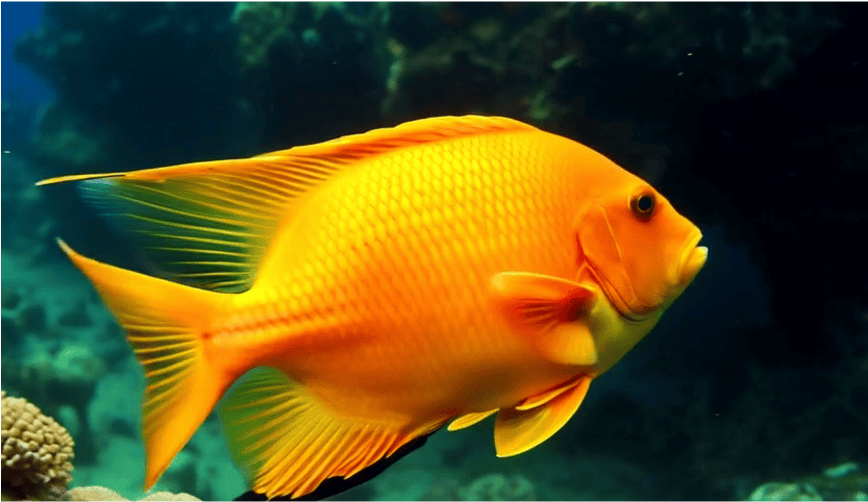} &
        \includegraphics[width=0.22\textwidth]{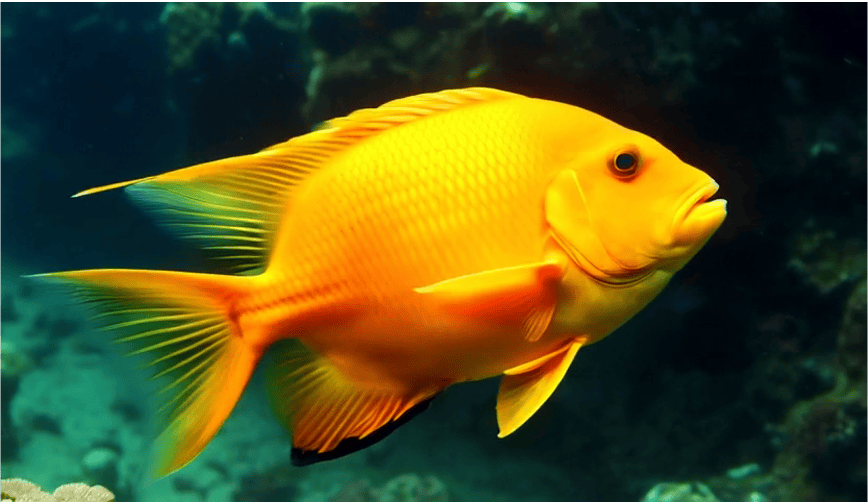} \\
        \multicolumn{4}{c}{\textbf{Baseline}} \\
        \includegraphics[width=0.22\textwidth]{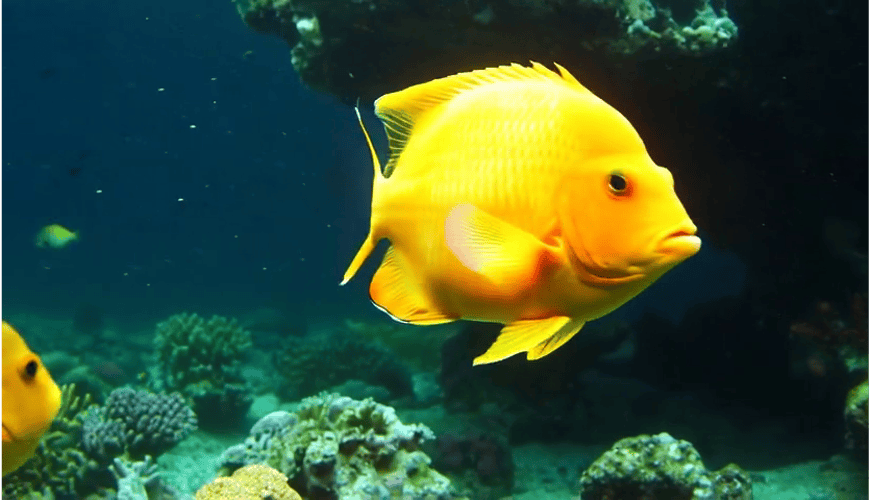} &
        \includegraphics[width=0.22\textwidth]{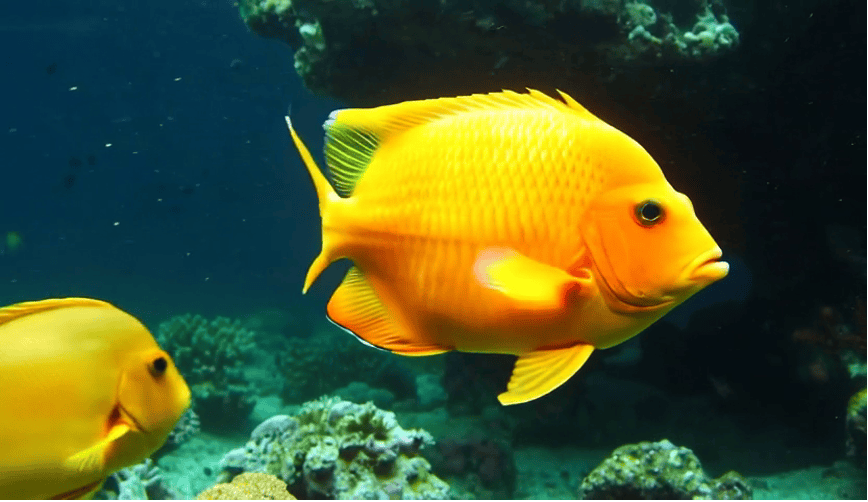} &
        \includegraphics[width=0.22\textwidth]{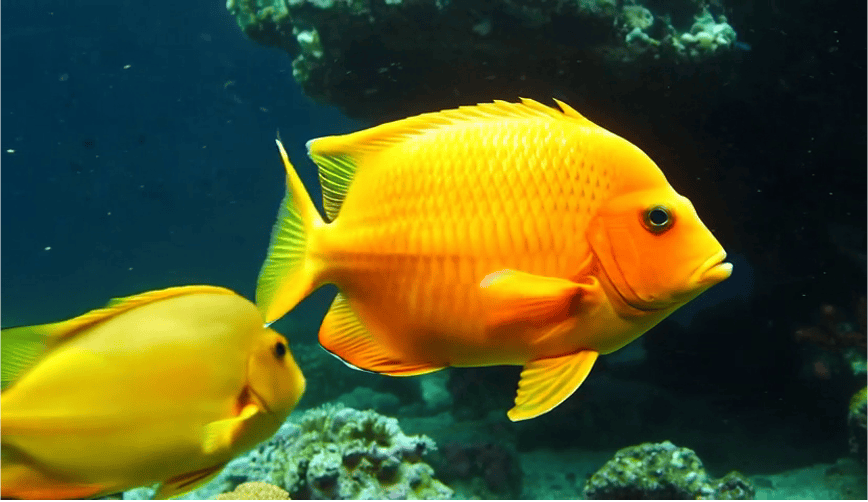} &
        \includegraphics[width=0.22\textwidth]{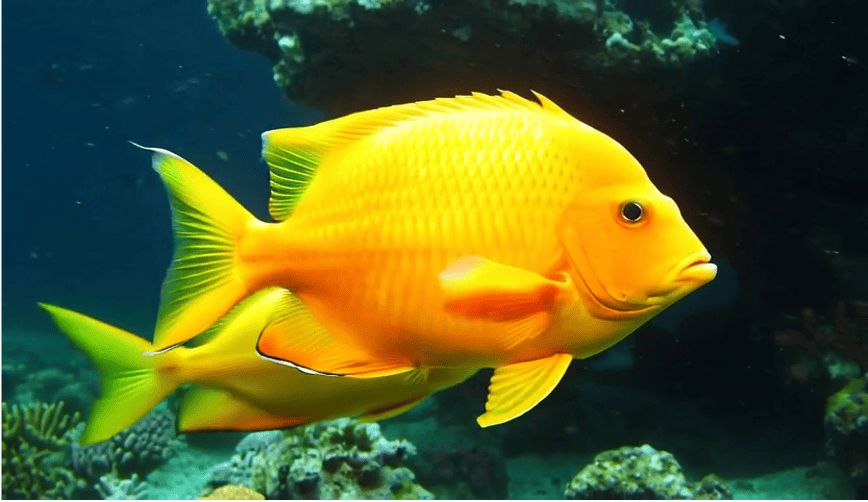} \\
        \multicolumn{4}{c}{\textbf{FPSAttention}} \\
    \end{tabular}
\end{table}

\begin{table}[h!]
    \centering
    \caption{Qualitative comparison on the dog group. Prompt: `A dog enjoying a peaceful walk'. Top: Baseline; Bottom: FPSAttention.}
    \label{tab:vis_dog}
    \begin{tabular}{cccc}
        \includegraphics[width=0.22\textwidth]{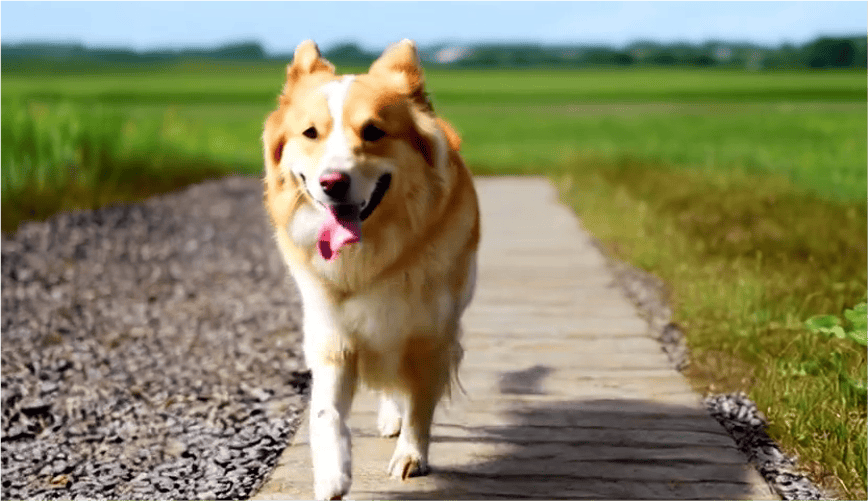} &
        \includegraphics[width=0.22\textwidth]{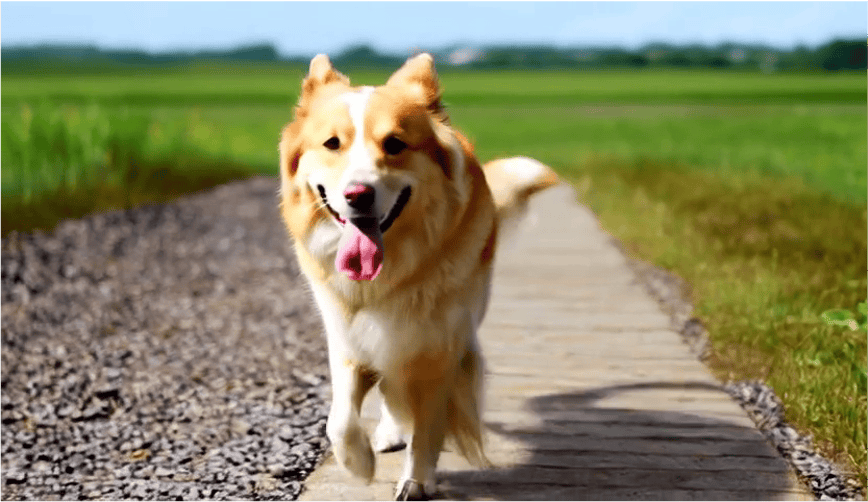} &
        \includegraphics[width=0.22\textwidth]{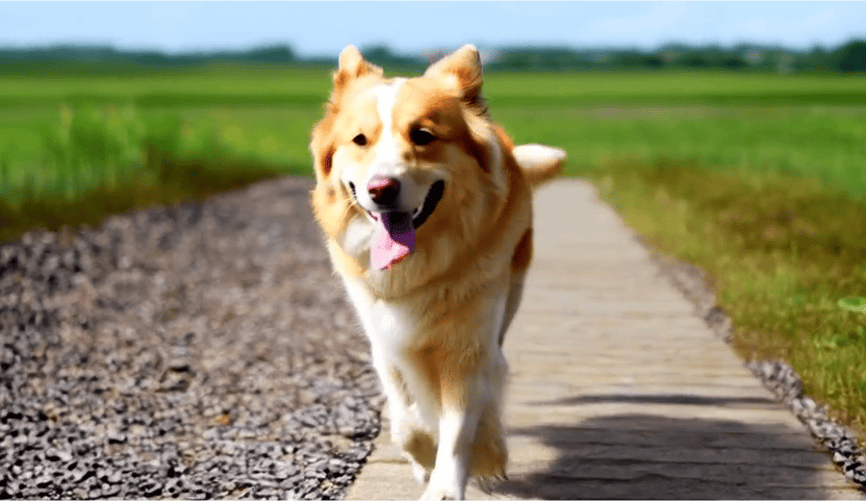} &
        \includegraphics[width=0.22\textwidth]{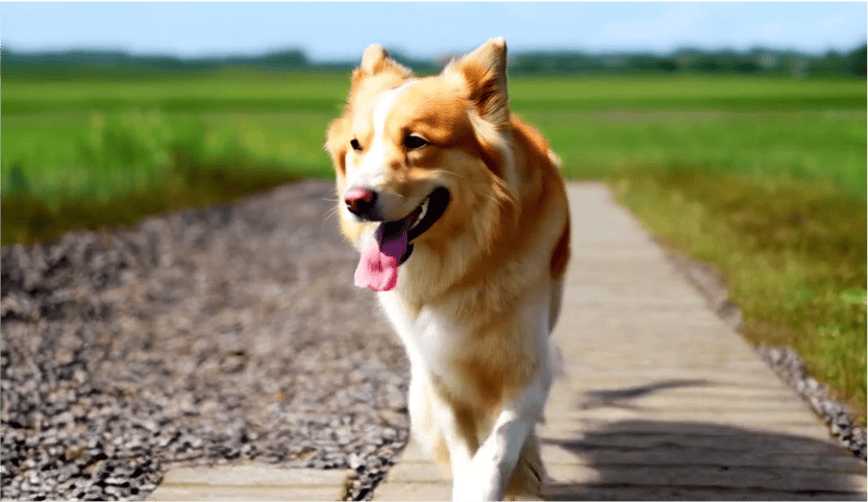} \\
        \multicolumn{4}{c}{\textbf{Baseline}} \\
        \includegraphics[width=0.22\textwidth]{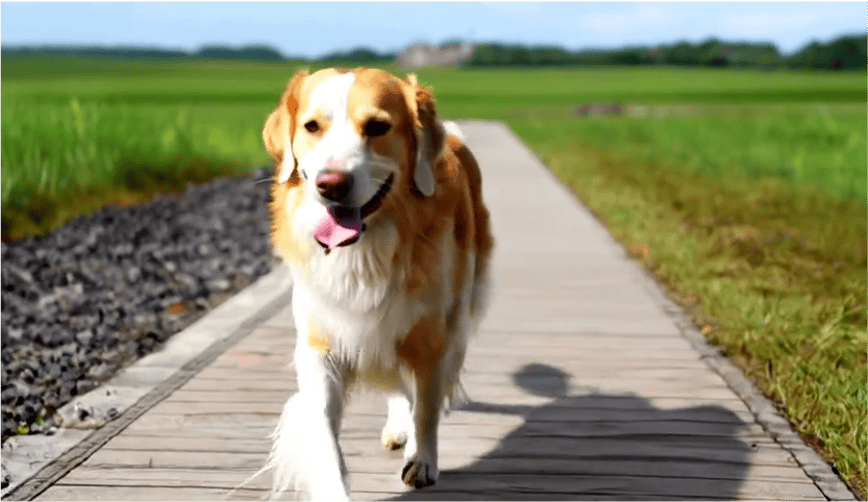} &
        \includegraphics[width=0.22\textwidth]{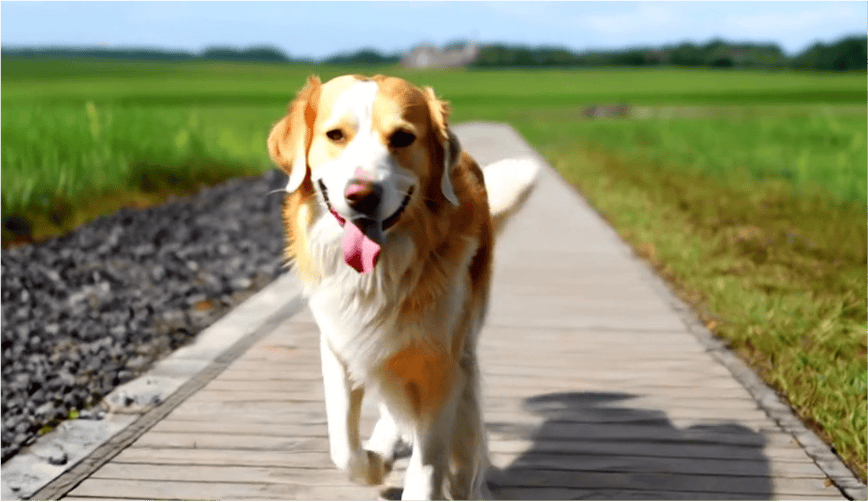} &
        \includegraphics[width=0.22\textwidth]{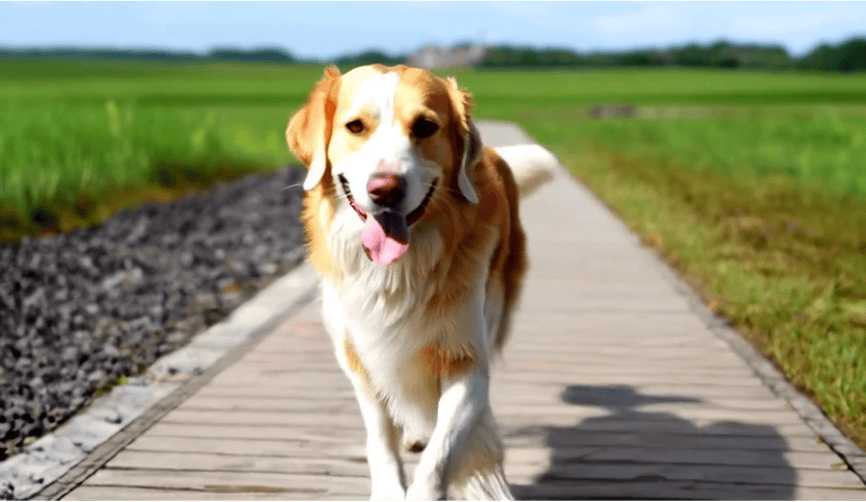} &
        \includegraphics[width=0.22\textwidth]{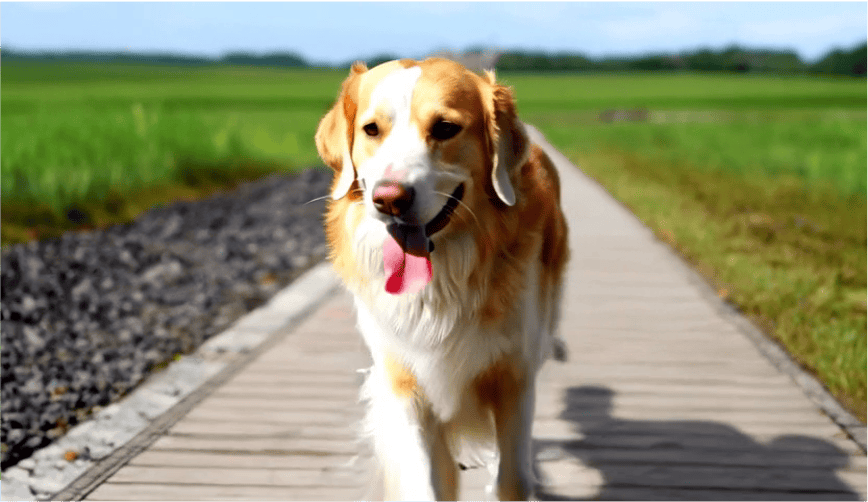} \\
        \multicolumn{4}{c}{\textbf{FPSAttention}} \\
    \end{tabular}
\end{table}

\begin{table}[h!]
    \centering
    \caption{Qualitative comparison on the desert group. Prompt: `Static view on a desert scene with an oasis palm trees and a clear calm pool of water'. Top: Baseline; Bottom: FPSAttention.}
    \label{tab:vis_desert}
    \begin{tabular}{cccc}
        \includegraphics[width=0.22\textwidth]{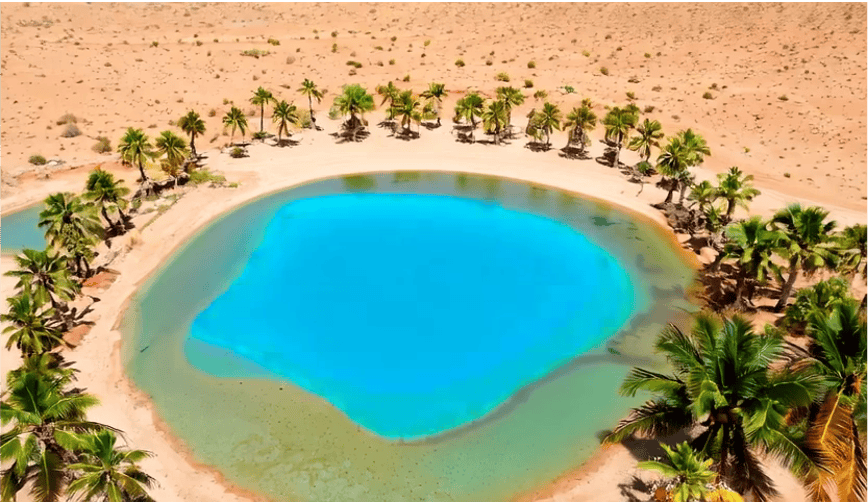} &
        \includegraphics[width=0.22\textwidth]{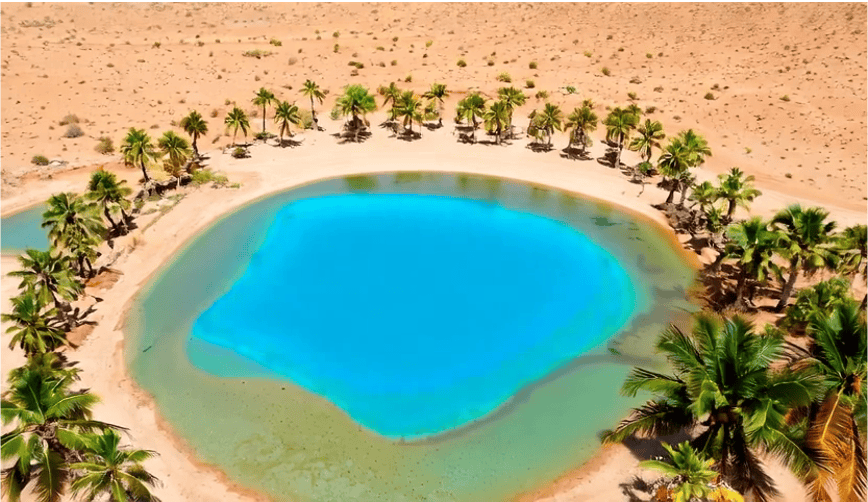} &
        \includegraphics[width=0.22\textwidth]{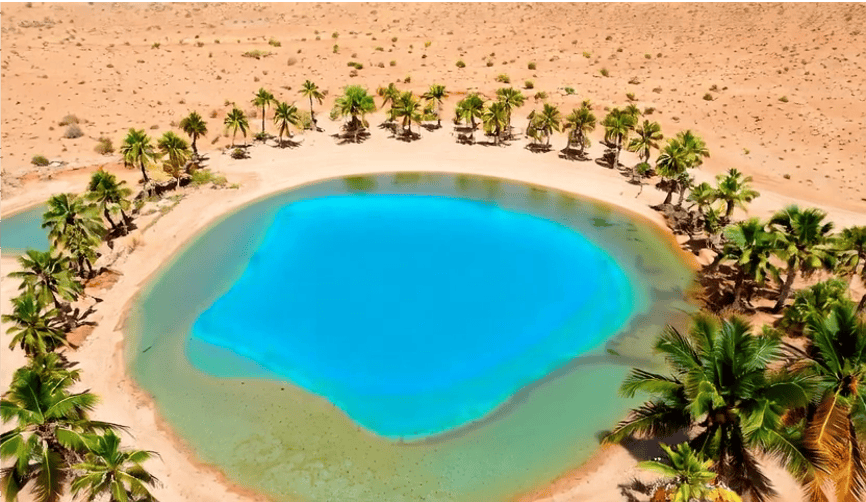} &
        \includegraphics[width=0.22\textwidth]{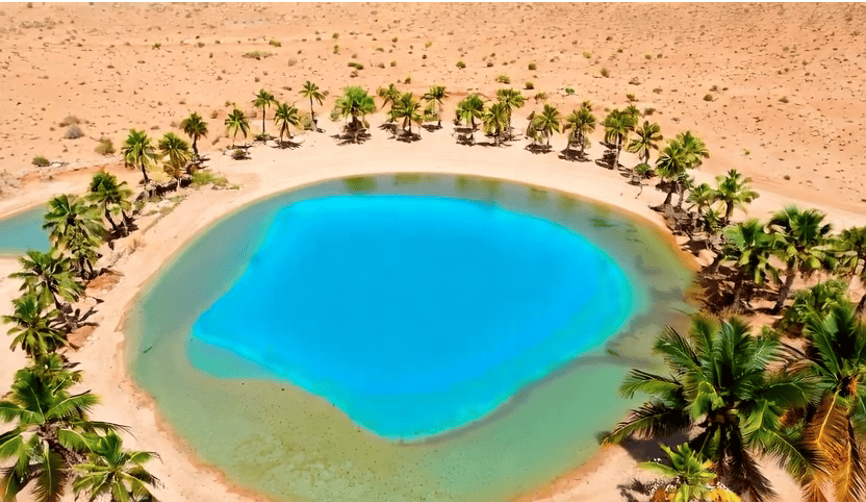} \\
        \multicolumn{4}{c}{\textbf{Baseline}} \\
        \includegraphics[width=0.22\textwidth]{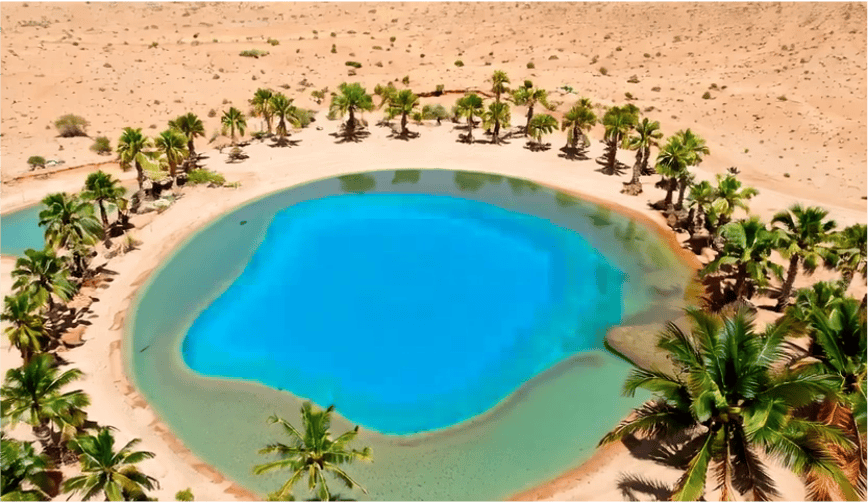} &
        \includegraphics[width=0.22\textwidth]{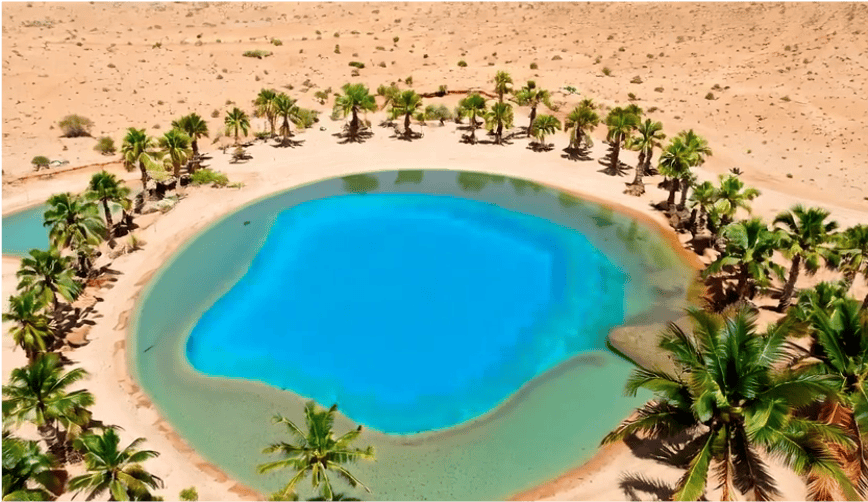} &
        \includegraphics[width=0.22\textwidth]{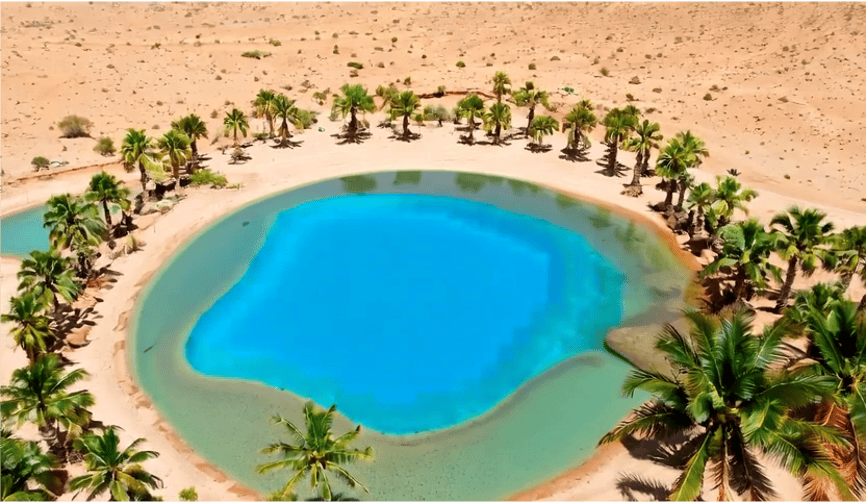} &
        \includegraphics[width=0.22\textwidth]{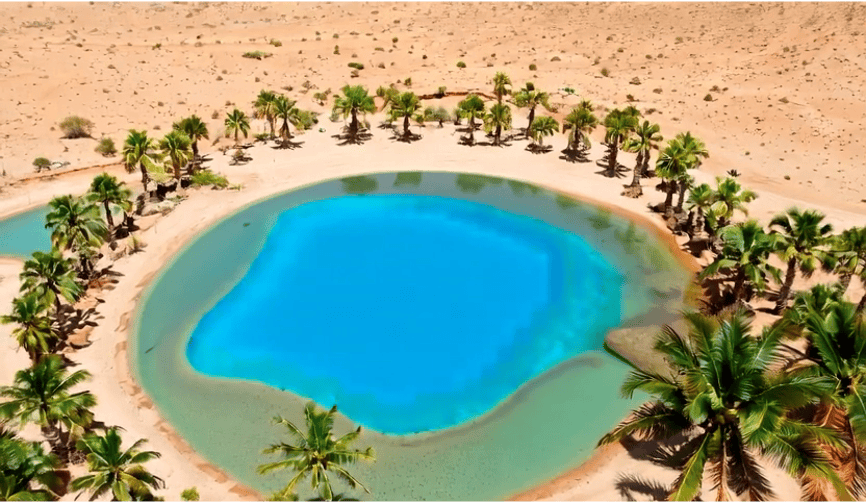} \\
        \multicolumn{4}{c}{\textbf{FPSAttention}} \\
    \end{tabular}
\end{table}

\begin{table}[h!]
    \centering
    \caption{Qualitative comparison on the couple group. Prompt: `A couple in formal evening wear going home get caught in a heavy downpour with umbrellas'. Top: Baseline; Bottom: FPSAttention.}
    \label{tab:vis_couple}
    \begin{tabular}{cccc}
        \includegraphics[width=0.22\textwidth]{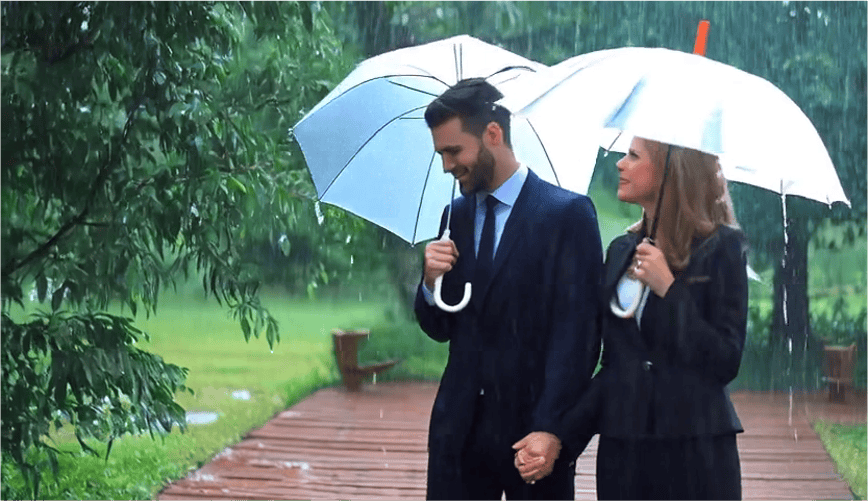} &
        \includegraphics[width=0.22\textwidth]{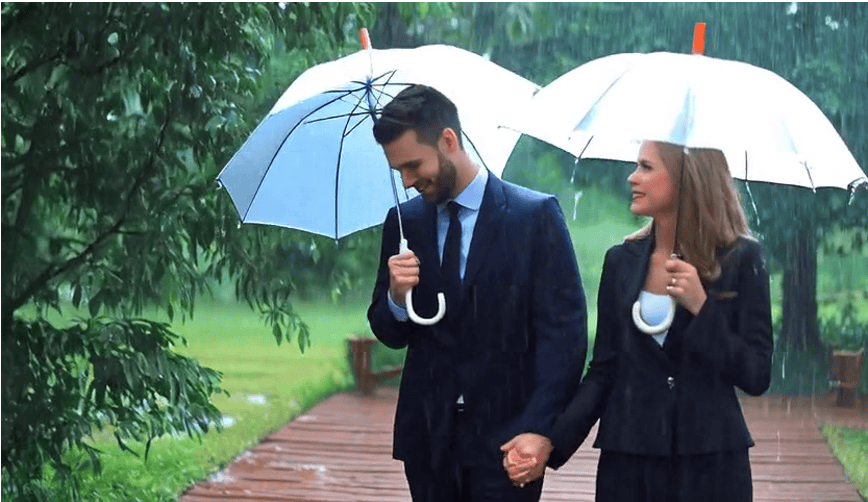} &
        \includegraphics[width=0.22\textwidth]{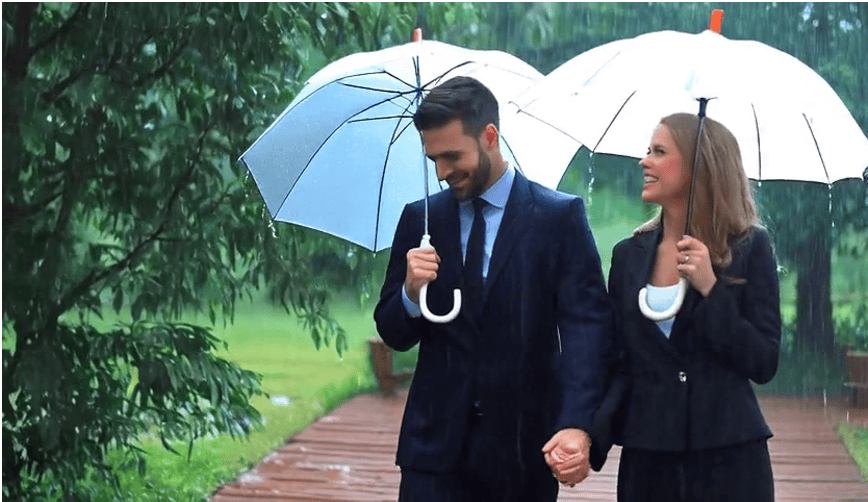} &
        \includegraphics[width=0.22\textwidth]{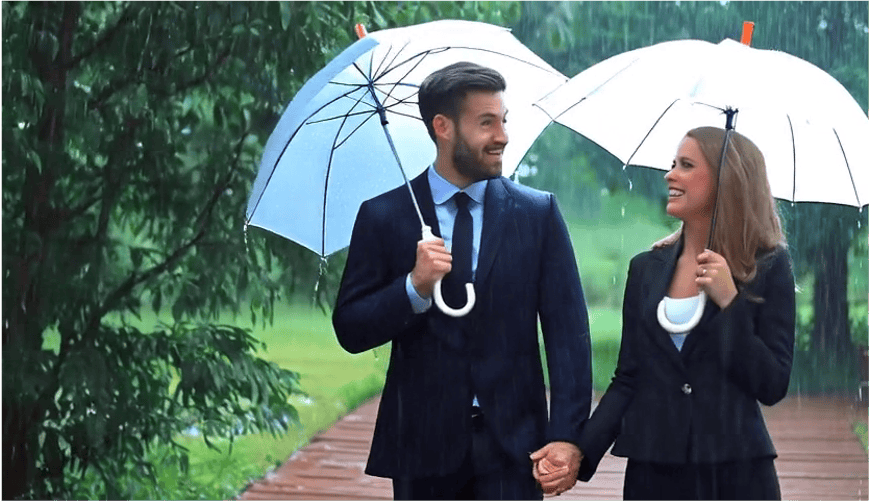} \\
        \multicolumn{4}{c}{\textbf{Baseline}} \\
        \includegraphics[width=0.22\textwidth]{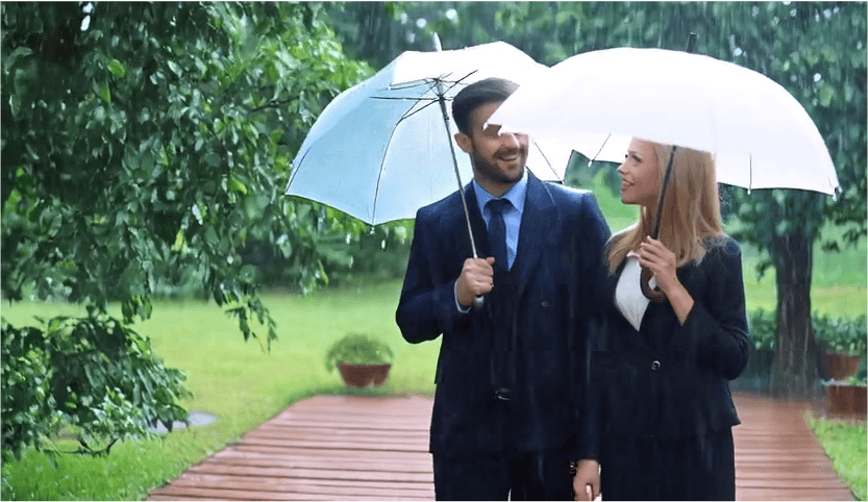} &
        \includegraphics[width=0.22\textwidth]{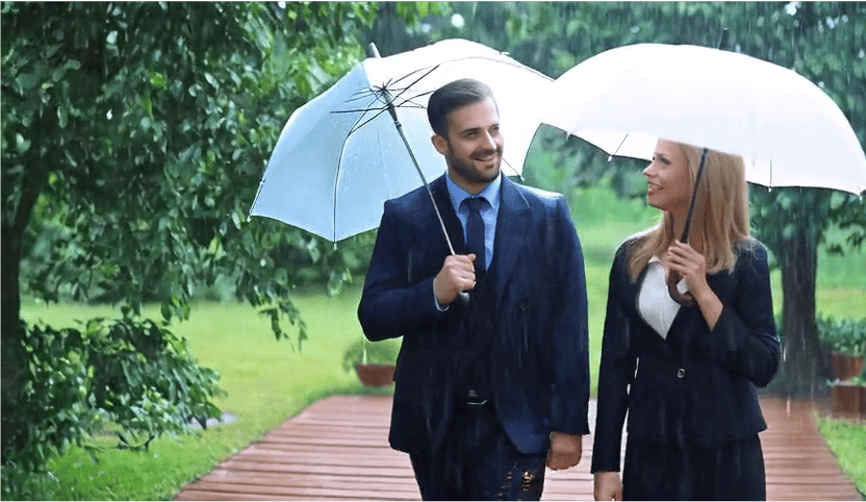} &
        \includegraphics[width=0.22\textwidth]{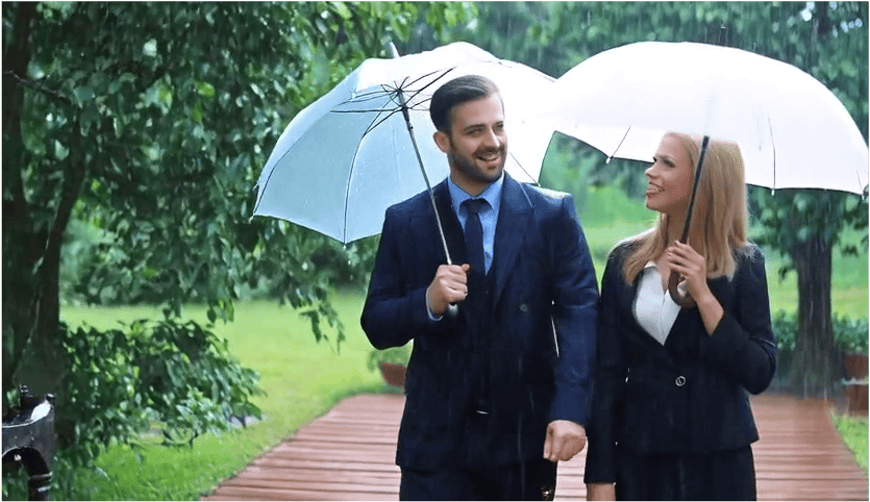} &
        \includegraphics[width=0.22\textwidth]{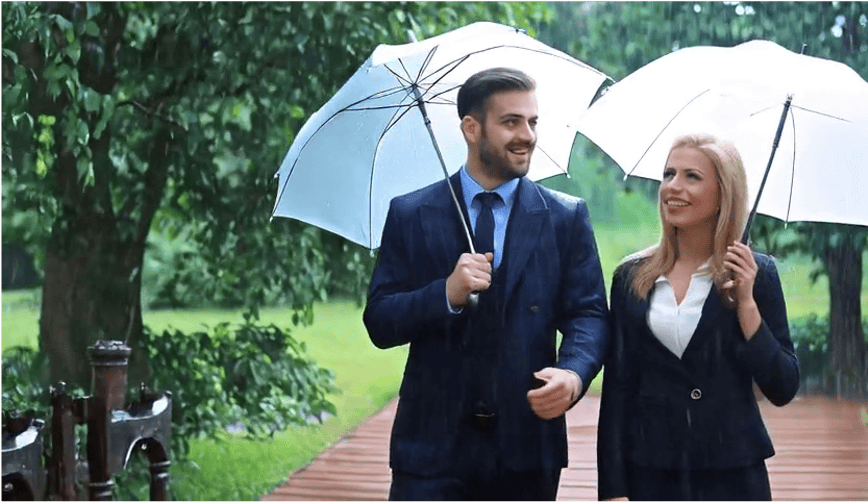} \\
        \multicolumn{4}{c}{\textbf{FPSAttention}} \\
    \end{tabular}
\end{table}

\begin{table}[h!]
    \centering
    \caption{Qualitative comparison on the train group. Prompt: `A train accelerating to gain speed'. Top: Baseline; Bottom: FPSAttention.}
    \label{tab:vis_train}
    \begin{tabular}{cccc}
        \includegraphics[width=0.22\textwidth]{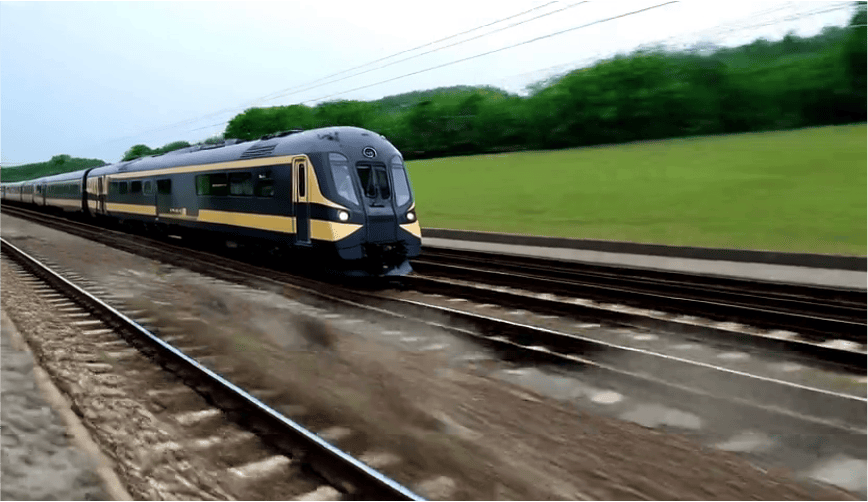} &
        \includegraphics[width=0.22\textwidth]{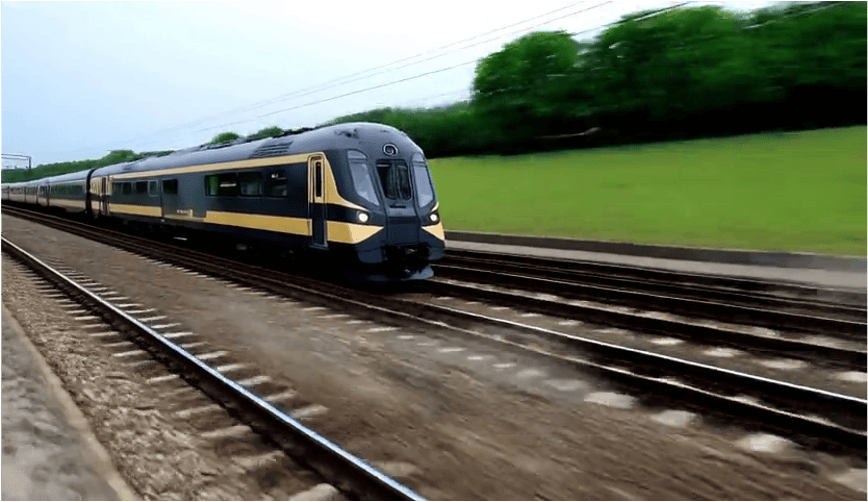} &
        \includegraphics[width=0.22\textwidth]{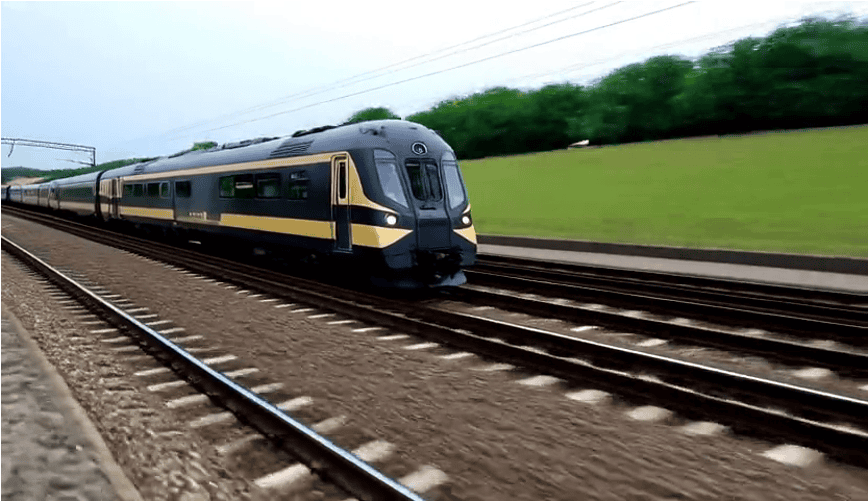} &
        \includegraphics[width=0.22\textwidth]{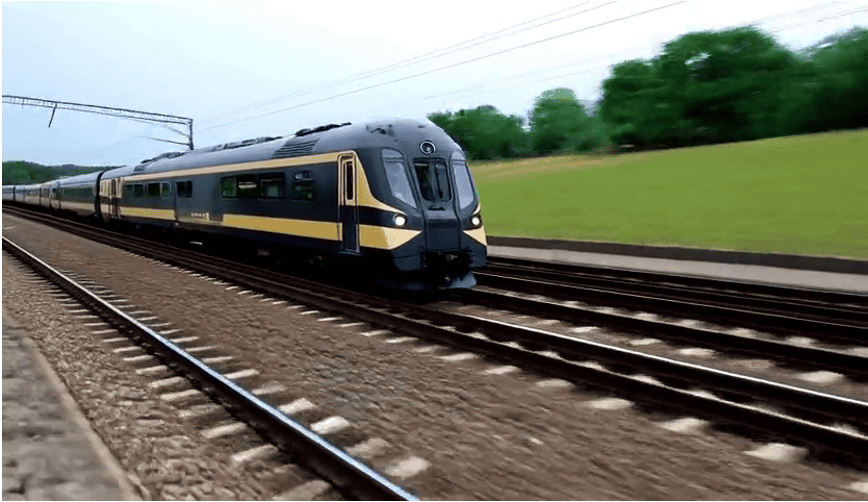} \\
        \multicolumn{4}{c}{\textbf{Baseline}} \\
        \includegraphics[width=0.22\textwidth]{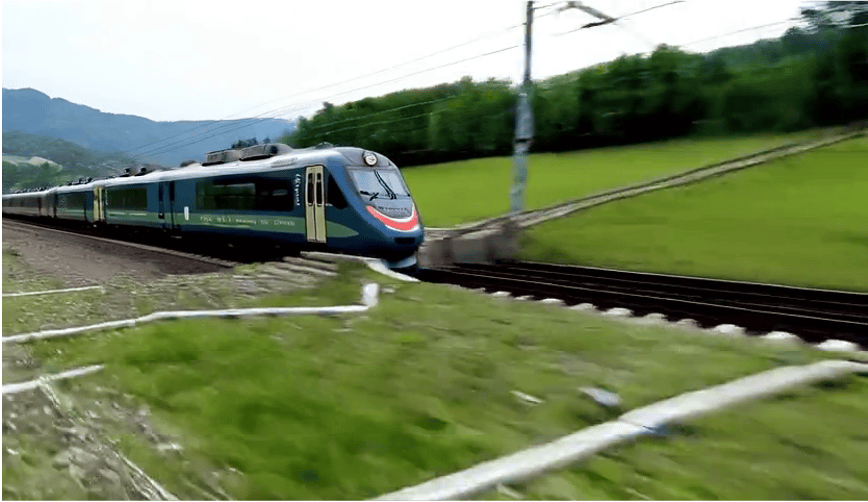} &
        \includegraphics[width=0.22\textwidth]{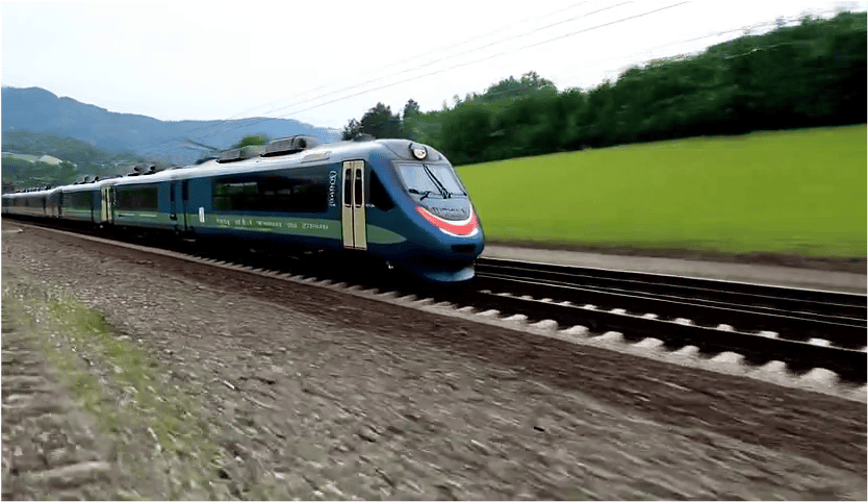} &
        \includegraphics[width=0.22\textwidth]{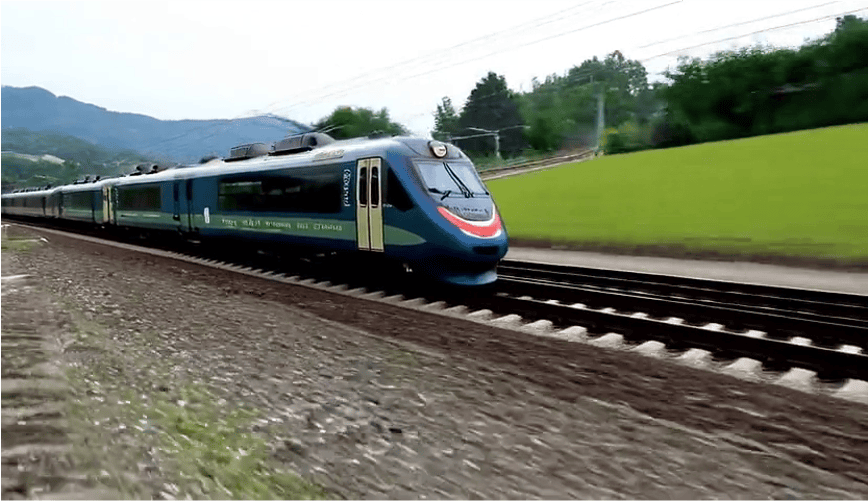} &
        \includegraphics[width=0.22\textwidth]{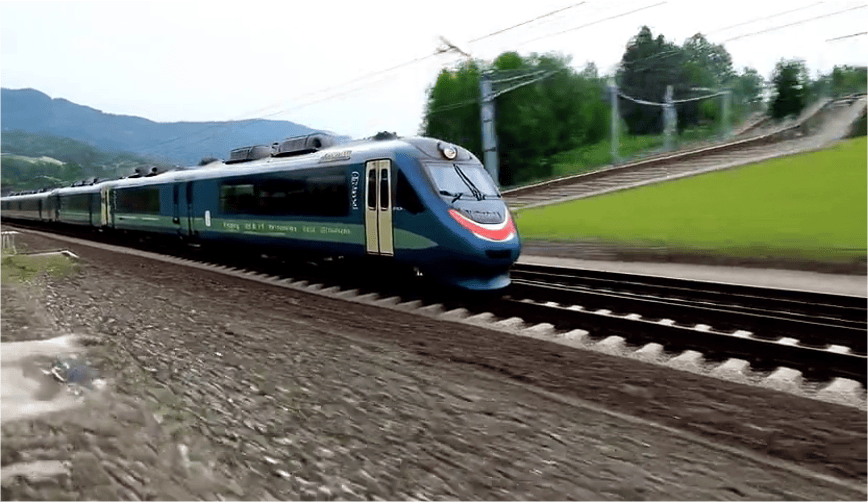} \\
        \multicolumn{4}{c}{\textbf{FPSAttention}} \\
    \end{tabular}
\end{table}

\begin{table}[h!]
    \centering
    \caption{Qualitative comparison on the rock group. Prompt: `A tranquil tableau of at the edge of the Arabian Desert, the ancient city of Petra beckoned with its enigmatic rock-carved façades'. Top: Baseline; Bottom: FPSAttention.}
    \label{tab:vis_train}
    \begin{tabular}{cccc}
        \includegraphics[width=0.22\textwidth]{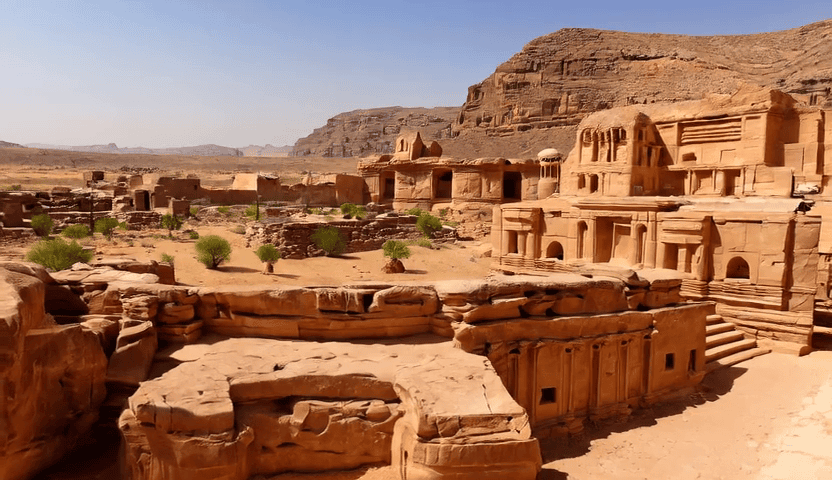} &
        \includegraphics[width=0.22\textwidth]{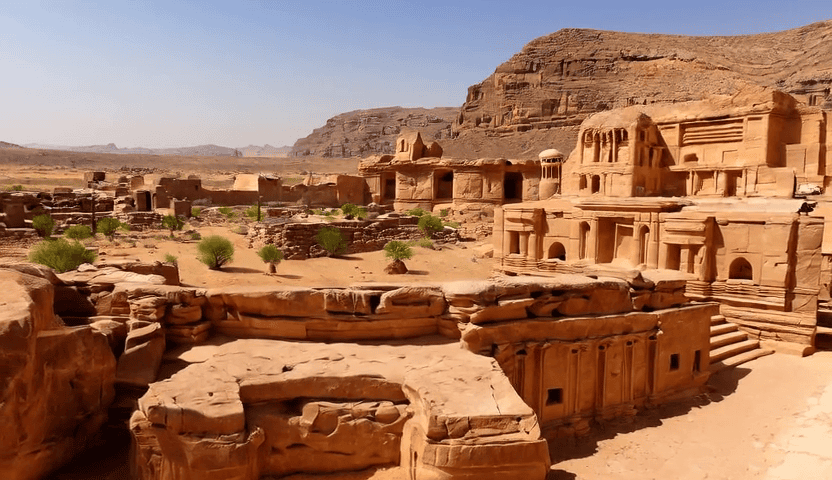} &
        \includegraphics[width=0.22\textwidth]{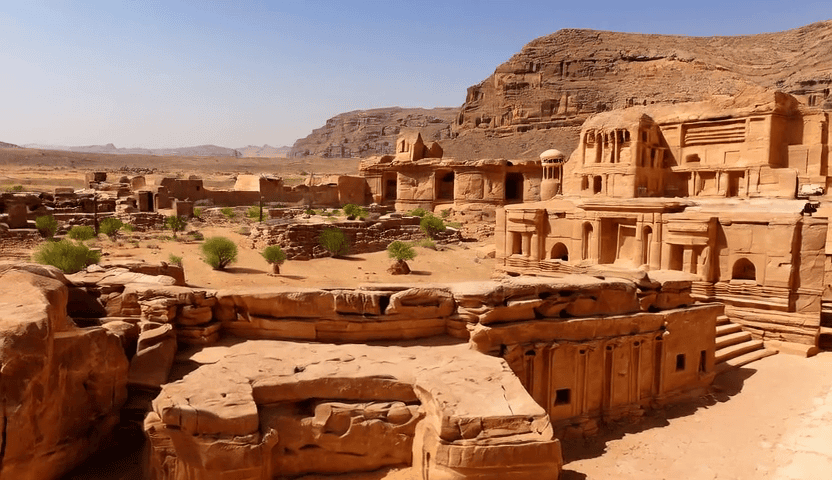} &
        \includegraphics[width=0.22\textwidth]{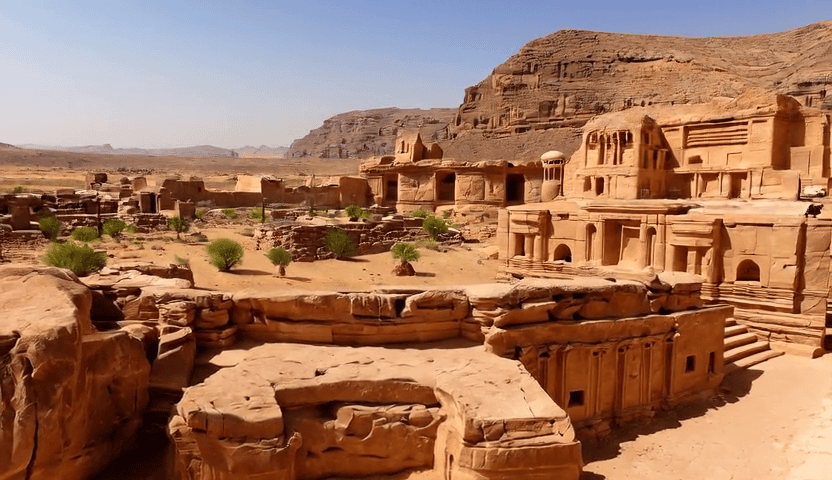} \\
        \multicolumn{4}{c}{\textbf{Baseline}} \\
        \includegraphics[width=0.22\textwidth]{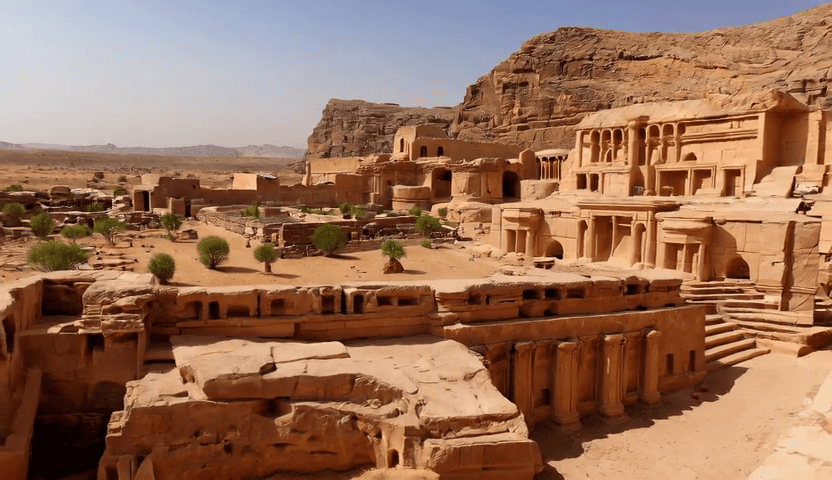} &
        \includegraphics[width=0.22\textwidth]{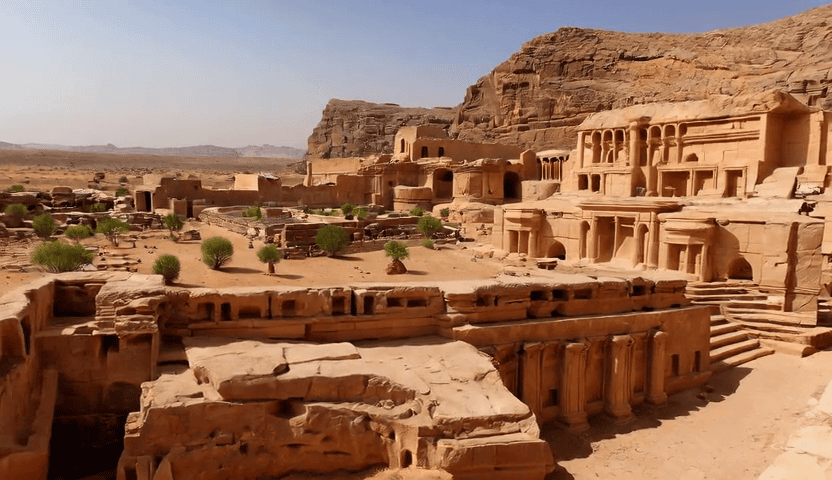} &
        \includegraphics[width=0.22\textwidth]{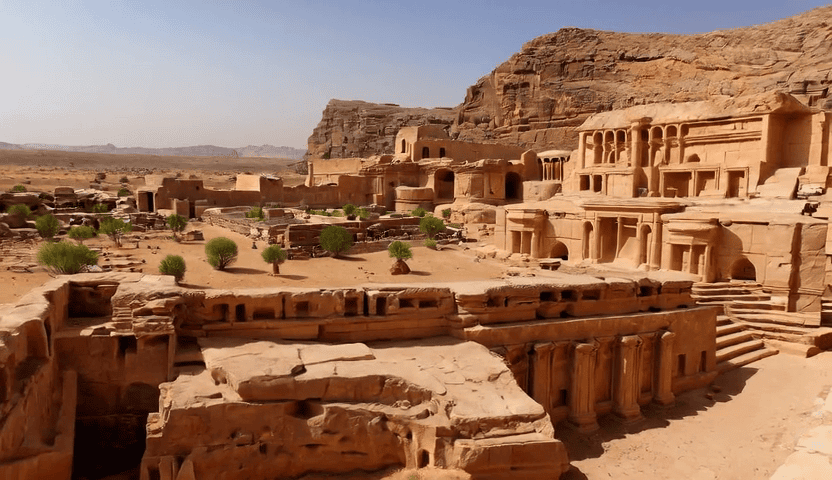} &
        \includegraphics[width=0.22\textwidth]{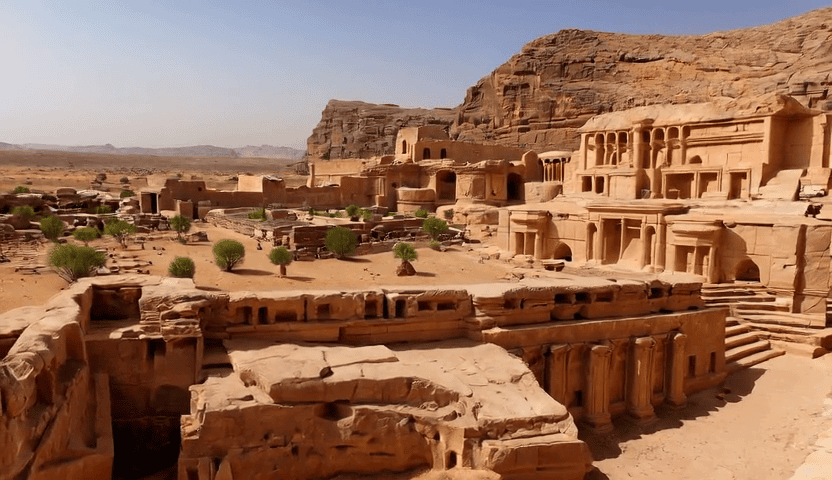} \\
        \multicolumn{4}{c}{\textbf{FPSAttention}} \\
    \end{tabular}
\end{table}

\begin{table}[h!]
    \centering
    \caption{Qualitative comparison on the nursery group. Prompt: `Nursery'. Top: Baseline; Bottom: FPSAttention.}
    \label{tab:vis_train}
    \begin{tabular}{cccc}
        \includegraphics[width=0.22\textwidth]{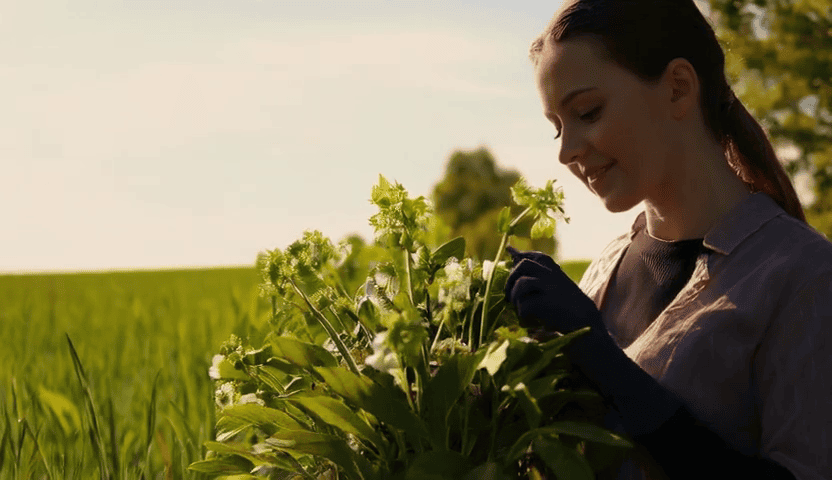} &
        \includegraphics[width=0.22\textwidth]{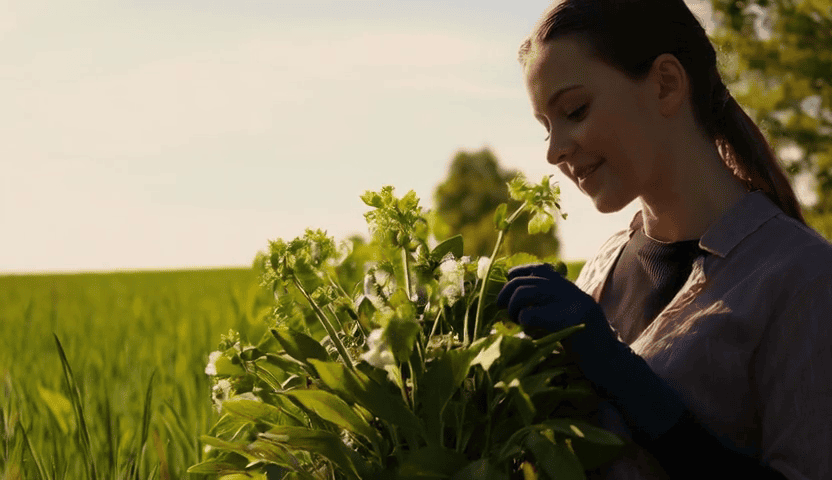} &
        \includegraphics[width=0.22\textwidth]{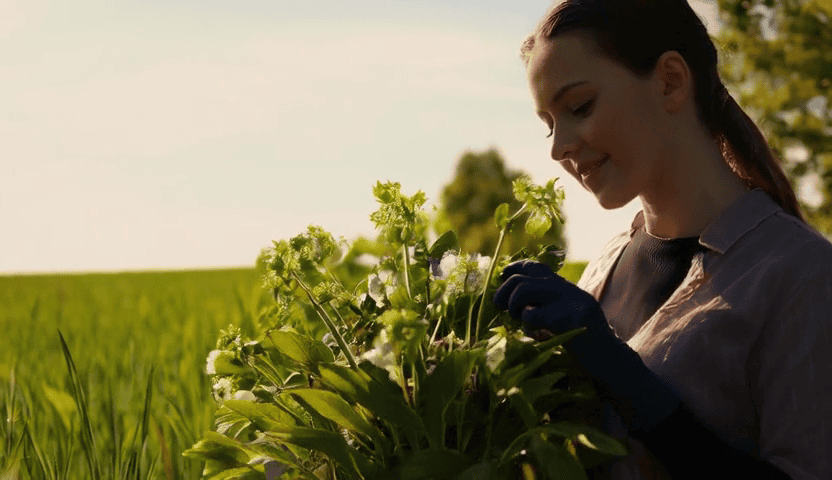} &
        \includegraphics[width=0.22\textwidth]{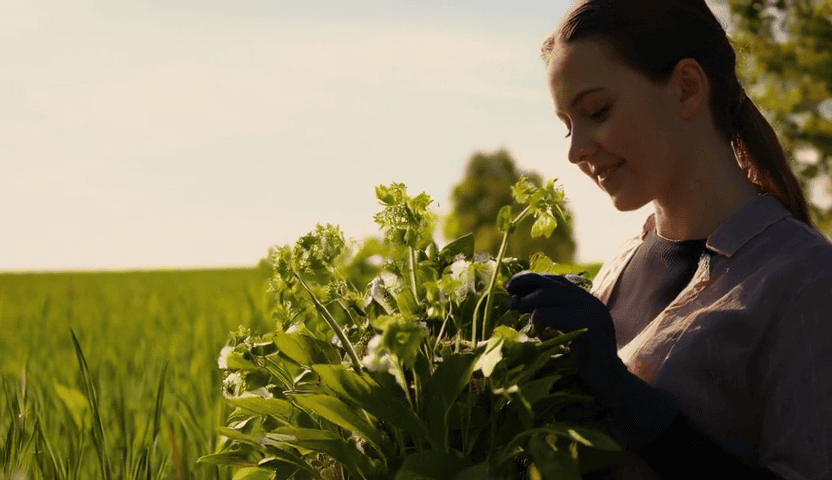} \\
        \multicolumn{4}{c}{\textbf{Baseline}} \\
        \includegraphics[width=0.22\textwidth]{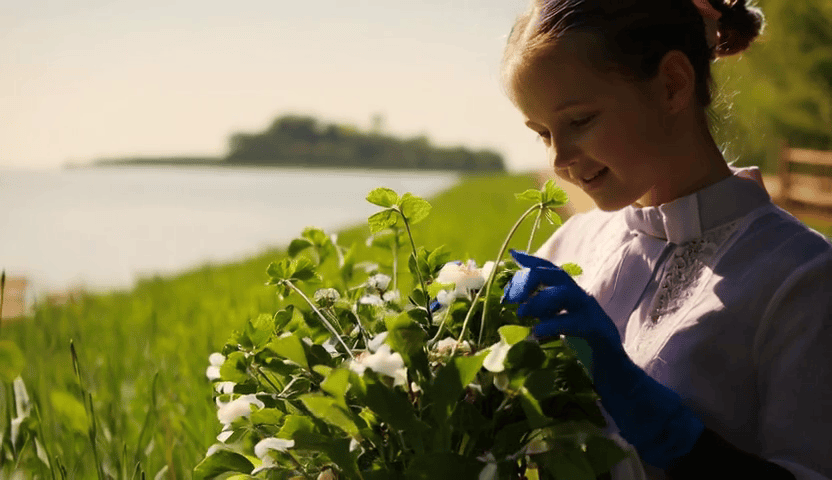} &
        \includegraphics[width=0.22\textwidth]{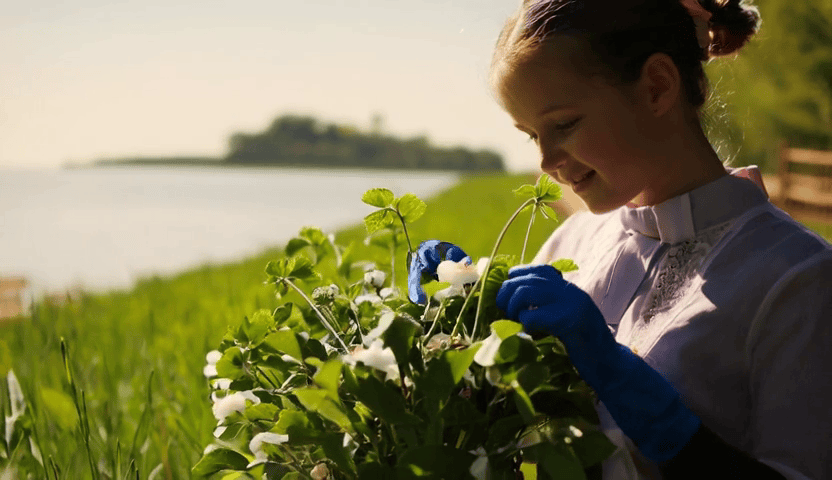} &
        \includegraphics[width=0.22\textwidth]{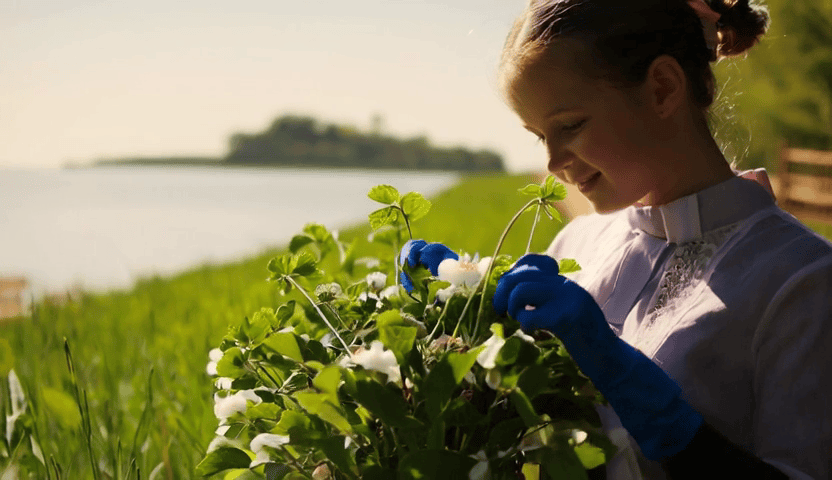} &
        \includegraphics[width=0.22\textwidth]{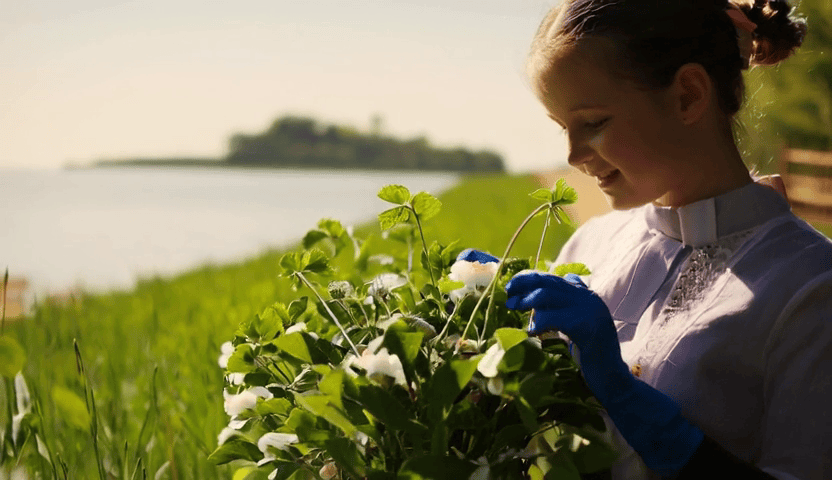} \\
        \multicolumn{4}{c}{\textbf{FPSAttention}} \\
    \end{tabular}
\end{table}

\begin{table}[h!]
    \centering
    \caption{Qualitative comparison on the snow group. Prompt: `Snow rocky mountains peaks canyon. snow blanketed rocky mountains surround and shadow deep canyons. The canyons twist and bend through the high elevat'. Top: Baseline; Bottom: FPSAttention.}
    \label{tab:vis_train}
    \begin{tabular}{cccc}
        \includegraphics[width=0.22\textwidth]{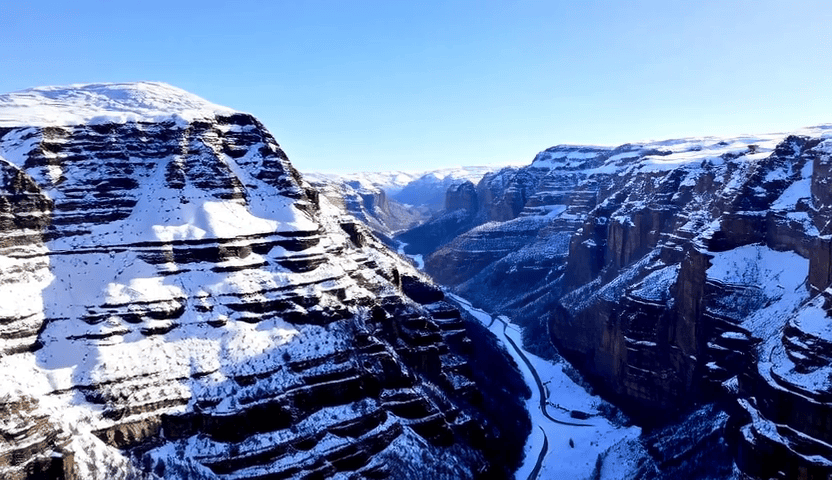} &
        \includegraphics[width=0.22\textwidth]{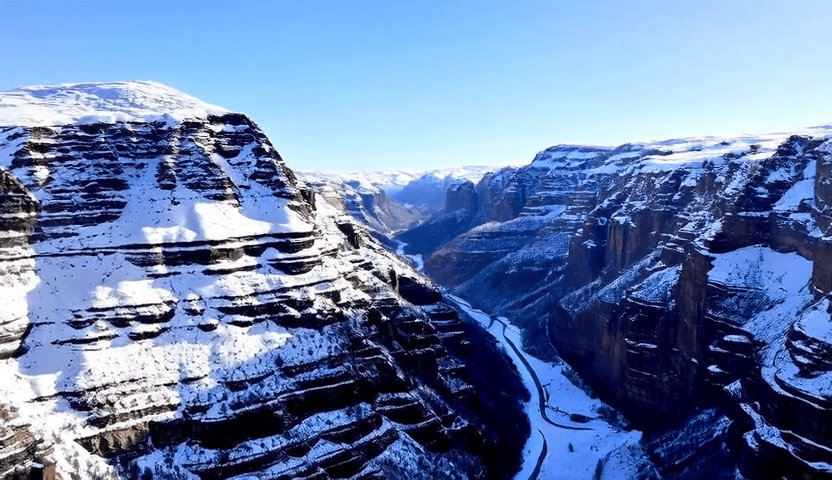} &
        \includegraphics[width=0.22\textwidth]{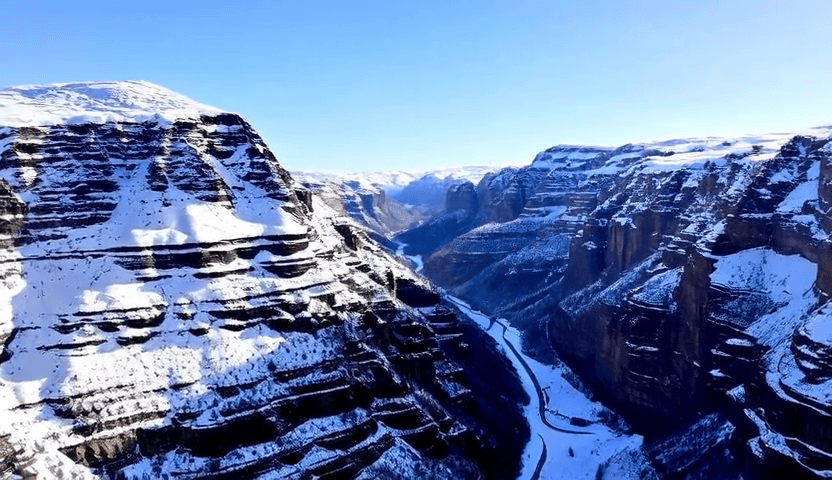} &
        \includegraphics[width=0.22\textwidth]{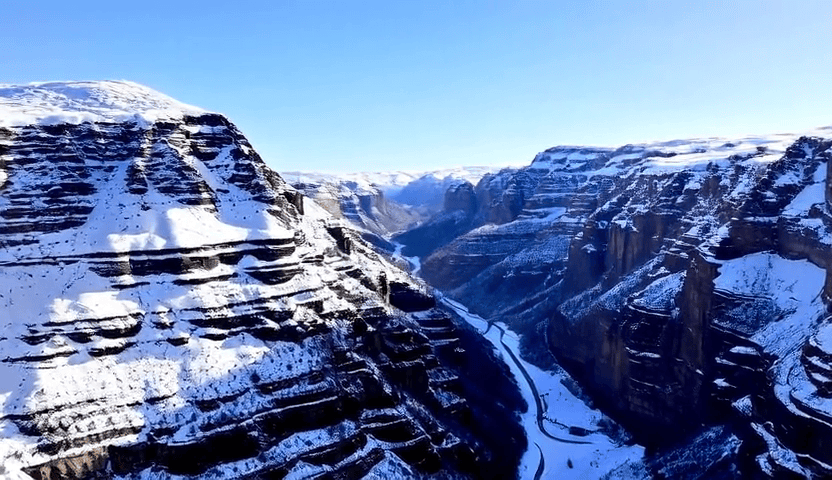} \\
        \multicolumn{4}{c}{\textbf{Baseline}} \\
        \includegraphics[width=0.22\textwidth]{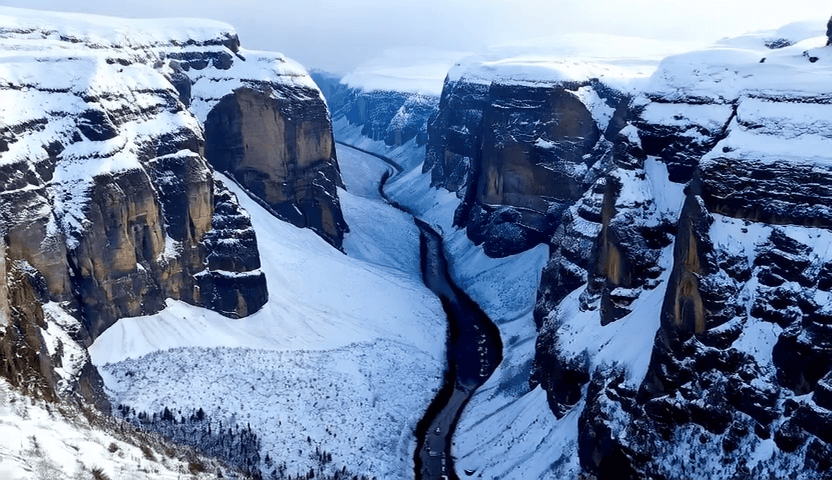} &
        \includegraphics[width=0.22\textwidth]{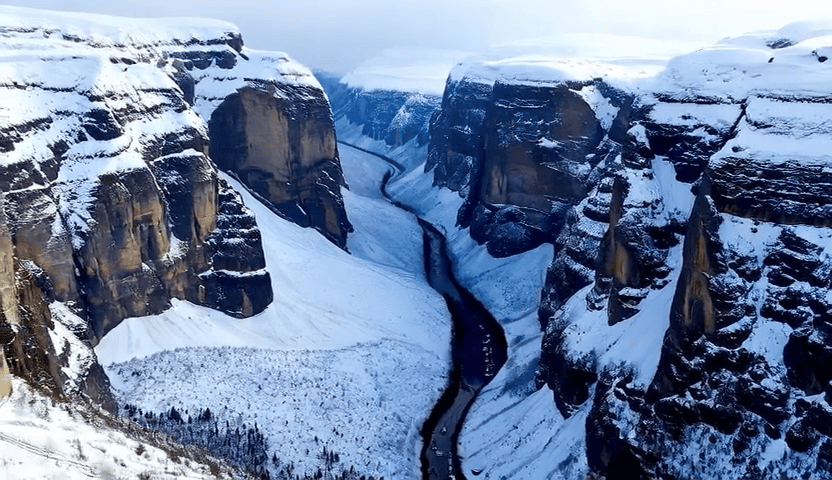} &
        \includegraphics[width=0.22\textwidth]{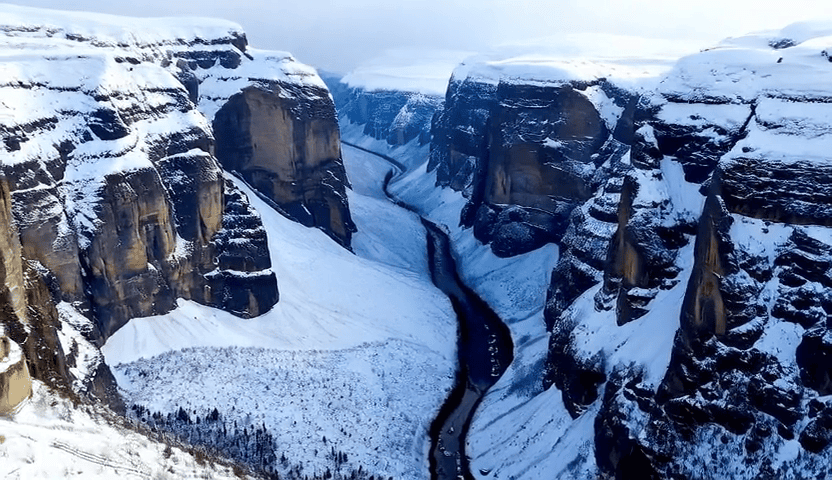} &
        \includegraphics[width=0.22\textwidth]{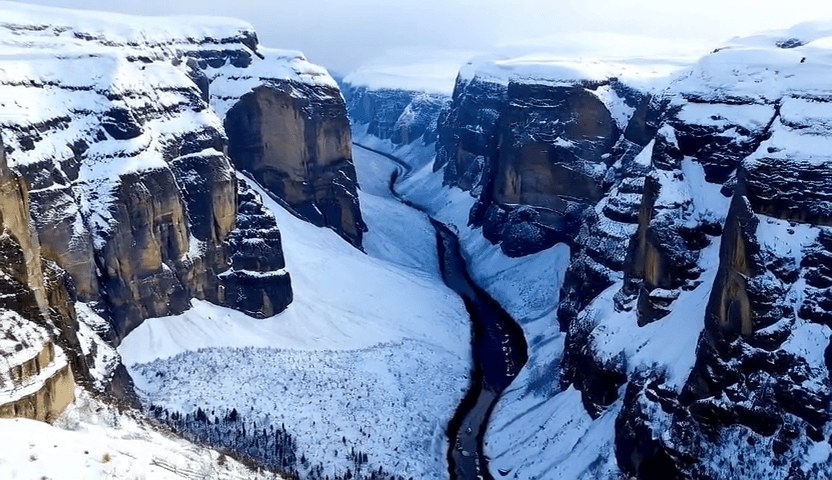} \\
        \multicolumn{4}{c}{\textbf{FPSAttention}} \\
    \end{tabular}
\end{table}

\begin{table}[h!]
    \centering
    \caption{Qualitative comparison on the book group. Prompt: `A person is reading book'. Top: Baseline; Bottom: FPSAttention.}
    \label{tab:vis_train}
    \begin{tabular}{cccc}
        \includegraphics[width=0.22\textwidth]{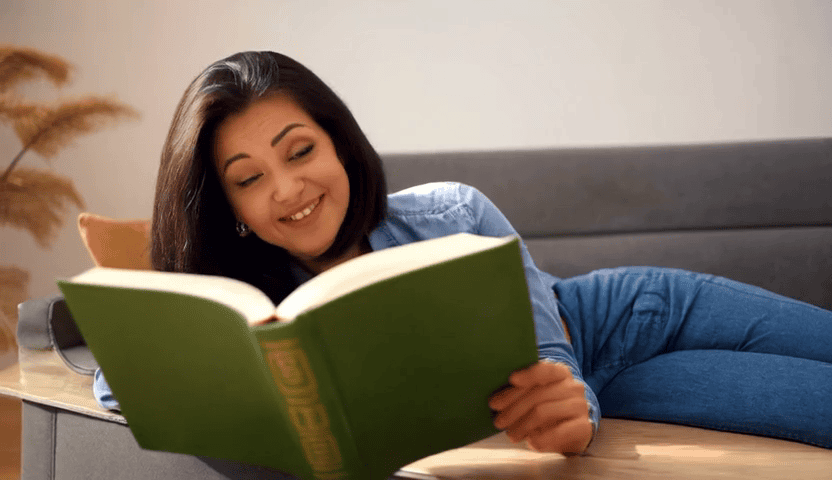} &
        \includegraphics[width=0.22\textwidth]{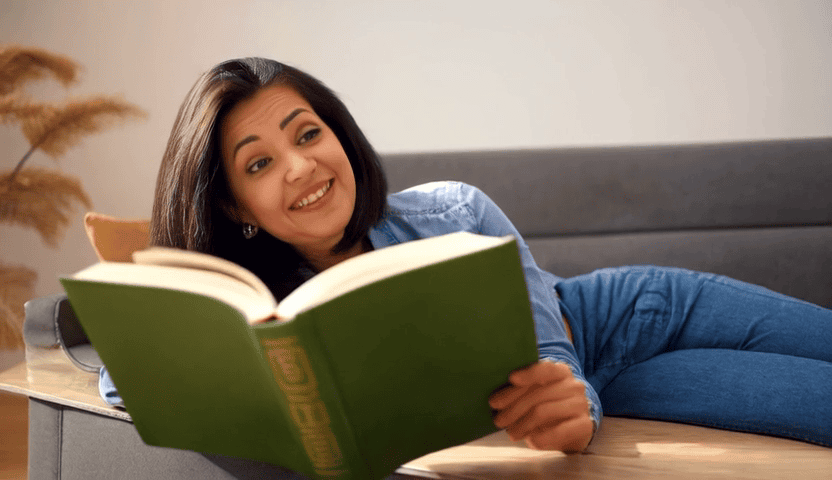} &
        \includegraphics[width=0.22\textwidth]{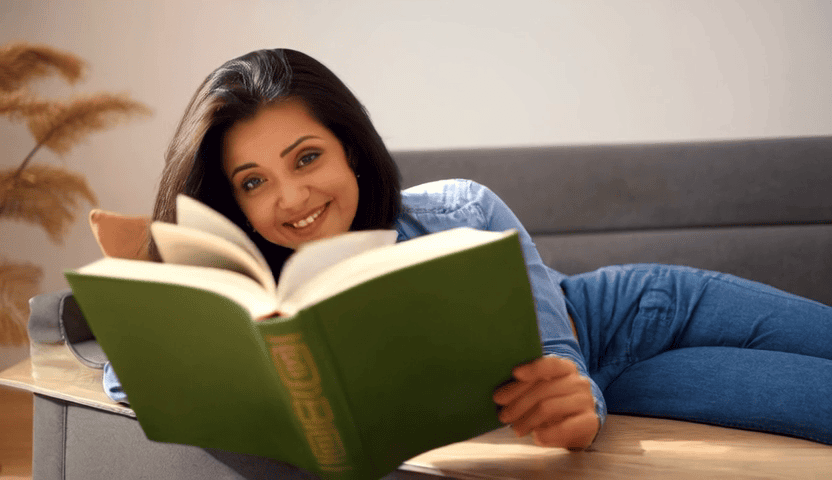} &
        \includegraphics[width=0.22\textwidth]{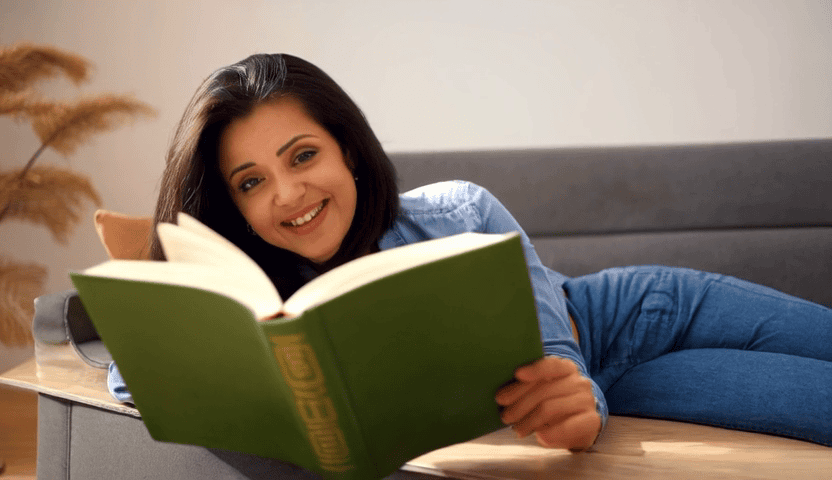} \\
        \multicolumn{4}{c}{\textbf{Baseline}} \\
        \includegraphics[width=0.22\textwidth]{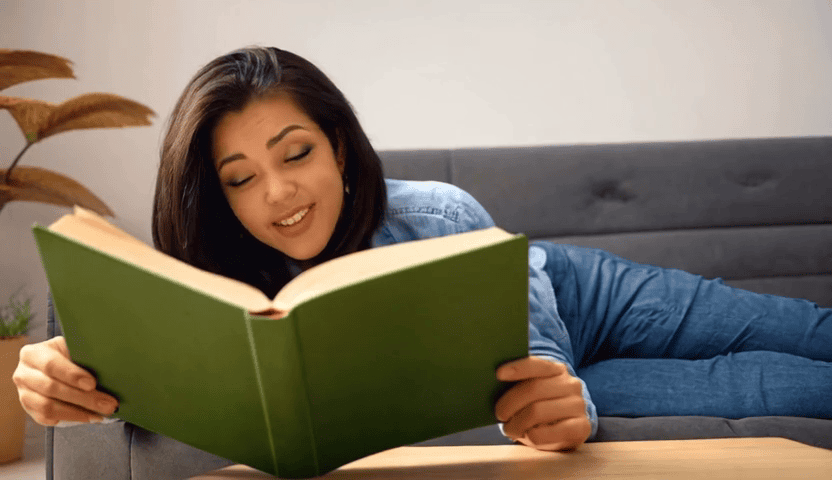} &
        \includegraphics[width=0.22\textwidth]{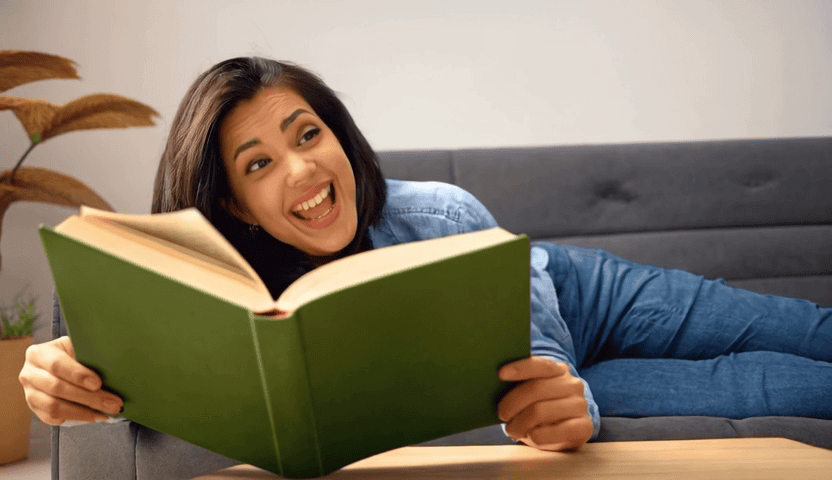} &
        \includegraphics[width=0.22\textwidth]{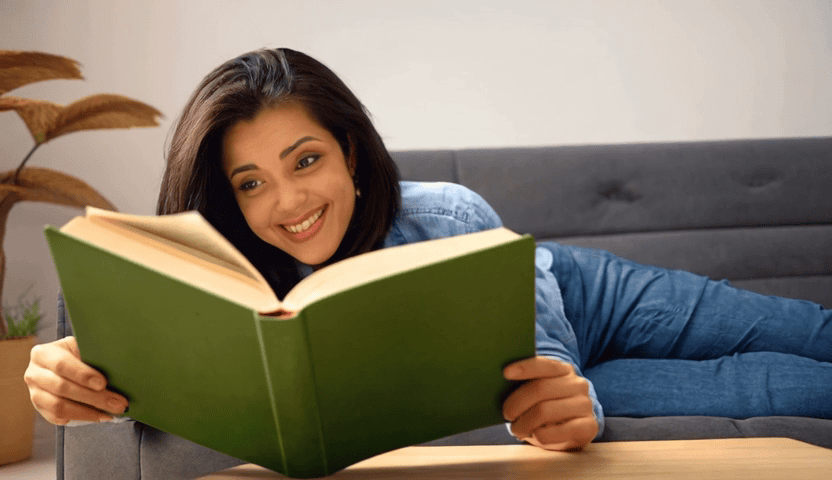} &
        \includegraphics[width=0.22\textwidth]{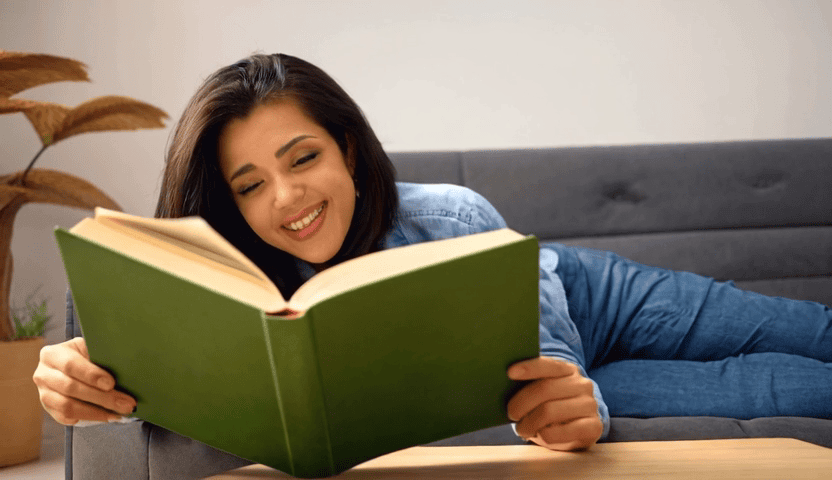} \\
        \multicolumn{4}{c}{\textbf{FPSAttention}} \\
    \end{tabular}
\end{table}

\begin{table}[h!]
    \centering
    \caption{Qualitative comparison on the space group. Prompt: `An astronaut flying in space'. Top: Baseline; Bottom: FPSAttention.}
    \label{tab:vis_train}
    \begin{tabular}{cccc}
        \includegraphics[width=0.22\textwidth]{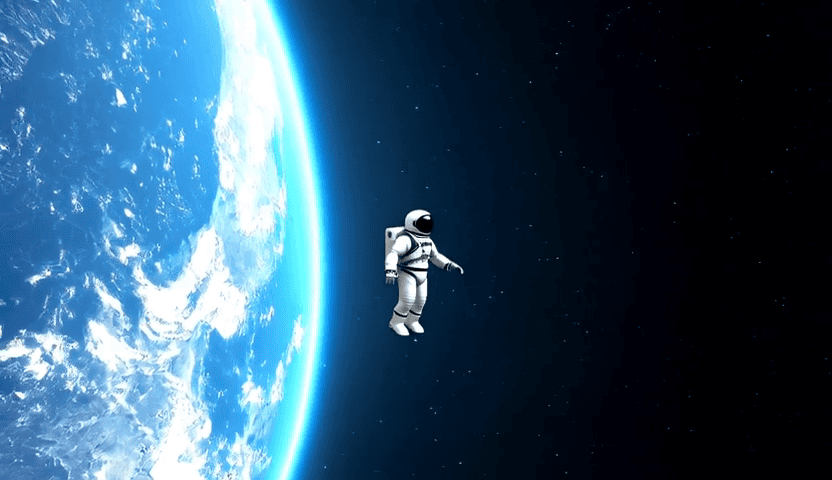} &
        \includegraphics[width=0.22\textwidth]{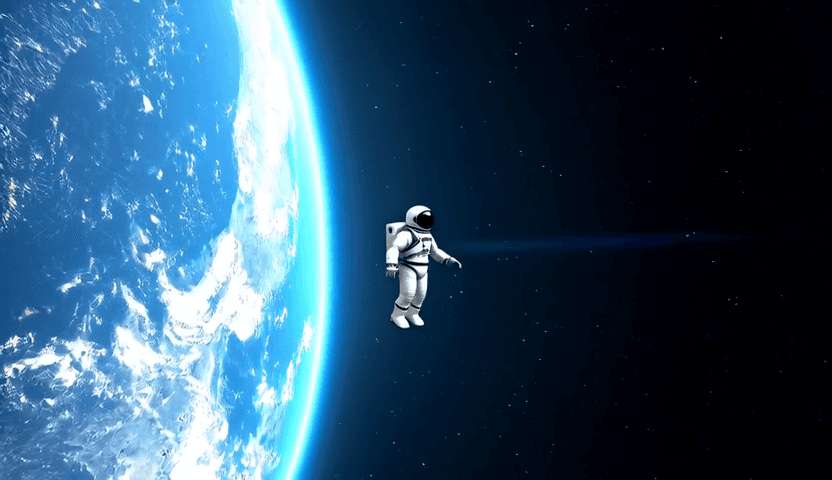} &
        \includegraphics[width=0.22\textwidth]{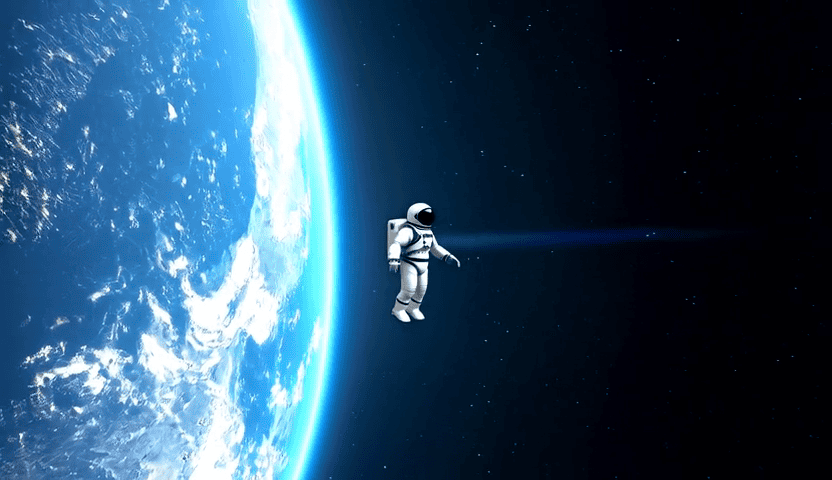} &
        \includegraphics[width=0.22\textwidth]{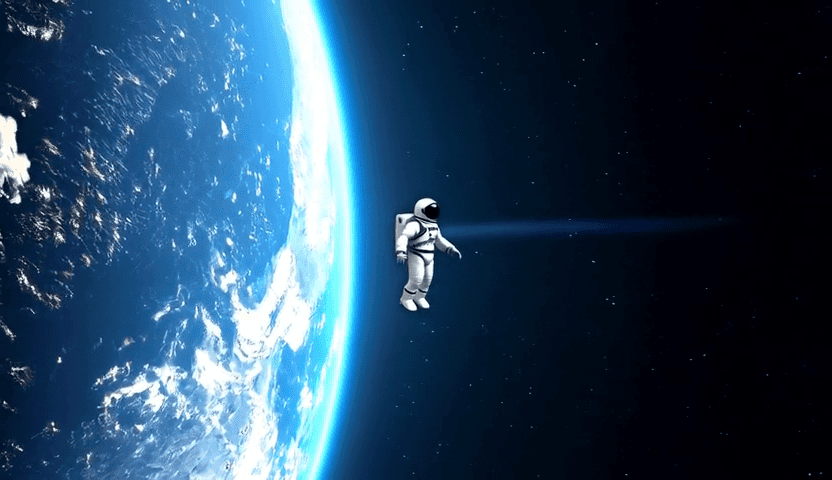} \\
        \multicolumn{4}{c}{\textbf{Baseline}} \\
        \includegraphics[width=0.22\textwidth]{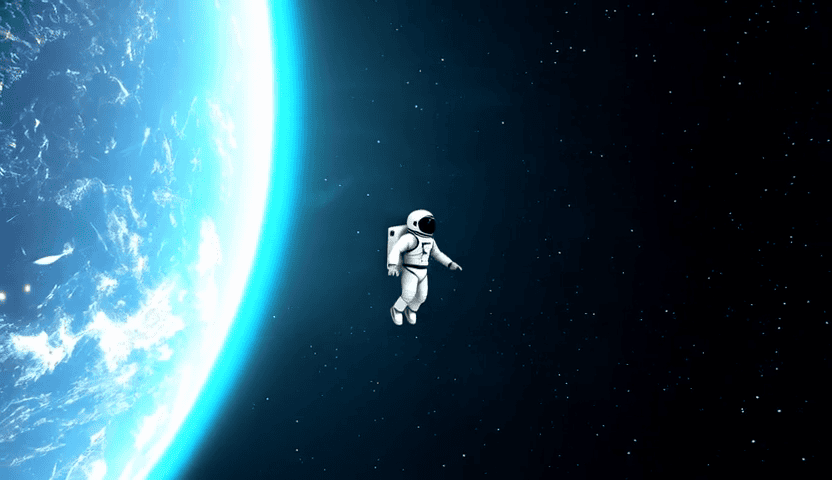} &
        \includegraphics[width=0.22\textwidth]{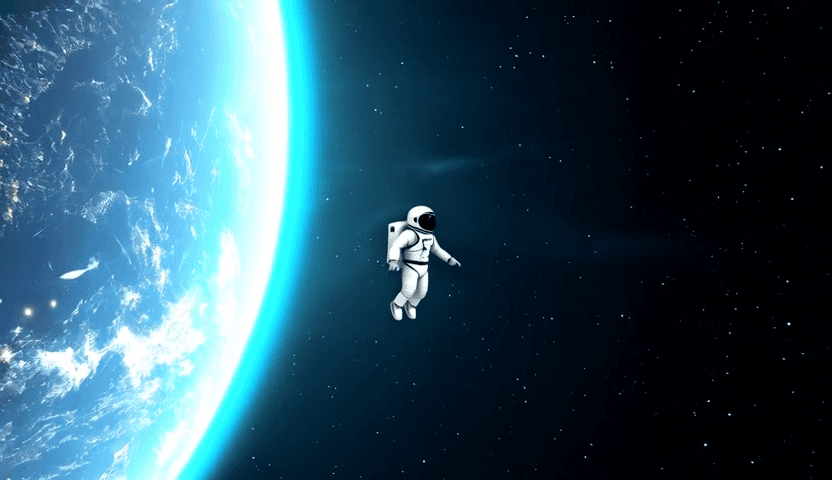} &
        \includegraphics[width=0.22\textwidth]{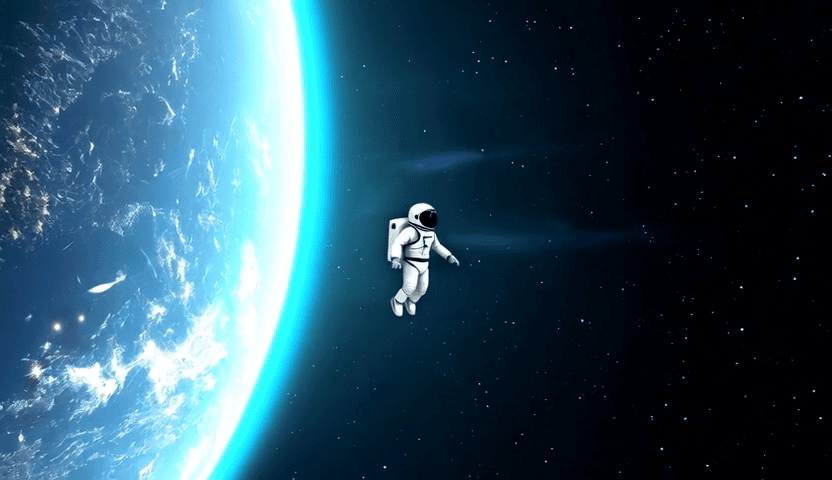} &
        \includegraphics[width=0.22\textwidth]{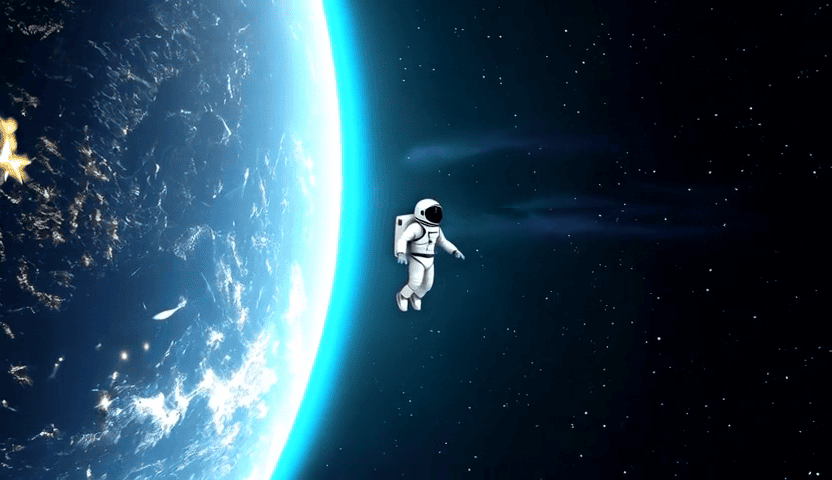} \\
        \multicolumn{4}{c}{\textbf{FPSAttention}} \\
    \end{tabular}
\end{table}

\begin{table}[h!]
    \centering
    \caption{Qualitative comparison on the panda group. Prompt: `A panda drinking coffee in a cafe in Paris'. Top: Baseline; Bottom: FPSAttention.}
    \label{tab:vis_train}
    \begin{tabular}{cccc}
        \includegraphics[width=0.22\textwidth]{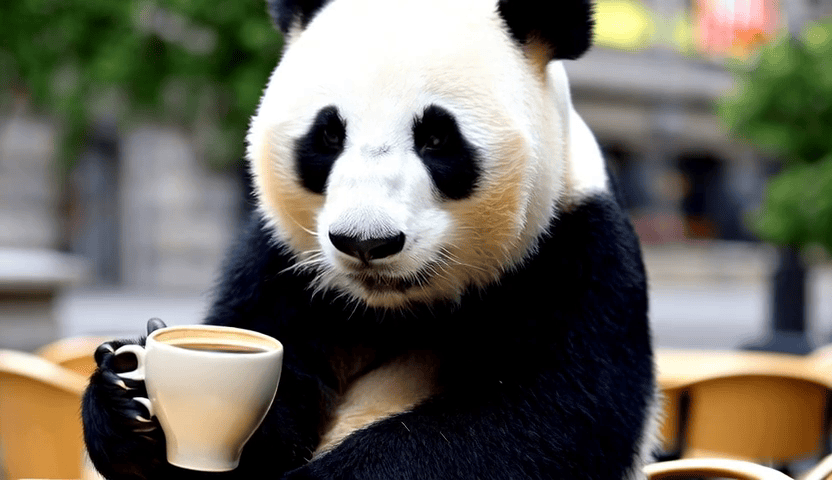} &
        \includegraphics[width=0.22\textwidth]{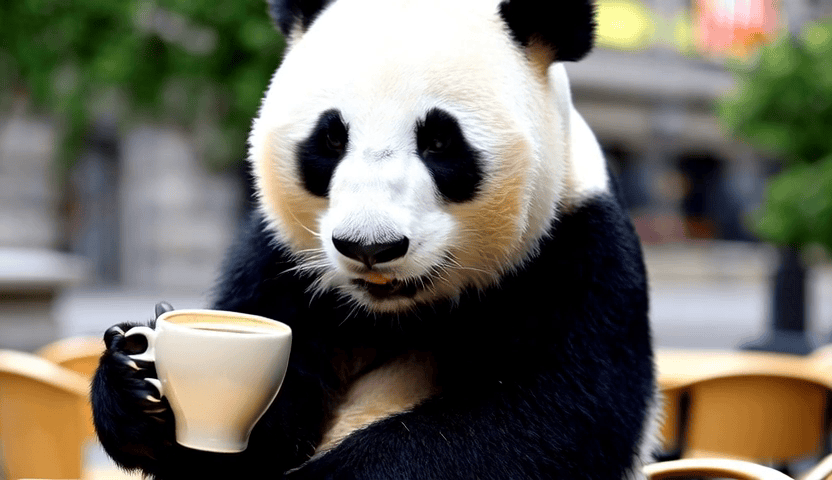} &
        \includegraphics[width=0.22\textwidth]{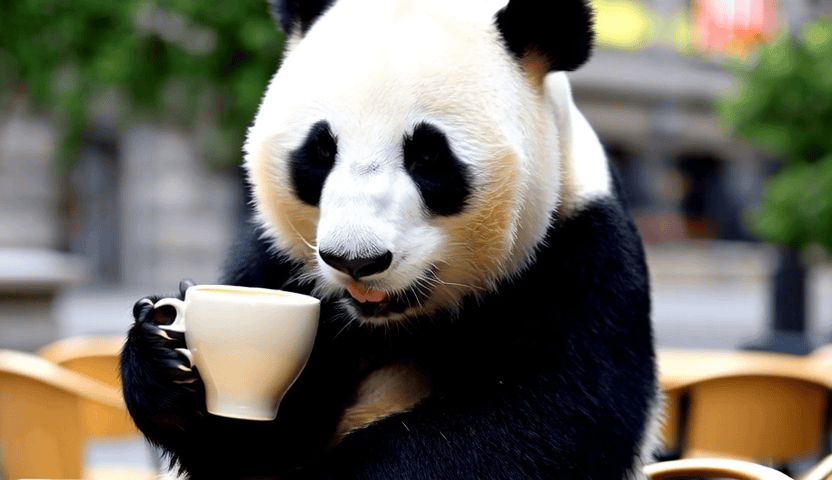} &
        \includegraphics[width=0.22\textwidth]{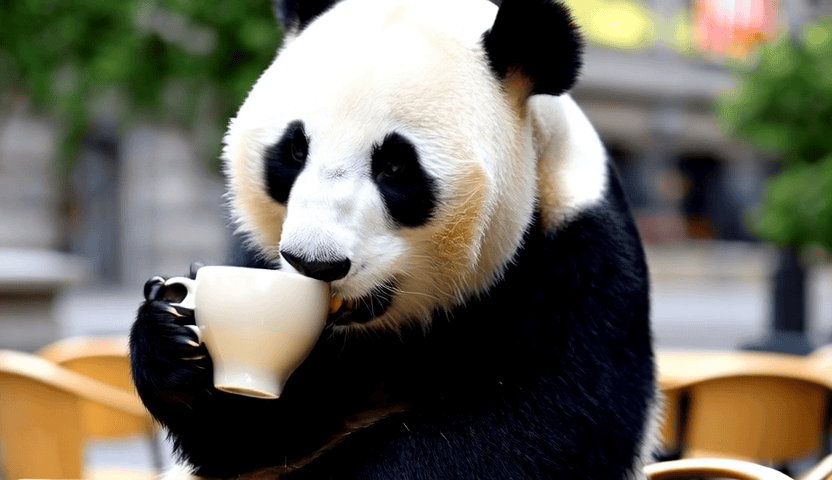} \\
        \multicolumn{4}{c}{\textbf{Baseline}} \\
        \includegraphics[width=0.22\textwidth]{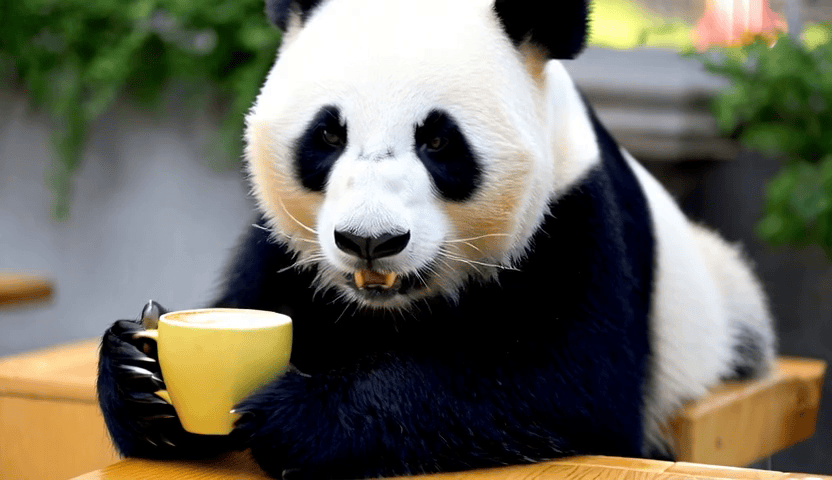} &
        \includegraphics[width=0.22\textwidth]{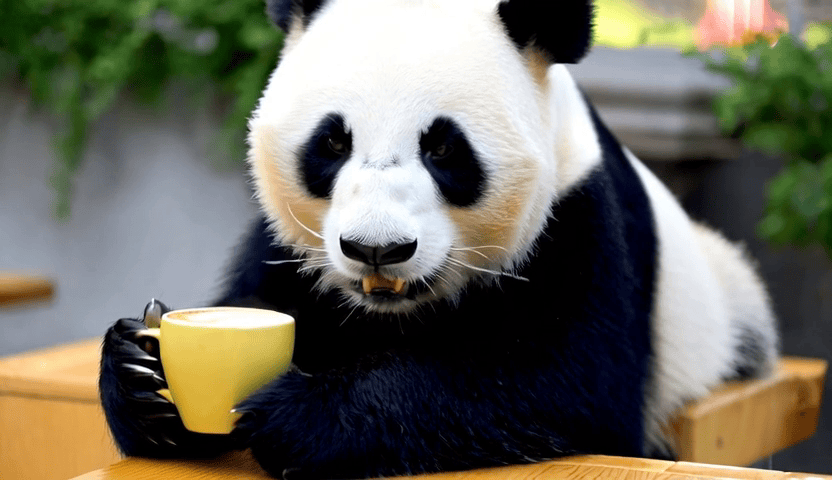} &
        \includegraphics[width=0.22\textwidth]{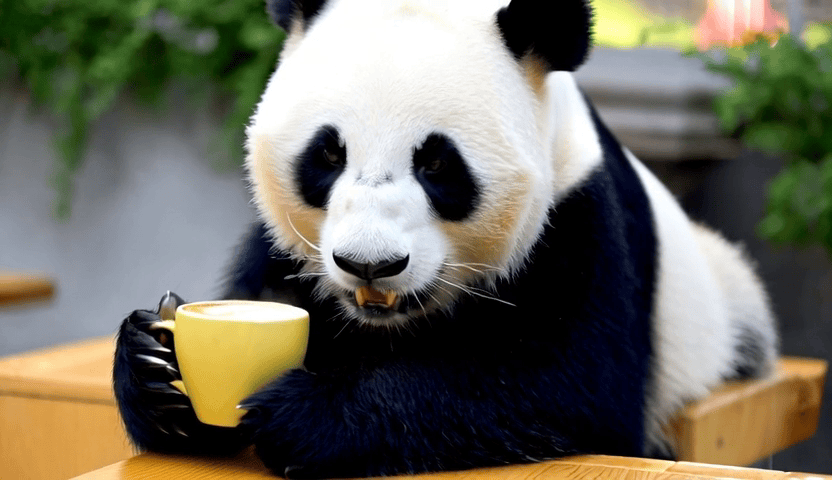} &
        \includegraphics[width=0.22\textwidth]{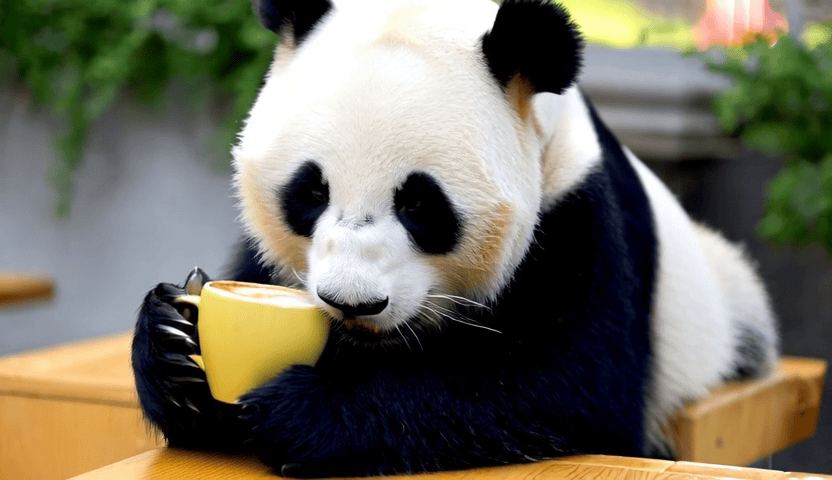} \\
        \multicolumn{4}{c}{\textbf{FPSAttention}} \\
    \end{tabular}
\end{table}

\newpage

\clearpage
\newpage

\bibliographystyle{plain}
\bibliography{reference}

\end{document}